\documentclass[lettersize,journal]{IEEEtran}
\usepackage{amsmath,amsfonts}

\usepackage{algorithmic}
\usepackage{algorithm}
\usepackage{array}
\usepackage[caption=false,font=normalsize,labelfont=sf,textfont=sf]{subfig}
\usepackage{textcomp}
\usepackage{stfloats}
\usepackage{url}
\usepackage{verbatim}
\usepackage{graphicx}
\usepackage{cite}
\usepackage{multirow}
\newcolumntype{M}[1]{>{\centering\arraybackslash}m{#1}}
\usepackage{xcolor}

\begin{document}

\title{An optimal pairwise merge algorithm improves the quality and consistency of nonnegative matrix factorization}

\author{Youdong Guo, Timothy E. Holy
        % <-this % stops a space
\thanks{Youdong Guo is with the Department of Neuroscience, Washington University in St. Louis, St. Louis, MO, 63130, USA}
\thanks{Timothy E. Holy is with the Departments of Neuroscience and Biomedical Engineering, Washington University in St. Louis, St. Louis, MO, 63130, USA (email: holy@wustl.edu) }}


% \IEEEpubidadjcol 

\maketitle

\begin{abstract}
Non-negative matrix factorization (NMF) is a key technique for feature extraction and widely used in source separation. 
However, existing algorithms may converge to poor local minima, or to one of several minima with similar objective value but differing feature parametrizations. 
% Additionally, the performance of NMF greatly depends on the number of components, but choosing the optimal count remains a challenge. 
Here we show that some of these weaknesses may be mitigated by performing NMF in a higher-dimensional feature space and then iteratively combining components with an analytically-solvable pairwise merge strategy. 
Experimental results demonstrate our method helps non-ideal NMF solutions escape to better local optima and achieve greater consistency of the solutions.
% Iterative merging also provides an efficient and informative framework for choosing the number of components. 
Despite these extra steps, our approach exhibits similar computational performance to established methods by reducing the occurrence of ``plateau phenomenon'' near saddle points. 
Moreover, the results also illustrate that our method is compatible with different NMF algorithms.
Thus, this can be recommended as a preferred approach for most applications of NMF.
\end{abstract}


\section{Introduction}
\label{introduction}
\IEEEPARstart{N}{on-negative} matrix factorization has gained popularity as an exceptional unsupervised method in extracting meaningful patterns from images, text, audio, etc.\cite{lee1999learning, cichocki2009fast, leplat2020blind, aonishi2022imaging, giovannucci2019caiman, gillis2021distributionally, tan2012automatic}. In many applications, it uncovers underlying components that have plausible interpretations, making it valuable for source separation and facilitating subsequent analyses such as classification\cite{jing2012snmfca, guillamet2003introducing, jiao2020hyper, xie2017decoding}, regression\cite{elosua2021spotlight}, and segmentation\cite{giovannucci2019caiman, maruyama2014detecting}. Given a non-negative data matrix $\mathbf{X}\in\mathbb{R}^{m\times n}$, NMF approximates it as a product of two non-negative matrices,
\begin{align}
    \begin{aligned}
        \label{NMF_eq}
        \mathbf{X} \approx \mathbf{\mathbf{WH}}\  \textrm{subject to}\ \ \mathbf{\mathbf{W}}\geq 0,\ \mathbf{H}\geq0
    \end{aligned}
\end{align}
where $\mathbf{W}\in\mathbb{R}^{m\times r}$ and $\mathbf{H}\in\mathbb{R}^{r\times n}$. $r$ is the rank of this factorization, and the columns and rows of $\mathbf{W}$ and $\mathbf{H}$, respectively, are called the components (a.k.a., features). $\mathbf{W}$ and $\mathbf{H}$ are obtained by minimizing an objective function which measures the similarity between $\mathbf{X}$ and $\mathbf{W}\mathbf{H}$. Here we employ one of the most widely used objective functions, the square Euclidean distance\cite{lin2007projected, hsieh2011fast}
 \begin{align}
    \begin{aligned}
    \label{SED_obj}
     D_E(\mathbf{X}||\mathbf{W}\mathbf{H})=\frac{1}{2}\left\lVert \mathbf{X}-\mathbf{WH}\right\rVert^2_F
    \end{aligned}
\end{align}

Although NMF is a crucial tool for source separation, its real-world performance is challenged.
Since (\ref{SED_obj}) is non-convex when optimizing $\mathbf{W}$ and $\mathbf{H}$ simultaneously, NMF is NP-hard \cite{vavasis2010complexity, gillis2014nonnegative}. 
Hence, the convergence to a global optimum is not deterministically guaranteed\cite{wang2012nonnegative}. 
Several algorithms have been developed to minimize (\ref{SED_obj}), including multiplicative updates (MU), alternating least squares (ALS), alternating nonnegative least squares (ANLS), and several variants of coordinate descent\cite{lee1999learning, lin2007projected, cichocki2009fast}. However, these methods are only be guaranteed to converge to stationary points\cite{gillis2014nonnegative}.
Algorithms have also been developed to find the global optimum for near-separable matrices\cite{arora2012computing, moitra2016almost, gillis2013fast, pan2019generalized} but not all matrices $\mathbf{X}$ are of this form and these algorithms may yield poor outcomes or be computationally inefficient when the requirement of separability is not satisfied
\cite{gillis2014nonnegative}. 
Other strategies for mitigating NP-hardness include improved initializations\cite{esposito2021review, fathi2023initialization} or adding regularization terms to the objective function to favor solutions with certain desirable characteristics\cite{jiao2020hyper, yuan2020convex, deng2022graph, huang2020robust, ince2022weighted}. However, to the best of our knowledge, none of them ensure the discovery of the global optimum of NMF.

To mitigate the problems discussed above, we propose a new approach called NMF-Merge, which utilizes over-complete NMF to generate a more comprehensive representation of the data and subsequently recombines extra components to reach the desired number of components. 
Conceptually, this approach is motivated by the idea that convergence of NMF becomes poor when one is forced to make difficult trade-offs in describing different features of the data matrix; thus, performing an initial factorization with an excessive number of components grants the opportunity to escape such constraints and reliably describe the full behavior of the data matrix. 
Later, any redundant or noisy components are identified and merged together.
We test our proposed methodology on multiple real-world datasets with different NMF algorithms, and our findings underscore two primary contributions: 
1) NMF-Merge typically achieves equivalent or superior local optima compared to standard NMF, and can help existing NMF solutions escape to more favorable local optima.
2) Starting from different initializations, the final $\mathbf{W}$ and $\mathbf{H}$ are typically more consistent.
% 3) Via the pairwise merge penalty, the method provides an efficient and informative heuristic for estimating the rank.

Beyond these contributions,  NMF-Merge boasts two additional advantages. 
First, the total computational demand is typically comparable to that of standard NMF, and our approach occasionally achieves faster completion times than its standard counterpart despite the additional steps in our pipeline.
The final NMF converges quickly by avoiding plateau phenomenon as illustrated via example in Appendix~\ref{Illustration of plateau phenomenon in NMF} and more systematically in Fig.~\ref{fig_trace_t_example}-\ref{fig_trace_phase}.
Second, our approach is highly adaptable when analyzing real-world data. 
When adjustments to the number of components are necessary, our method seamlessly increases or decreases components, allowing users to efficiently and easily adjust NMF solutions in an interactive setting.

\section{NMF-Merge}
In this section, we will describe the whole pipeline and each part of NMF-Merge. The whole pipeline of NMF-Merge is visualized in Fig.  \ref{whole_pipe}.  
\subsection{The Pipeline of NMF-Merge}
\label{The Pipeline of NMF-Merge}
\begin{figure*}[!t]
    \centering
    \includegraphics[width= 6.6in]{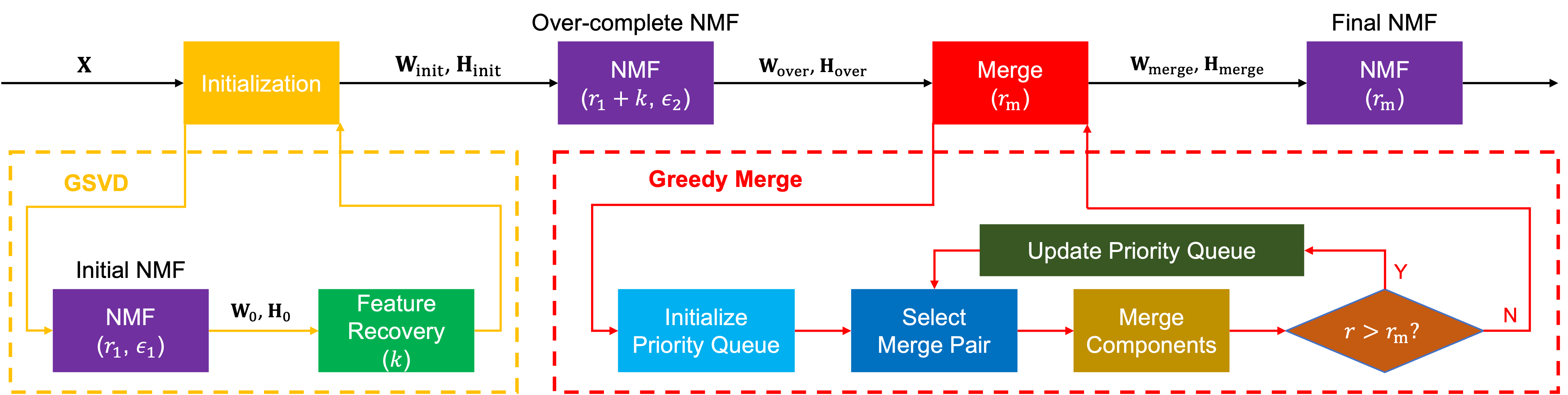}%
    \caption{The whole pipeline of NMF-Merge.}
    \label{whole_pipe}
\end{figure*}

The core of the proposed method consists of the ``Over-complete NMF'' and ``Merge'' stages in Fig.  \ref{whole_pipe}. 
Supposing the desired rank is $r_\mathrm{m}$, over-complete NMF (with rank $r_\mathrm{o}>r_\mathrm{m}$) is less likely to miss important features or patterns within the data.
Assuming the outputs of over-complete NMF are $\mathbf{W}_\mathrm{over}$ and $\mathbf{H}_\mathrm{over}$, we then iteratively reduce the number of components using a globally optimal pairwise merge algorithm, resulting in $\mathbf{W}_\mathrm{merge}$ and $ \mathbf{H}_\mathrm{merge}$ with $r_\mathrm{m}$ components. 
This merge process eliminates the extra components.
% , but also aids in determining the number of features present in $\mathbf{X}$ (discussed in Sec. \ref{NMF Components number selection}) by monitoring the loss during merging.
The merge sequence is chosen by a greedy algorithm, which selects the remaining pair with least merge penalty.
The merged matrices, $\mathbf{W}_\mathrm{merge}$ and $ \mathbf{H}_\mathrm{merge}$, are subsequently refined through NMF (Final NMF) under a stringent tolerance to ensure full optimization and to achieve the final factorization. 
While one could start from rank-$r_\mathrm{o}$ NMF directly, to improve the capture of lower rank features we instead adopt a generalized singular value decomposition (GSVD)-based feature recovery method to initialize over-complete NMF\cite{guo2024gsvdnmfrecoveringmissingfeatures}.
The GSVD feature recovery starts from an initial NMF (rank $r_1$), which roughly learns features from $\mathbf{X}$ (resulting in $\mathbf{W}_0$ and $\mathbf{H}_0$). 
Next, we augment $\mathbf{W}_0$ and $\mathbf{H}_0$ with $k$ additional components produced by the GSVD-based feature recovery method, which deterministically finds features present in $\mathbf{X}$ that are not in initial NMF solutions (summarized in Sec. \ref{Generalized singular value decomposition (GSVD) based feature recovery}). 
Over-complete NMF's rank $r_o = r_1+k$ should ideally be large enough to cover all meaningful features in $\mathbf{X}$. Our GSVD recovery method provides an option to incrementally increase the number of components for over-complete NMF, if the initial NMF components is not adequate. 
Although $r_1$ could be a parameter when using the pipeline as a method to implement NMF, we set and test $r_1=r_\mathrm{m}$ in this paper.
In practice, if a non-ideal NMF solution has already been obtained, it is advantageous to use this non-ideal NMF solution as the output of the ``initial NMF'' step and then proceed with subsequent steps in the pipeline to improve the solution.
% This is valuable for applications that need to run NMF multiple times under different conditions\cite{giovannucci2019caiman}. \tim{Is this last sentence necessary? Is it the only situation in which this might be useful? What do you actually mean by "different conditions"?}

\subsection{Pairwise Components Merge}
\label{Pairwise Components Merge}
    Consider two components, $\mathbf{w}_p$, $\mathbf{h}_p$ and $\mathbf{w}_q$, $\mathbf{h}_q$, where $\mathbf{w}_p,\ \mathbf{w}_q\in\mathbb{R}^{m\times 1}$ are $p$-th and $q$-th columns in $\mathbf{W}$, $\mathbf{h}_p,\ \mathbf{h}_q\in\mathbb{R}^{n\times 1}$ are $p$-th and $q$-th columns in $\mathbf{H}^\mathrm{T}$. $\mathbf{w}_p$, $\mathbf{w}_q$ are normalized to $\lVert\mathbf{w}_p\rVert^2=\lVert\mathbf{w}_q\rVert^2=1$, where $\lVert\cdot\rVert$ denotes the $\ell_2$ norm. The matrix $\mathbf{W}\mathbf{H}$ can be denoted as
    \begin{align}
        \mathbf{W}\mathbf{H} = \sum_{i\neq p,q}\mathbf{w}_i\mathbf{h}_i^\mathrm{T}+ \mathbf{w}_p\mathbf{h}_p^\mathrm{T}+\mathbf{w}_q\mathbf{h}_q^\mathrm{T}
    \end{align}
    Pairwise merge evaluates a potential merge into a single component as described by
    \begin{align}
        \begin{aligned}
        \mathbf{W}_{\mathrm{merge}}\mathbf{H}_{\mathrm{merge}} = \sum_{i\neq p,q}\mathbf{w}_i\mathbf{h}_i^\mathrm{T}+\mathbf{w}_\mathrm{m}\mathbf{h}_\mathrm{m}^\mathrm{T}
        \end{aligned}
    \end{align}
    We propose to obtain $\mathbf{w}_\mathrm{m}$ and $\mathbf{h}_\mathrm{m}$ by minimizing
    \begin{equation}
    \label{DE_X_X_merge}
        D_E(\mathbf{W}\mathbf{H}||\mathbf{W}_{\mathrm{merge}}\mathbf{H}_{\mathrm{merge}}) 
        = \lVert\mathbf{w}_p\mathbf{h}_p^\mathrm{T}+\mathbf{w}_q\mathbf{h}_q^\mathrm{T}-\mathbf{w}_\mathrm{m}\mathbf{h}_\mathrm{m}^\mathrm{T}\rVert^2
    \end{equation}
    Note that this makes no reference to the initial data matrix $\mathbf{X}$. One could solve (\ref{DE_X_X_merge}) as a rank-1 NMF problem in $\mathbf{w}_p\mathbf{h}_p^\mathrm{T}+\mathbf{w}_q\mathbf{h}_q^\mathrm{T}$\cite{giovannucci2019caiman}, exploiting standard or polynomial time algorithms \cite{vavasis2010complexity, ghalamkari2022fast}.
    
    However, in Appendix~\ref{Derivation of optimal merge model} we show that any $\mathbf{w}_\mathrm{m}$ that minimizes (\ref{DE_X_X_merge}) is a linear combination of $\mathbf{w}_p$ and $\mathbf{w}_q$. Thus, we model the merged component $\mathbf{w}_\mathrm{m}$ as% the linear combination of $\mathbf{w}_p$ and $\mathbf{w}_q$
    \begin{align}
        \begin{aligned}
        \label{linear_merge_model}
        \mathbf{w}_\mathrm{m} = \alpha\mathbf{w}_p+\beta\mathbf{w}_q, \  \textrm{subject to}\ \ \lVert\mathbf{w}_\mathrm{m}\rVert^2=1
        \end{aligned}
    \end{align}
    We seek an analytic solution for $\alpha$ and $\beta$.
    Define the magnitudes $c = \mathbf{w}_p^\mathrm{T}\mathbf{w}_q$ ($0\leq c\leq1$), $h_p = \lVert\mathbf{h}_\mathbf{p}\rVert$, $h_p = \lVert\mathbf{h}_\mathbf{q}\rVert$, $h_m = \lVert\mathbf{h}_\mathbf{m}\rVert$.
    We also define $g$ as $\mathbf{h}_p^\mathrm{T}\mathbf{h}_q = gh_ph_q$ (so that $g$ is to $\mathbf{h}$ as $c$ is to $\mathbf{w}$).
    To handle the normalization constraint, a Lagrange-multiplier term is added to (\ref{DE_X_X_merge}), 
    \begin{align}
        \begin{aligned}
        \label{Merge_penalty_P}
            P =& D_E(\mathbf{X}||\mathbf{X}_{\mathrm{merge}})+\lambda\left[\left(\alpha\mathbf{w}_p+\beta\mathbf{w}_q\right)^2-1\right]\\
            =& h_m^2-2\left(\alpha+\beta c\right)\mathbf{h}_p^\mathrm{T}\mathbf{h}_\mathrm{m}-2\left(\alpha c+\beta\right)\mathbf{h}_q^\mathrm{T}\mathbf{h}_\mathrm{m}\\
            & +h_p^2+2cgh_ph_q+h_q^2
            +\lambda\left(\alpha^2+2\alpha\beta c+\beta^2-1\right)
        \end{aligned}
    \end{align}
    The stationary point of $P$ with respect to $\mathbf{h}_\mathbf{m}$ is 
    \begin{align}
    \label{optimal_h_m}
        \mathbf{h}_\mathrm{m} = \left(\alpha+\beta c\right)\mathbf{h}_p+\left(\alpha c+\beta\right)\mathbf{h}_q
    \end{align}
    Substituting (\ref{optimal_h_m}) into (\ref{Merge_penalty_P}) yields
    \begin{align}
        \label{Merge_penalty_PQ}
        \begin{aligned}
            P = & -\mathbf{u}^\mathrm{T}\mathbf{Q}_1\mathbf{u}+\lambda\left(\mathbf{u}^\mathrm{T}\mathbf{Q}_2\mathbf{u}-1\right)\\
            &+h_p^2+2cgh_ph_q+h_q^2
        \end{aligned}
    \end{align}
    where $\mathbf{u}=\left[\alpha,\ \beta\right]^\mathrm{T}$, and where $\mathbf{Q}_1$ and $\mathbf{Q}_2$ are defined in Appendix~\ref{Derivation of final solution}.
    $\mathbf{Q}_2$ is a positive semi-definite matrix because $c \leq 1$. At the minimum of $P$, $\mathbf{u}$ and $\lambda$ satisfy the generalized eigenvalue problem
    \begin{align}
        \begin{aligned}
            \label{gen_eigen}
            \mathbf{Q}_1\mathbf{u}=\lambda \mathbf{Q}_2\mathbf{u}
        \end{aligned}
    \end{align}
    Solving (\ref{gen_eigen}) yields,
    \begin{align}
        \begin{aligned}
            \label{eigen_values}
            &\lambda_\mathrm{max} = \frac{\tau}{2}+\sqrt{\frac{\tau^2}{4}-\delta}\\
            &\lambda_\mathrm{min} = \frac{\tau}{2}-\sqrt{\frac{\tau^2}{4}-\delta}
        \end{aligned}
    \end{align}
    where
    \begin{align}
        \begin{aligned}
        \label{tau&delta}
            & \tau = h_p^2+2cgh_ph_q+h_q^2\\
            & \delta = \left(1-c^2\right)\left(1-g^2\right)h_p^2h_q^2
        \end{aligned}
    \end{align}
    and
    \begin{align}
        \begin{aligned}
            \label{u_vec}
            \mathbf{u} & = \frac{1}{\sqrt{1+2c\xi+\xi^2}}\left[
                                \xi,\ 1
                            \right]^\mathrm{T}
        \end{aligned}
    \end{align}
    where
    \begin{align}
        \begin{aligned}
            \label{xi}
                \xi = \frac{\lambda_\mathrm{max}-h_q^2-cgh_ph_q}{\left(gh_ph_q+ch_q^2\right)}
        \end{aligned}
    \end{align}
    In Appendix~\ref{sec:proof-nonneg}, we show that $\mathbf{\xi}$ is nonnegative, and thus (\ref{linear_merge_model}) and (\ref{optimal_h_m}) preserve the nonnegativity of the solution.
    At this minimum, $P$ is 
    \begin{align}
        \begin{aligned}
            \label{minimum_P}
            P = \lambda_\mathrm{min}
        \end{aligned}
    \end{align}
    
    Implementing this analytically-optimal merge is extremely efficient with time complexity $O(m+n)$, the cost of computing the dot products of components.
    In practice, merging single pair of components is summarized in Algorithm \ref{merge2to1}.
    \begin{algorithm}[H]
        \caption{Merge Two Components}\label{merge2to1}
        \begin{algorithmic}
            \STATE 
            \STATE \textbf{Input:} $\mathbf{w}_p,\mathbf{h}_p,\mathbf{w}_q,\mathbf{h}_q$. $\mathbf{w}_p,\mathbf{w}_q$
            are required to be normalized to unit $\ell_2$-norm length.
            \STATE \textbf{Output:} $P$, the merge penalty (loss)
            \STATE \hspace{0.5cm} $c \gets \mathbf{w}_p^\mathrm{T}\mathbf{w}_q$
            \STATE \hspace{0.5cm} $h_p \gets \lVert \mathbf{h}_p \rVert$
            \STATE \hspace{0.5cm} $h_q \gets \lVert \mathbf{h}_q \rVert$
            \STATE \hspace{0.5cm} $g \gets \mathbf{h}_p^\mathrm{T} \mathbf{h}_q / (h_p h_q)$
            \STATE \hspace{0.5cm} Compute $\lambda_\mathrm{min}$ from (\ref{eigen_values})
            \STATE \hspace{0.5cm} $P \gets \lambda_\mathrm{min}$
            \STATE \textbf{If $\mathbf{w}_\mathrm{m}$, $\mathbf{h}_\mathrm{m}$ are needed:}
            \STATE \hspace{0.5cm} Compute $\mathbf{u}$ from (\ref{u_vec})
            \STATE \hspace{0.5cm} $\alpha \gets u_1$
            \STATE \hspace{0.5cm} $\beta \gets u_2$
            \STATE \hspace{0.5cm} $\mathbf{w}_\mathrm{m} \gets \alpha\mathbf{w}_p+\beta\mathbf{w}_q$
            \STATE \hspace{0.5cm} $\mathbf{h}_\mathrm{m} \gets \left(\alpha+\beta c\right)\mathbf{h}_p+\left(\alpha c+\beta\right)\mathbf{h}_q$
        \end{algorithmic}
        \label{merge_2_alg}
    \end{algorithm}

    Compared to a general-purpose NMF algorithm, this analytic solution is many-fold more efficient. 
    To evaluate the merge penalty of a single pair (the main bottleneck in picking which pair to merge, see below), only $P$ is required, which can be computed in $O(m+3n)$, primarily attributed to computing dot products of the component vectors (the first four lines of Algorithm~\ref{merge2to1}).
    When $r=1$, the time complexity of a single iteration of standard (naive) HALS\cite{cichocki2009fast} is $O(mn)$, potentially orders of magnitude slower than our analytic merge.
    However, in this special case we already have a rank-2 factored representation of the target matrix $\mathbf{X}_2=[\mathbf{w}_p\,\mathbf{w}_q][\mathbf{h}_p\, \mathbf{h}_q]^\mathrm{T}$, and rank-1 NMF can be accelerated by skipping the computation of the dense matrix $\mathbf{X}_2$, and instead using the factored form for all internal computations.
    Even with this optimization, the time complexity of one iteration of accelerated rank 2-to-1 HALS (Algorithm 2 in \cite{cichocki2009fast}) is $O(6(m+n))$, more than twice that of Algorithm~\ref{merge2to1}.
    More significantly, rank 2-to-1 NMF generally needs multiple iterations (empirically of order 10 iterations for random initialization), thus implying that these two methods differ in performance by at least 20-fold. Empirical benchmarking (not shown) bears out this large performance advantage for Algorithm~\ref{merge2to1}.

    The overall merge phase exploits a greedy strategy.  
    Given that $\mathbf{W}_r$ and $\mathbf{H}_r$ consist of $r$ components, a priority queue is initialized with merge errors (\ref{minimum_P}) for all possible pairs of components. 
    The optimal reduction from $r$ components to $r-1$ components is determined by selecting the component pair associated with the minimal merge error from the priority queue.
    After finishing the merge, we compute (\ref{minimum_P}) between the new component and each remaining previous component and insert it into the priority queue. During later merges, we continue to draw the lowest value from the priority queue, dropping any associated with previously merged components (and thus no longer available).
    The maximum number ($N_{\mathrm{merge}}$) of pairwise merge-candidates required to consolidate from $r$ components down to $r_\mathrm{m}$ components is
    \begin{align}
        \begin{aligned}
        \label{total_n_merge}
            N_{\mathrm{merge}} =\frac{r(3r-11)}{2}-(r_\mathrm{m}-4)(r_\mathrm{m}+1)
        \end{aligned}
    \end{align}
    While (\ref{total_n_merge}) is quadratic in $r$, the total time complexity of the merge phase is $O(N_{\mathrm{merge}}(m+n))$, which is smaller than the time ($O(mnr)$) for just a single NMF iteration ($r\ll\min(m,n)$).
    Thus, the merge phase does not contribute substantially to the overall time complexity of the pipeline.
    If the desired number of components is known, we stop merging when that number is reached. 
    % Alternatively, we merge until only one component remains, monitoring the merge error of the chosen pair at each iteration. Use of this merge penalty trajectory in selecting the number of components is discussed in Sec.\ref{NMF Components number selection}.

    \subsection{Generalized singular value decomposition (GSVD) based feature recovery}
    \label{Generalized singular value decomposition (GSVD) based feature recovery}
    As mentioned in Sec.\ref{The Pipeline of NMF-Merge}, we use GSVD-based feature recovery to increase the robustness of over-complete NMF \cite{guo2024gsvdnmfrecoveringmissingfeatures}.
While one could alternatively perform over-complete NMF directly with $r_\mathrm{o} = r_1 + k$ components, our choice to augment standard NMF with a deterministic step (later refined with over-complete NMF) is guided by three principal considerations. 
First, as previously noted, the NP-hard characteristic of NMF may lead over-complete NMF to settle on suboptimal solutions that fail to encapsulate all the necessary features, potentially causing the final results to converge to an unfavorable local minimum. 
Second, in practice, if the number of components in over-complete NMF turns out to be insufficient due to a misestimated initial NMF rank or an incorrect determination of the number of extra components needed, adding more components is both efficient and convenient.  
This step can speed up the entire pipeline as it avoids a complete re-run of the over-complete NMF. 
Third, as mentioned in Sec. \ref{The Pipeline of NMF-Merge}, it enables the NMF-Merge to improve existing non-ideal NMF solutions.
The full details of GSVD-based method are described in \cite{guo2024gsvdnmfrecoveringmissingfeatures}.

    \section{Experimental Results}
    In this section, we introduce the datasets, compare  standard NMF and the proposed NMF-Merge, and analyze convergence. All experiments were performed using Julia 1.10 on the Washington University in St.~Louis RIS scientific computing platform with Intel\_Xeon\_Gold6242CPU280GHz with 8G RAM. 
    We tested and compared the performance with Hierarchical alternating least squares (HALS)\cite{cichocki2009fast}, Greedy Coordinate Descent (GCD)\cite{hsieh2011fast}, Alternating Least Squares using Projected Gradient descent (ALSPGrad)\cite{lin2007projected}, Multiplicative Updating (MU)\cite{lee2000algorithms} to perform NMF. Later analysis on consistency, efficiency and parameters is perform on the results from using HALS since it provides the best combination of speed and solution quality among all widely used NMF algorithms\cite{gillis2014nonnegative, gillis2012accelerated, gillis2011nonnegative}.
    In this manuscript, we used a stopping criterion based on the maximum relative change in the Frobenius norm of the columns of $\mathbf{W}\in\mathbb{R}^{m\times r}$ and the rows of $\mathbf{H}\in\mathbb{R}^{r\times n}$
    \begin{align}
        \begin{aligned}
            \label{hals_stop_condition}
            \lVert\mathbf{w}_j^{(k+1)}-\mathbf{w}_j^{(k)}\rVert^2&\leq\epsilon\lVert\mathbf{w}_j^{(k+1)}+\mathbf{w}_j^{(k)}\rVert^2\\
            \lVert\mathbf{h}_j^{(k+1)}-\mathbf{h}_j^{(k)}\rVert^2&\leq\epsilon\lVert\mathbf{h}_j^{(k+1)}+\mathbf{h}_j^{(k)}\rVert^2
        \end{aligned}
    \end{align}
    for all $j=1, 2, \dots, r$,
    where $\mathbf{w}_j$ and $\mathbf{h}_j$ are the $j$-th column and $j$-th row in $\mathbf{W}$ and $\mathbf{H}$ respectively, and where $\mathbf{v}^{(k)}$ indicates the value of $\mathbf{v}$ on the $k$-th iteration.
    Unless otherwise specified, $\epsilon=10^{-4}$ was used and the maximum number of iterations for NMF was set to $10^8$, so that all NMF stages in the pipeline converged by these criteria given $\epsilon$. The relative fitting error between $\mathbf{W}\mathbf{H}$ and the input matrix $\mathbf{X}$, $100\lVert \mathbf{X}-\mathbf{W}\mathbf{H} \rVert_2^2/\lVert \mathbf{X} \rVert_2^2$, was used to evaluate the local optima of NMF. 
    This measures how well the factorization fit the input matrix.
    
    \subsection{Data sets}
    \label{data sets}    
\begin{figure}[!t]
        \centering
        \subfloat[]{\includegraphics[width=1.5in]{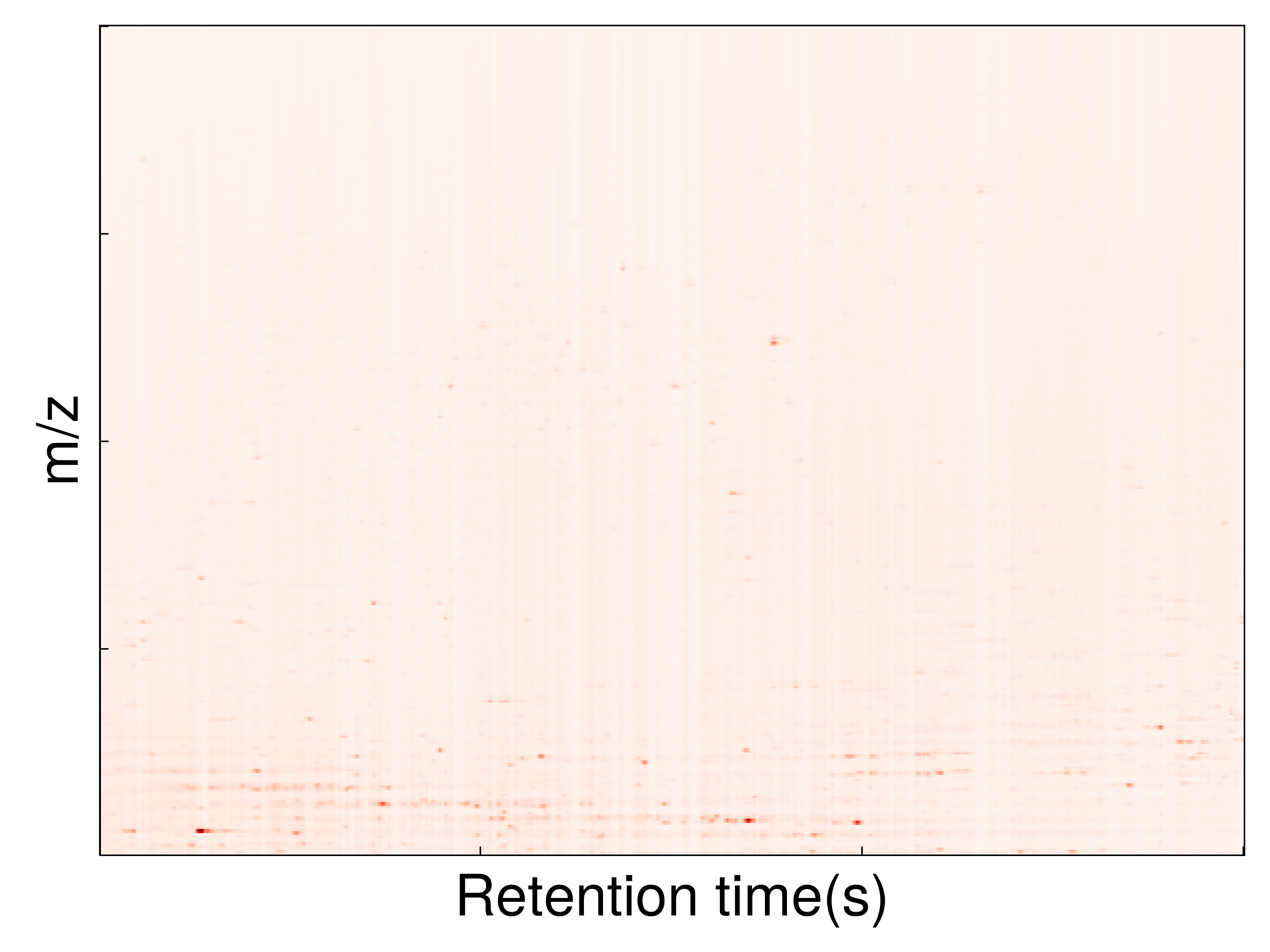}}%
        \subfloat[]{\includegraphics[width=1.5in]{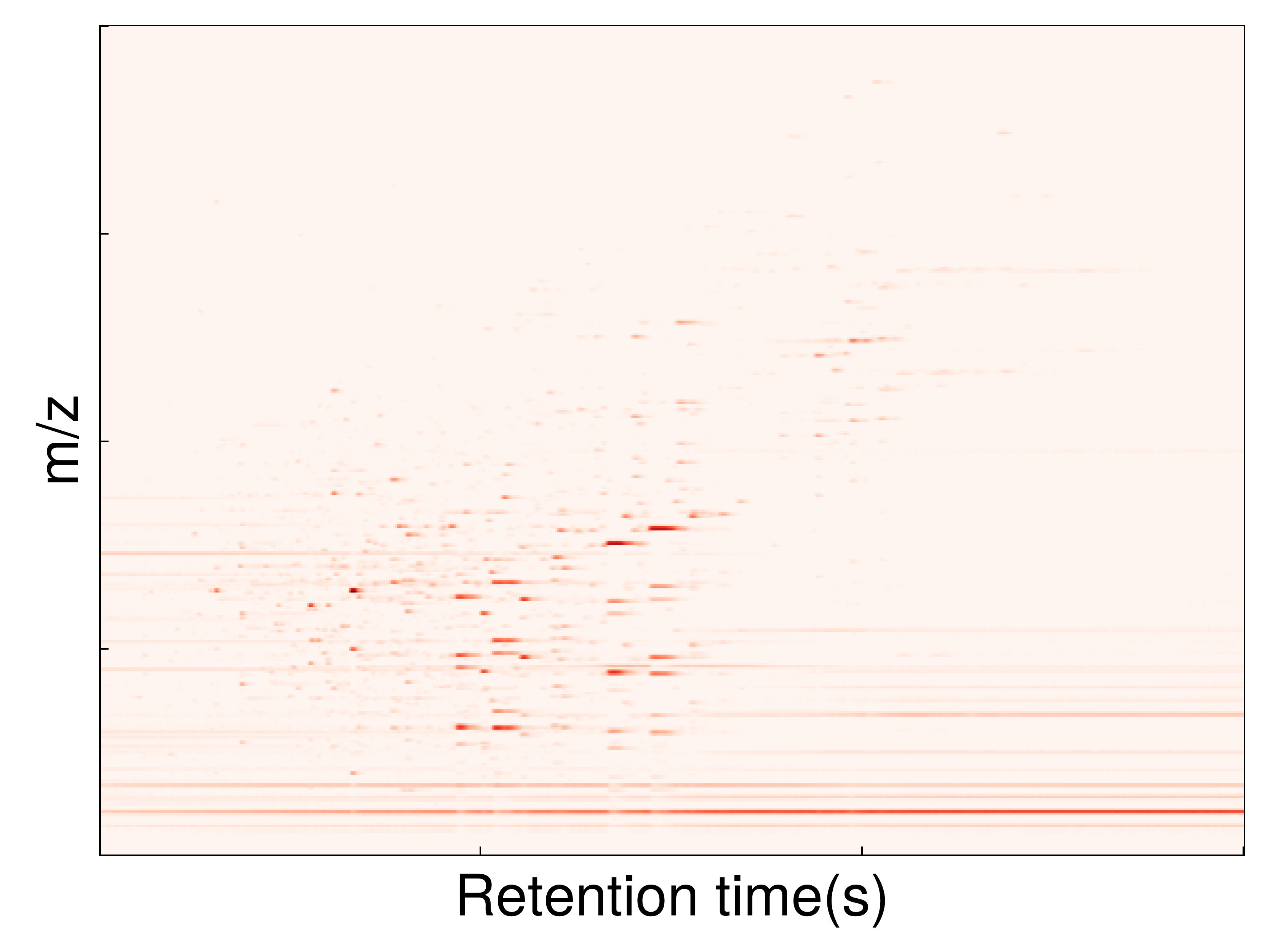}}%
        
        \subfloat[]{\includegraphics[width=1.5in]{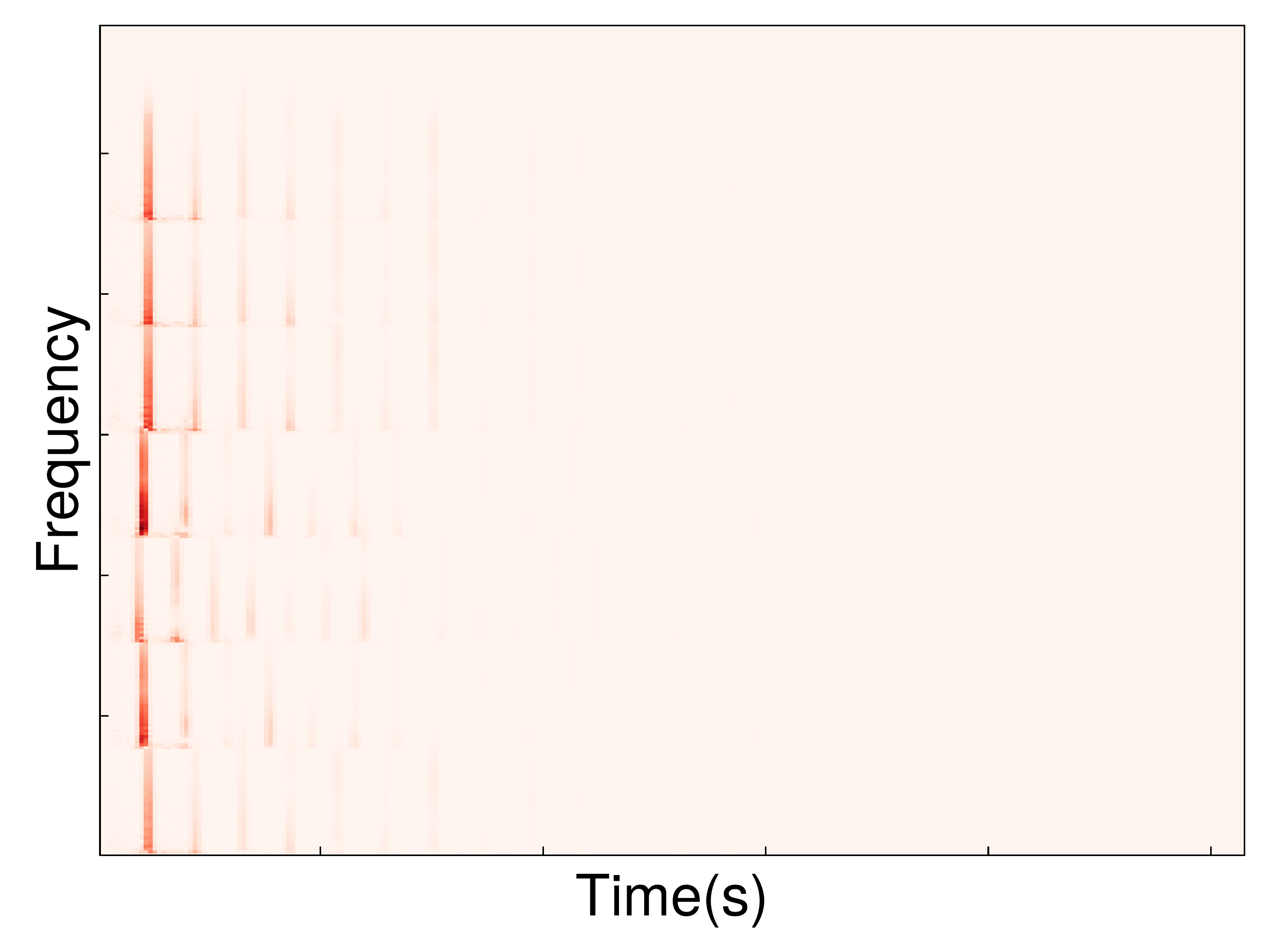}}%
        \subfloat[]{\includegraphics[width=1.5in]{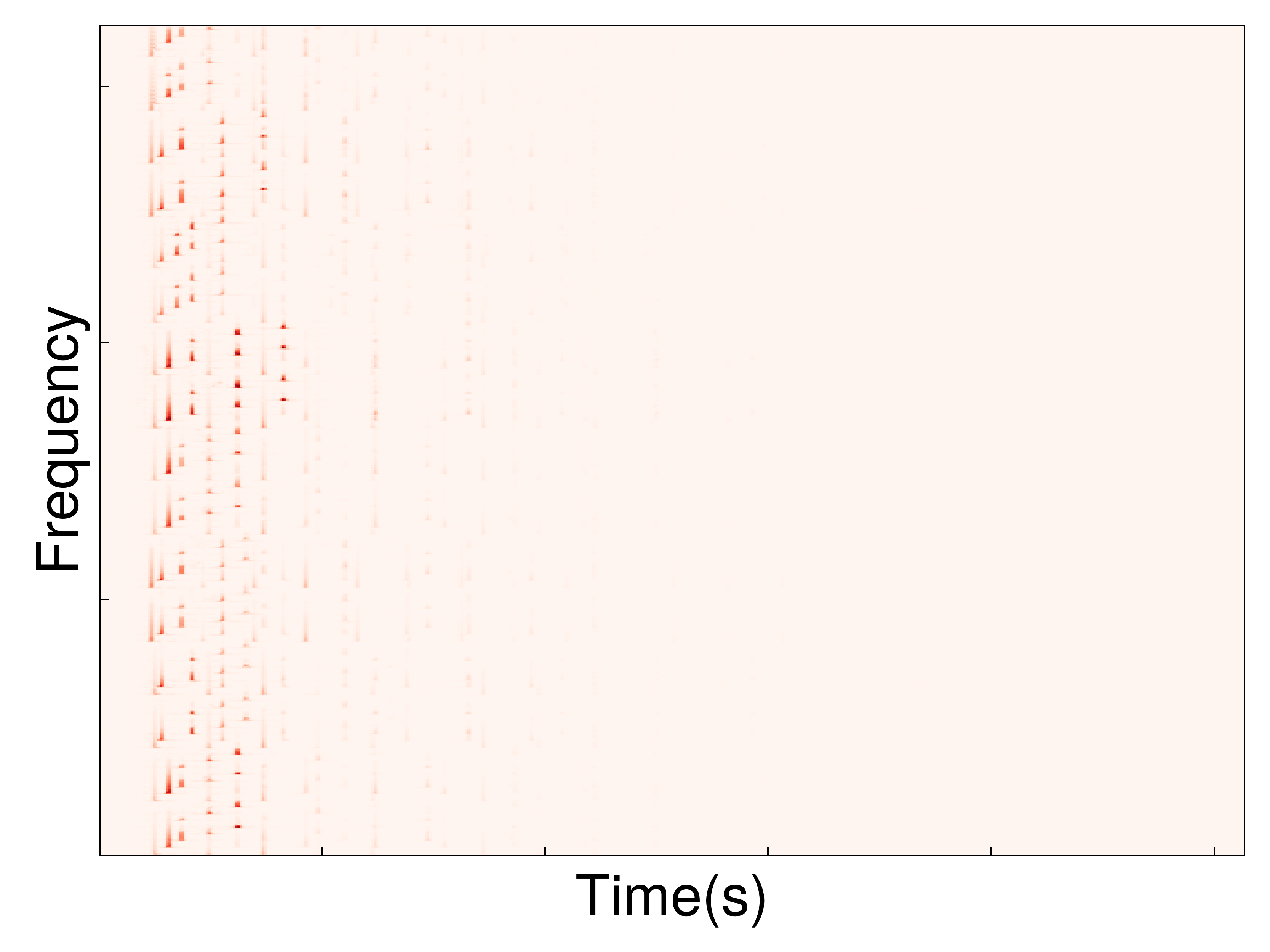}}%
        \caption{(a) LCMS1. (b) LCMS2. (c) The amplitude spectrogram of ``Mary had a little lamb''. (d) The amplitude spectrogram of ``Prelude and Fugue No.1 in C major''}
        \label{fig_datasets}
    \end{figure}

    In the study, six datasets were employed to evaluate the performance of NMF-Merge. 
    These include two liquid chromatography-mass spectrometry datasets, LCMS1 and LCMS2, two audio datasets, and two face image datasets.  LCMS data are shown as Fig.  \ref{fig_datasets}(a) and (b) and their sources are (doi:10.25345/C5KP7TV9T and doi:10.25345/C58C9R77T). 
    The audio datasets feature the first measure of ``Mary had a little lamb'' and the first 30 seconds of ``Prelude and Fugue No.1 in C major'' by J.S. Bach, played by Glenn Gould, both of which are taken from reference \cite{leplat2020blind}. 
    Their amplitude spectrograms, which are the short-time Fourier transforms of the recorded signals, are shown in  Fig.  \ref{fig_datasets}(c) and (d). 
    The face image datasets used are the MIT CBCL face images cited in reference \cite{weyrauch2004component} and the ORL face images from AT\&T Laboratories Cambridge (https://cam-orl.co.uk/facedatabase.html). The sizes of all six datasets are presented in Table~\ref{data_sets}. 

    We choose the number of components for the six datasets as specified in Table~\ref{dataset_table}. Where possible, these choices were based on previous literature with the same datasets\cite{leplat2020blind,cichocki2009fast,lin2007projected}.
    For the LCMS datasets not previously studied, we tested several rank-selection methods\cite{lin2020optimization,tan2012automatic, gillis2014nonnegative}. Because these methods gave divergent answers, ultimately we selected one among the recommended values based on visual inspection. To ensure our results were not overly-dependent on having the ``right'' number of components, we also tested the impact of varying components number (Fig.~\ref{fig_scatter_spec_r_spec_tol}(e)).
    \begin{table}[!t]
    \renewcommand{\arraystretch}{1.5}
        \centering
        \caption{Description of data sets \label{dataset_table}}
        \label{data_sets}
        \begin{tabular}{cM{4cm}cc} \hline 
             No.& Data sets & Size & $r$\\ \hline \hline
             1& LCMS1 & $400\times600$ & 17\\ 
             2& LCMS2 & $400\times600$ & 23\\ 
             3& Mary had a little lamb (MHLL) & $257\times294$ & 3\\ 
             4& Prelude and Fugue No.1 in C major (P\&F No.1) & $513\times647$ &13 \\  
             5& MIT CBCL face images & $361\times2429$ & 49\\  
             6& ORL face images & $10304\times400$ & 25\\ \hline
        \end{tabular}     
    \end{table}

    \subsection{NMF-Merge helps NMF converge to better local optima}
    
    % \begin{figure}[!t]
    % \centering
    %     \subfloat[]{\includegraphics[width=1.15in]{images/fig_scatter_all_r/scatters_89200_t1_4_t2_4_r_16_25.pdf}}%
    %     \subfloat[]{\includegraphics[width=1.15in]{images/fig_scatter_all_r/scatters_89747_t1_4_t2_4_r_16_25.pdf}}%
    %     \subfloat[]{\includegraphics[width=1.15in]{images/fig_scatter_all_r/scatters_mary_t1_4_t2_4_r_2_7.pdf}}%
        
    %     \subfloat[]{\includegraphics[width=1.15in]{images/fig_scatter_all_r/scatters_prelude_t1_4_t2_4_r_10_15.pdf}}
    %     \subfloat[]{\includegraphics[width=1.15in]{images/fig_scatter_all_r/scatters_CBCL_t1_4_t2_4_r_45_53.pdf}}%
    %     \subfloat[]{\includegraphics[width=1.15in]{images/fig_scatter_all_r/scatters_ORL_t1_4_t2_4_r_21_29.pdf}}
    %     \caption{The comparison of standard NMF and NMF-Merge: (a) LCMS1. (b) LCMS2. (c) Mary had a little lamb. (d) Prelude and Fugue No.1 in C major. (e) CBCL. (f) ORL.}
    %     \label{fig_scatter_all_r}
    % \end{figure}

      \label{NMF-Merge helps NMF converge to better local optima}
        \begin{figure*}[!t]
        \centering
            \subfloat[]{\includegraphics[width=6.8in]{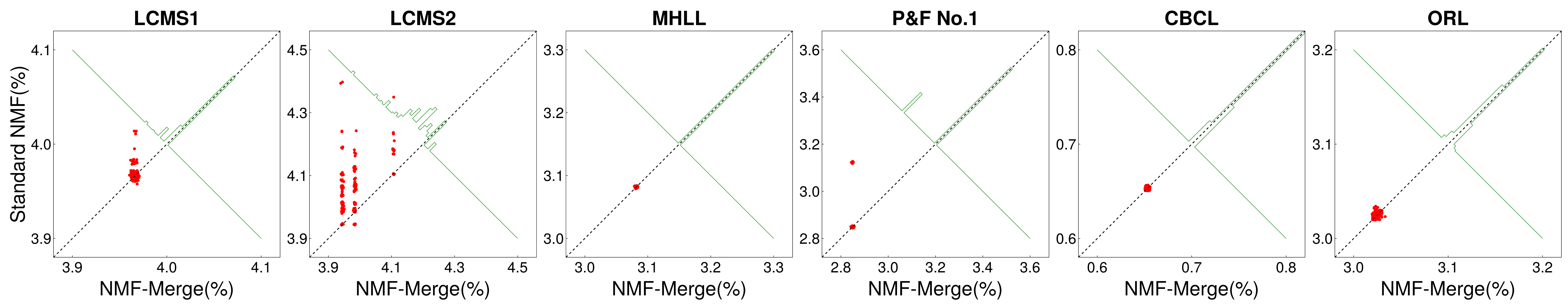}}%
            
            \subfloat[]{\includegraphics[width=6.8in]{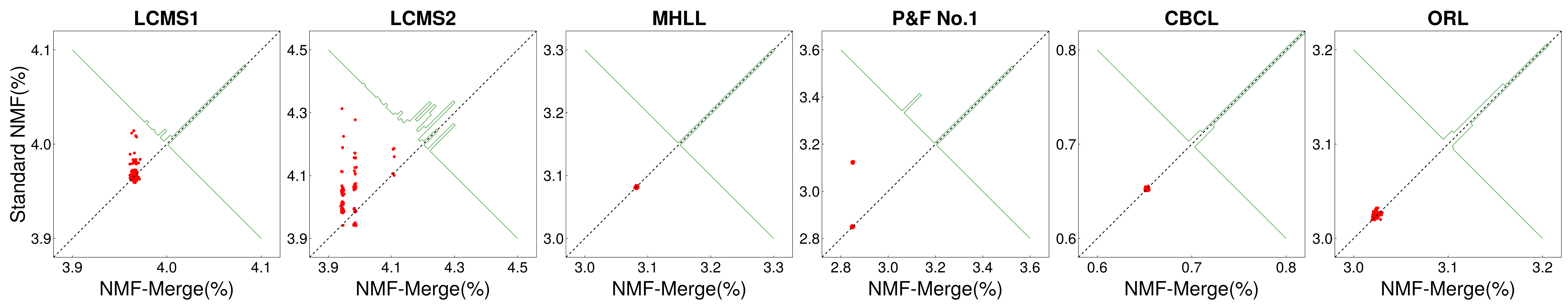}}%
           
            \subfloat[]{\includegraphics[width=6.8in]{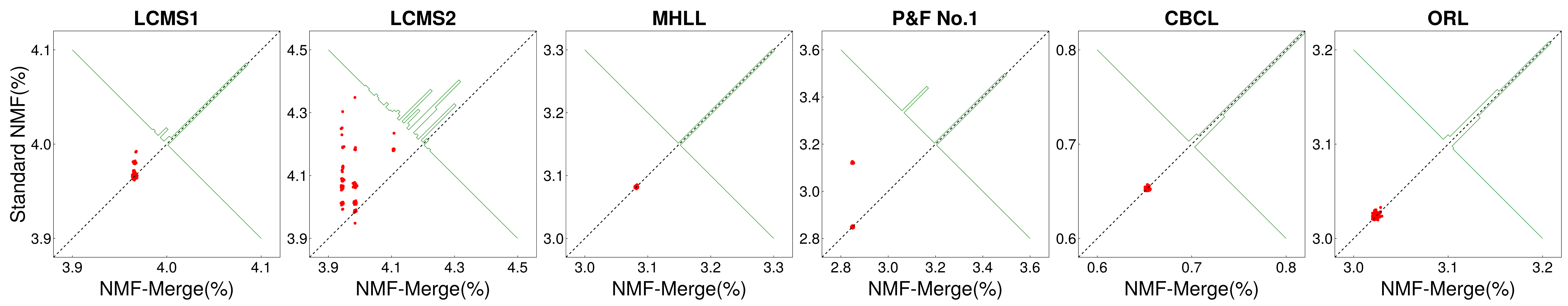}}%
            
            \subfloat[]{\includegraphics[width=6.8in]{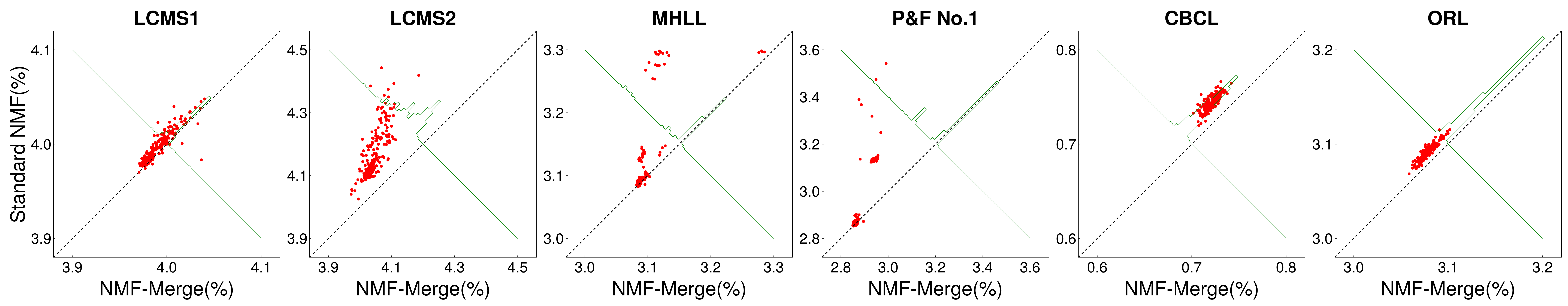}}%
            
            \subfloat[]{\includegraphics[width=6.8in]{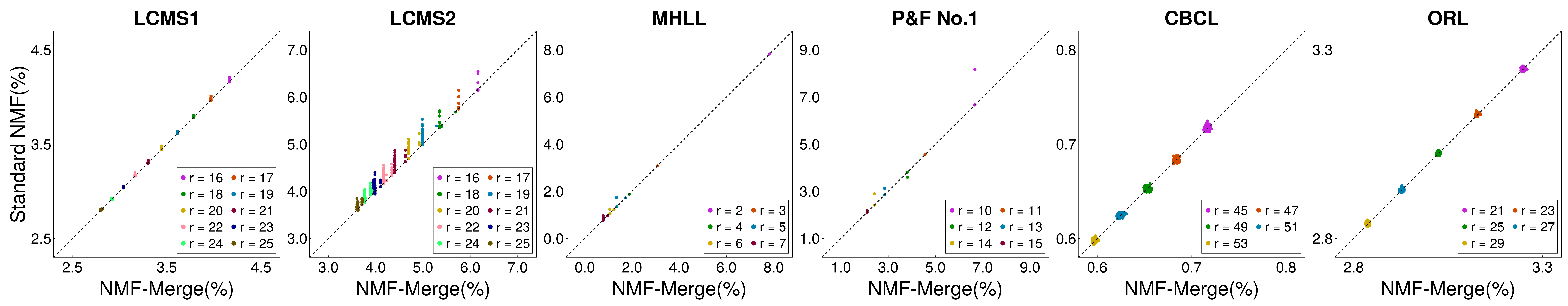}}%
            \caption{The comparison of standard NMF and NMF-Merge: (a) HALS. (b) GCD. (c) ALSPGrad. (d) MU. (e) HALS with multiple number of components.}
            \label{fig_scatter_spec_r_spec_tol}
    \end{figure*}

    \begin{table*}
        \renewcommand{\arraystretch}{1.5}
        \centering
        \caption{Comparison of NMF-Merge and standard NMF using deterministic initialization}
        \label{diff_ini_compare}
        \begin{tabular}{|M{1.3cm}|M{1.5cm}||M{1.9cm}|M{1.9cm}|M{1.9cm}|M{1.9cm}|M{1.9cm}|M{1.9cm}|}\hline
            \multicolumn{2}{|c||}{} & \multicolumn{6}{c|}{Fitting error (\%) : Standard NMF / NMF-Merge} \\\hline
             \multicolumn{2}{|c||}{Data sets} & LCMS1 & LCMS2 & MHLL & P\&F No.1 & CBCL & ORL \\\hline
             \multicolumn{2}{|c||}{$r$} & 17 & 23 & 3 & 13 & 49 & 25 \\\hline
             \multicolumn{2}{|c||}{$k$} & 3 & 5 & 1 & 3 & 10 & 5 \\\hline  \hline
              % Algorithm & Initialization & \multicolumn{6}{c|}{Fitting error (\%) : Standard NMF / NMF-Merge} \\\hline
           \multirow{4}{*}{HALS} & Random & $3.97\pm0.01$ / $3.97\pm0.00$ & $\mathbf{4.06\pm0.08}$ / $\mathbf{3.97\pm0.04}$ & $3.08\pm0.00$ / $3.08\pm0.00$ & $\mathbf{2.91\pm0.11}$ / $\mathbf{2.85\pm0.00}$ & $0.65\pm0.00$ / $0.65\pm0.00$ & $3.02\pm0.00$ / $3.02\pm0.00$ \\ \cline{2-8}
           & NNDSVD & 3.97 / 3.97 & \textbf{4.05 / 3.98} & 3.08 / 3.08 & 2.85 / 2.85 & 0.65 / 0.65 & 3.02 / 3.02 \\ \cline{2-8}
           & NNDSVDa & 3.97 / 3.97 & \textbf{4.07 / 3.98} & 3.08 / 3.08 & 2.85 / 2.85 & 0.65 / 0.65 & 3.02 / 3.02\\ \cline{2-8}
           & NNDSVDar & 3.97 / 3.97 & \textbf{4.05 / 3.98} & 3.08 / 3.08 & 2.85 / 2.85 & 0.65 / 0.66 & 3.02 / 3.02 \\  \hline \hline
           
           \multirow{4}{*}{GCD} & Random & $3.97\pm0.01$ / $3.97\pm0.00$ & $\mathbf{4.03\pm0.07}$ / $\mathbf{3.97\pm0.03}$ & $3.08\pm0.00$ / $3.08\pm0.00$ & $\mathbf{2.90\pm0.11}$ / $\mathbf{2.85\pm0.00}$ & $0.65\pm0.00$ / $0.65\pm0.00$ & $3.02\pm0.00$ / $3.02\pm0.00$ \\ \cline{2-8}
           & NNDSVD & 3.98 / 3.97 & 3.99 / 3.98 & 3.08 / 3.08 & 2.85 / 2.85 & 0.65 / 0.65 & 3.02 / 3.02 \\ \cline{2-8}
           & NNDSVDa & 3.97 / 3.97 & \textbf{4.00 / 3.94} & 3.08 / 3.08 & 2.85 / 2.85 & 0.65 / 0.65 & 3.03 / 3.03\\ \cline{2-8}
           & NNDSVDar & 3.98 / 3.97 & 3.99 / 3.98 & 3.08 / 3.08 & 2.85 / 2.85 & 0.65 / 0.65 & 3.02 / 3.02 \\  \hline \hline
           
           \multirow{4}{*}{ALSPGrad} & Random & $3.97\pm0.01$ / $3.97\pm0.00$ & $\mathbf{4.07\pm0.06}$ / $\mathbf{3.97\pm0.04}$ & $3.08\pm0.00$ / $3.08\pm0.00$ & $\mathbf{2.92\pm0.12}$ / $\mathbf{2.85\pm0.00}$ & $0.65\pm0.00$ / $0.65\pm0.00$ & $3.02\pm0.00$ / $3.02\pm0.00$ \\ \cline{2-8}
           & NNDSVD & 3.97 / 3.97 & 3.99 / 3.98 & 3.08 / 3.08 & 2.85 / 2.85 & 0.65 / 0.65 & 3.02 / 3.03 \\ \cline{2-8}
           & NNDSVDa & 3.97 / 3.97 & \textbf{4.19 / 3.94} & 3.08 / 3.08 & 2.85 / 2.85 & 0.65 / 0.66 & 3.02 / 3.02\\ \cline{2-8}
           & NNDSVDar & 3.97 / 3.97 & 3.99 / 3.98 & 3.08 / 3.08 & 2.85 / 2.85 & 0.65 / 0.65 & 3.02 / 3.03 \\  \hline \hline
           
           \multirow{4}{*}{MU} & Random & $\mathbf{4.00\pm0.02}$ / $\mathbf{3.99\pm0.01}$ & $\mathbf{4.18\pm0.08}$ / $\mathbf{4.05\pm0.03}$ & $\mathbf{3.12\pm0.06}$ / $\mathbf{3.10\pm0.03}$ & $\mathbf{2.92\pm0.13}$ / $\mathbf{2.88\pm0.03}$ & $\mathbf{0.74\pm0.01}$ / $\mathbf{0.72\pm0.00}$ & $\mathbf{3.09\pm0.01}$ / $\mathbf{3.08\pm0.01}$ \\ \cline{2-8}
           & NNDSVD & \textbf{5.09 / 4.86} & \textbf{5.89 / 4.99} & \textbf{3.24 / 3.20} & \textbf{4.5 / 3.89} & \textbf{1.66 / 1.46} & \textbf{4.05 / 3.90} \\ \cline{2-8}
           & NNDSVDa & 4.01 / 4.00 & \textbf{4.43 / 4.16} & 3.09 / 3.09 & 2.85 / 2.85 & \textbf{0.79 / 0.75} & 3.10 / 3.09\\ \cline{2-8}
           & NNDSVDar & 4.01 / 4.00 & 4.08 / 4.07 & 3.09 / 3.09 & 2.85 / 2.85 & \textbf{0.77 / 0.74} & \textbf{3.12 / 3.10}\\  \hline
        \end{tabular}
    \end{table*}
    
    As mentioned in Sec. \ref{The Pipeline of NMF-Merge}, by transiently augmenting the rank and merging extra components down to the desired rank, NMF-Merge provides a mechanism for existing non-ideal NMF solutions to escape from poor local minima and re-converge to better ones.
    This approach should be independent of the particular algorithm chosen to optimize (\ref{SED_obj}).
    To test these claims, we compared NMF-Merge and standard NMF using four algorithms listed above on our six datasets in Table \ref{data_sets}.
    The initial NMF was performed on 200 random initializations (Initial NMF in Fig. \ref{whole_pipe}). 
    The number of extra components for GSVD feature recovery was set to $20\%$ of the initial NMF rank (the impact of the choice of this number will be analyzed in Sec. \ref{The choice of extra components number}). 
    Fig. \ref{fig_scatter_spec_r_spec_tol} provides a comparative illustration between standard NMF performance and that of the NMF-Merge. 
    The vertical axis represents the fitting error of the standard NMF, and 
    the horizontal axis denotes the fitting error from the final NMF stage of NMF-Merge. For each point in the scatter plot, the two employed the same random initialization.
    Due to the convergence to specific local optima, results cluster at certain points in the figure. 
    For better visualization of the number of points in each cluster, in these scatter plots (but not for other analyses) we added small random perturbations ($10^{-3}\times$range of axis) to the fitting error.
    The green line in each subplot represents a histogram of the distribution of distances from each point to the diagonal.
    The mean and standard deviation of each case in Fig. \ref{fig_scatter_spec_r_spec_tol}(a)-(d) are shown in the rows with ``Random'' in Table \ref{diff_ini_compare}.

    Examination of the results presented in Fig.~\ref{fig_scatter_spec_r_spec_tol} and Table \ref{diff_ini_compare} reveals that (\ref{SED_obj}) for LCMS and audio data possesses single or multiple local optima, which are identified by all four algorithms.
    For MU, some of the clusters in Fig.~\ref{fig_scatter_spec_r_spec_tol}(d) are stationary points as they could be further refined by HALS.
    % \tim{This is a bit problematic: it's the *objective* that has multiple minima. That's independent of the algorithm. The question is, which minima are found by each algorithm?}
    % \youdongimportant{For the melody``Mary Had a Little Lamb'', only one local optimum was identified by HALS, GCD, and ALSGrad. In contrast, for ``Prelude and Fugue No. 1 in C Major'', two local optima were discerned by these three algorithms.}
    % \youdongimportant{When applying MU to two audio datasets, it attains similar local optima as the other three algorithms but also converges poorly to several other stationary points. 
    % Testing with $\epsilon=10^{-6}$ did not improve convergence. 
    % Additionally, when HALS initialized with MU solutions, it converged to the local optima identified by the other algorithms.}
    NMF for the face images data also exhibits multiple local optima by four algorithms, albeit the differences are subtle (as the matrices $\mathbf{W}$ and $\mathbf{H}$ are not identical, a point that is discussed further in the following section). 
    Fig.~\ref{fig_scatter_spec_r_spec_tol} indicates that NMF-Merge either matches or improves on the performance of standard NMF in the majority of cases for every algorithm we tested, as evidenced by the preponderance of data points situated above or along the diagonal line. 
    To elaborate further, when a particular initialization leads standard NMF to converge to a favorable local optimum, NMF-Merge exhibits comparable performance. 
    Conversely, if the initialization lead standard NMF to an unfavorable local optimum, NMF-Merge often guides the convergence towards a more favorable solution. Only rarely does NMF-Merge guide optimization towards less favorable minima. 
    
    Due to the overlap of many points at the same optima, the histogram of diagonal-distances (green lines) may be a more reliable indicator of the prevalence of each kind of outcome.  
    When working with HALS, GCD and ALSPGrad, the advantages of NMF-Merge are particularly apparent on the two LCMS datasets and ''Prelude and Fugue No.1 in C major'', and it is approximately comparable to standard NMF on the remaining data sets. When using MU, its advantages are more pronounced across all six datasets.
    
    To compare the results more comprehensively, we also varied the number of
components ($r$) for each dataset (LCMS1 and LCMS2: $r=16, \cdots, 25$, Mary had a little lamb: $r=2, 3, \cdots, 7$, Prelude and Fugue No.1 in C major: $r=10, 11, \cdots, 15$, MIT CBCL face images: $r=45, 47, \cdots, 53$, ORL face images: $r = 21, 23, \cdots, 29$) and the results with HALS are shown in Fig. \ref{fig_scatter_spec_r_spec_tol}(e). Consistent with the results in Fig. \ref{fig_scatter_spec_r_spec_tol}(a)-(d), the proposed method outperforms standard NMF in the majority cases.
   
    So far, we have compared the performance of NMF-Merge and standard NMF with random initialization. 
    We also examined other well-established deterministic initialization methods, such as NNDSVD, NNDSVDa and NNDSVDar with different NMF algorithms.
    The results, documented in Table~\ref{diff_ini_compare}, reveal that the proposed method is consistently competitive with or surpasses the performance of these initializations across all datasets. 
    Most notably, it demonstrates enhanced performance over the standard NMF for four algorithms on the LCMS2 dataset and for MU on six datasets.

    Overall, the proposed method surpasses the performance of standard NMF when commencing from random initializations. 
    Even with high-quality deterministic initialization, the proposed method demonstrates the potential to further elevate the performance of NMF or, at the very least, to maintain the existing quality of results.

    \subsection{NMF-Merge improves the consistency of NMF solutions}
    \label{NMF-Merge mitigates the non-uniqueness of NMF solutions}
    \begin{figure*}[!b]
        \centering
            \subfloat[]{\includegraphics[width=1.18in]{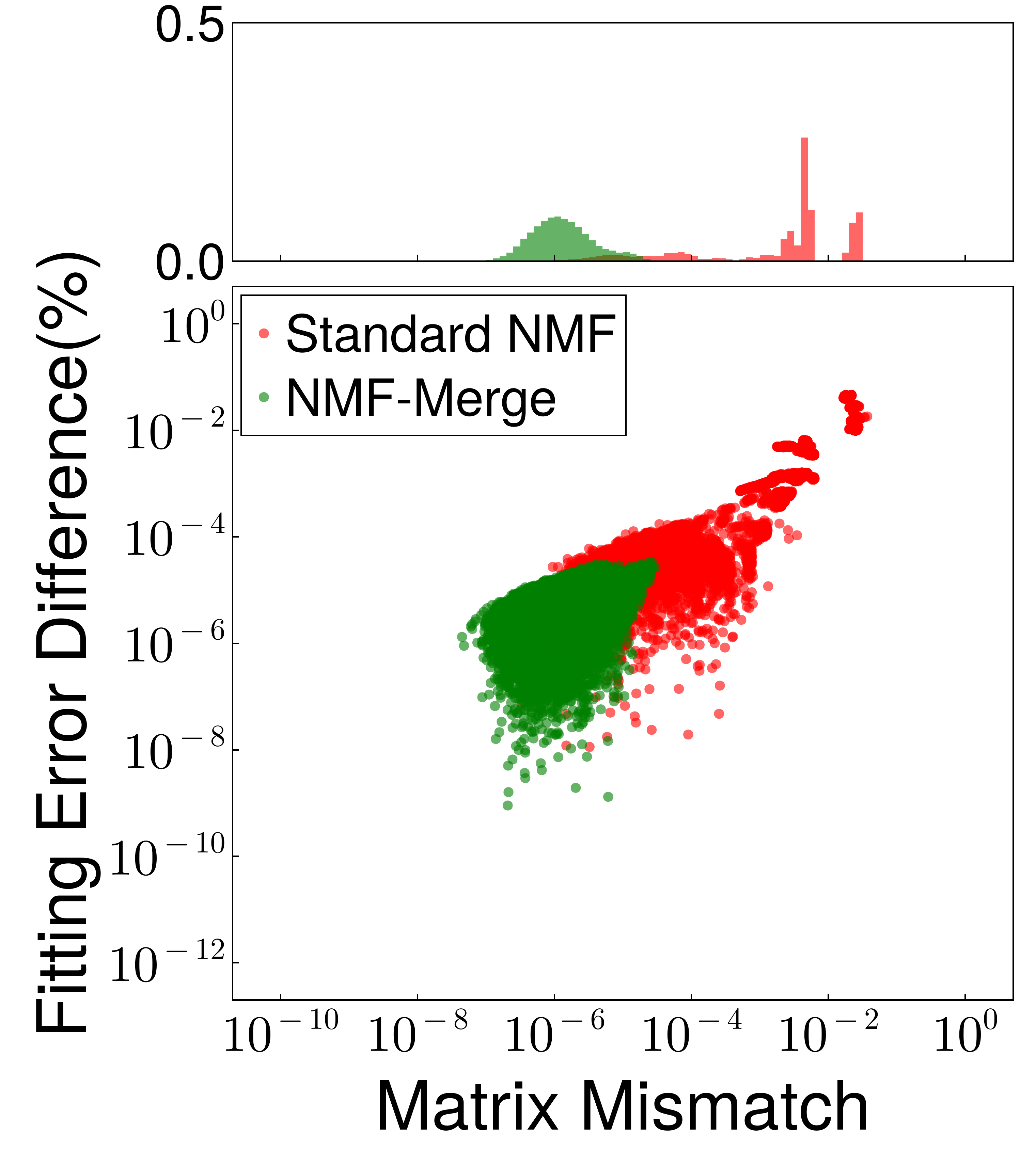}}%
            \subfloat[]{\includegraphics[width=1.18in]{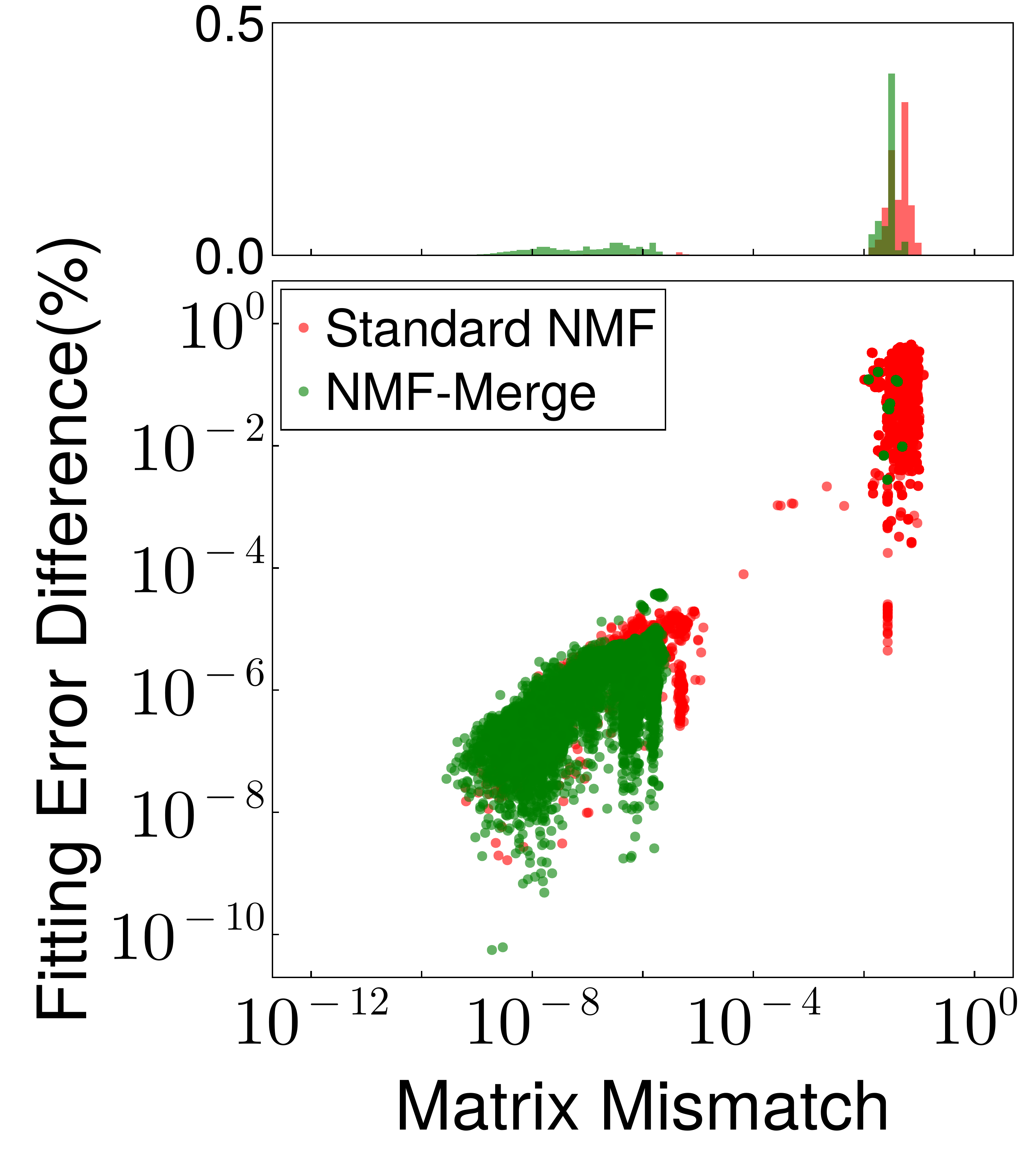}}%
            \subfloat[]{\includegraphics[width=1.18in]{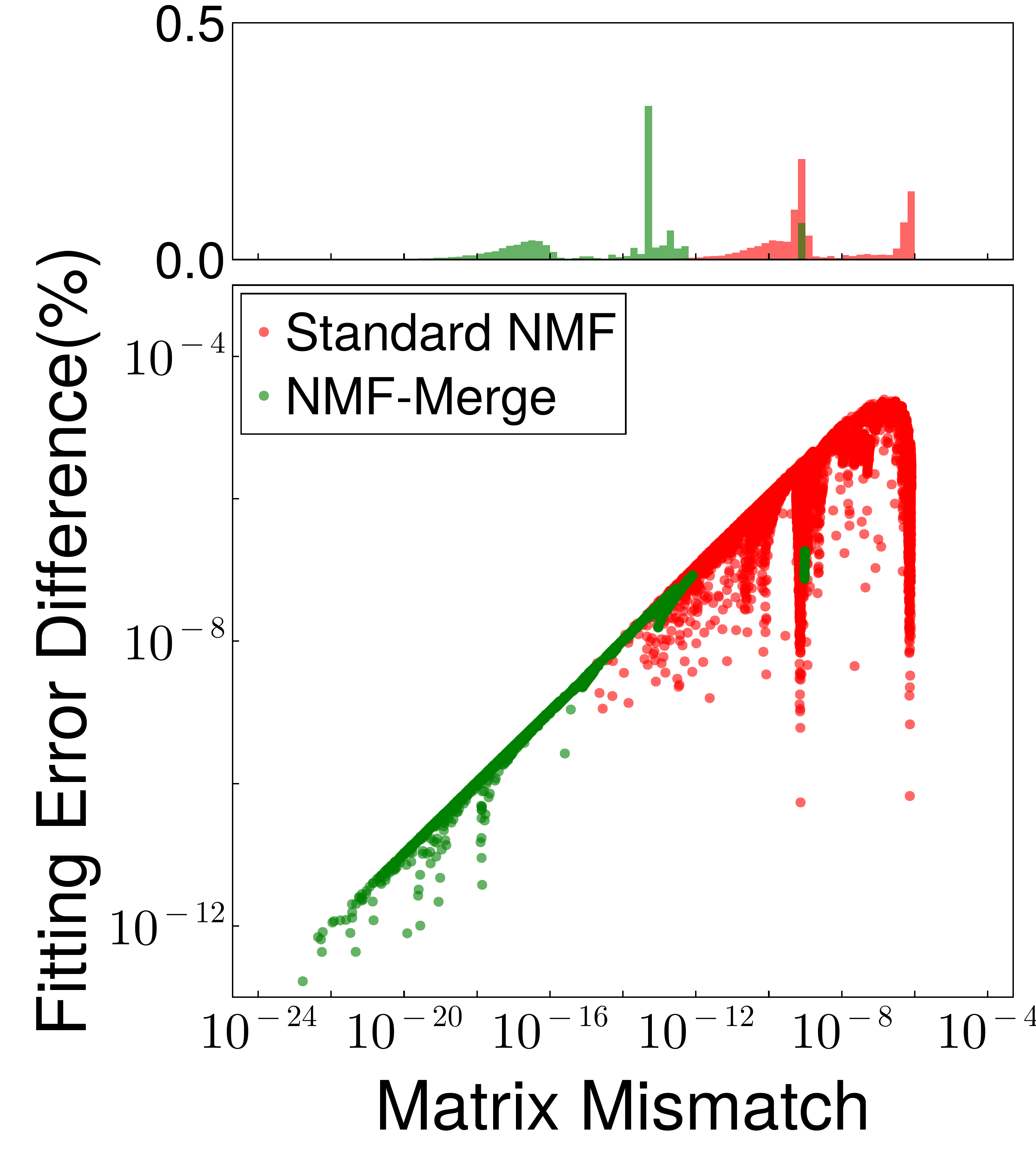}}%
            \subfloat[]{\includegraphics[width=1.18in]{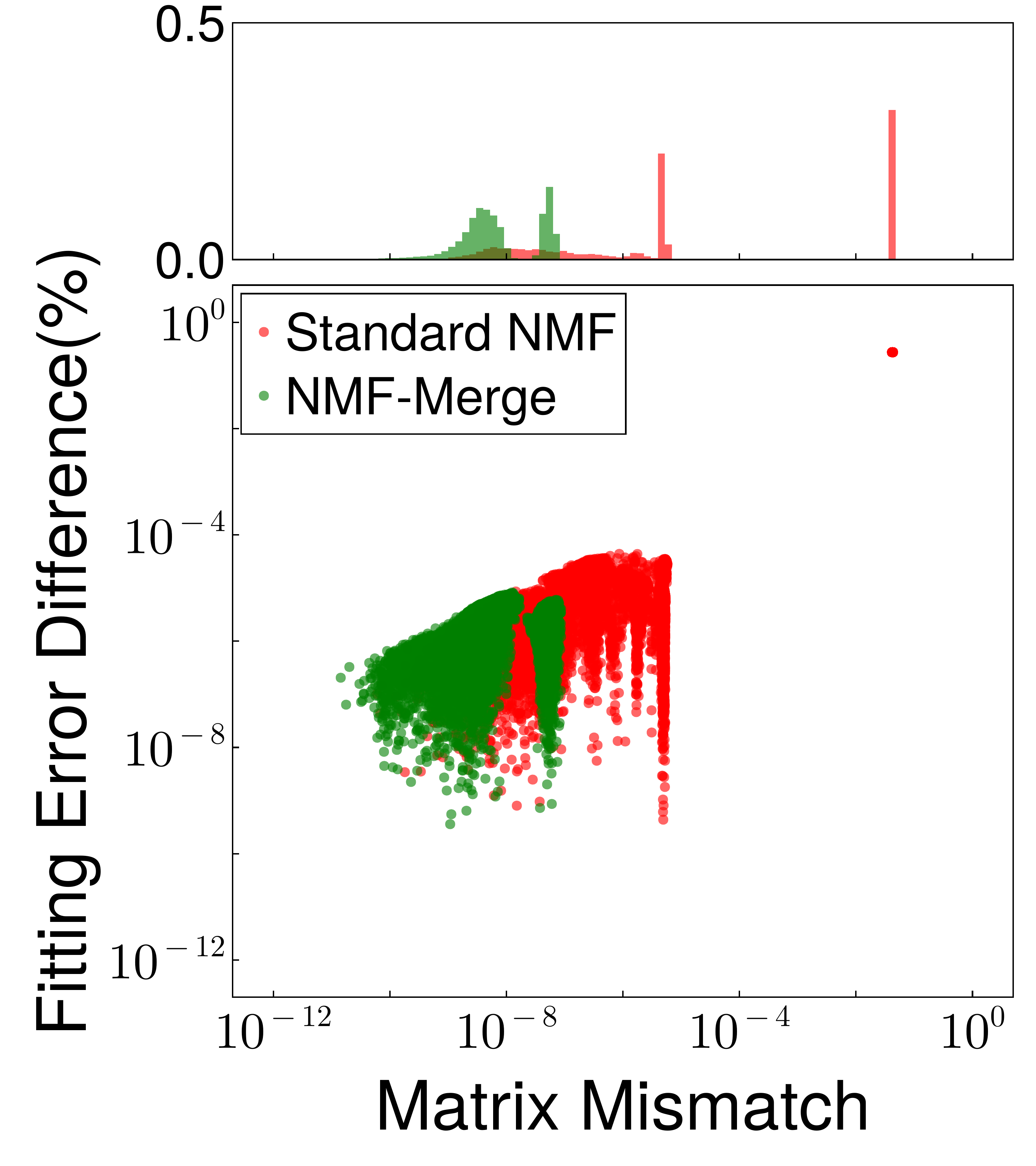}}%
            \subfloat[]{\includegraphics[width=1.18in]{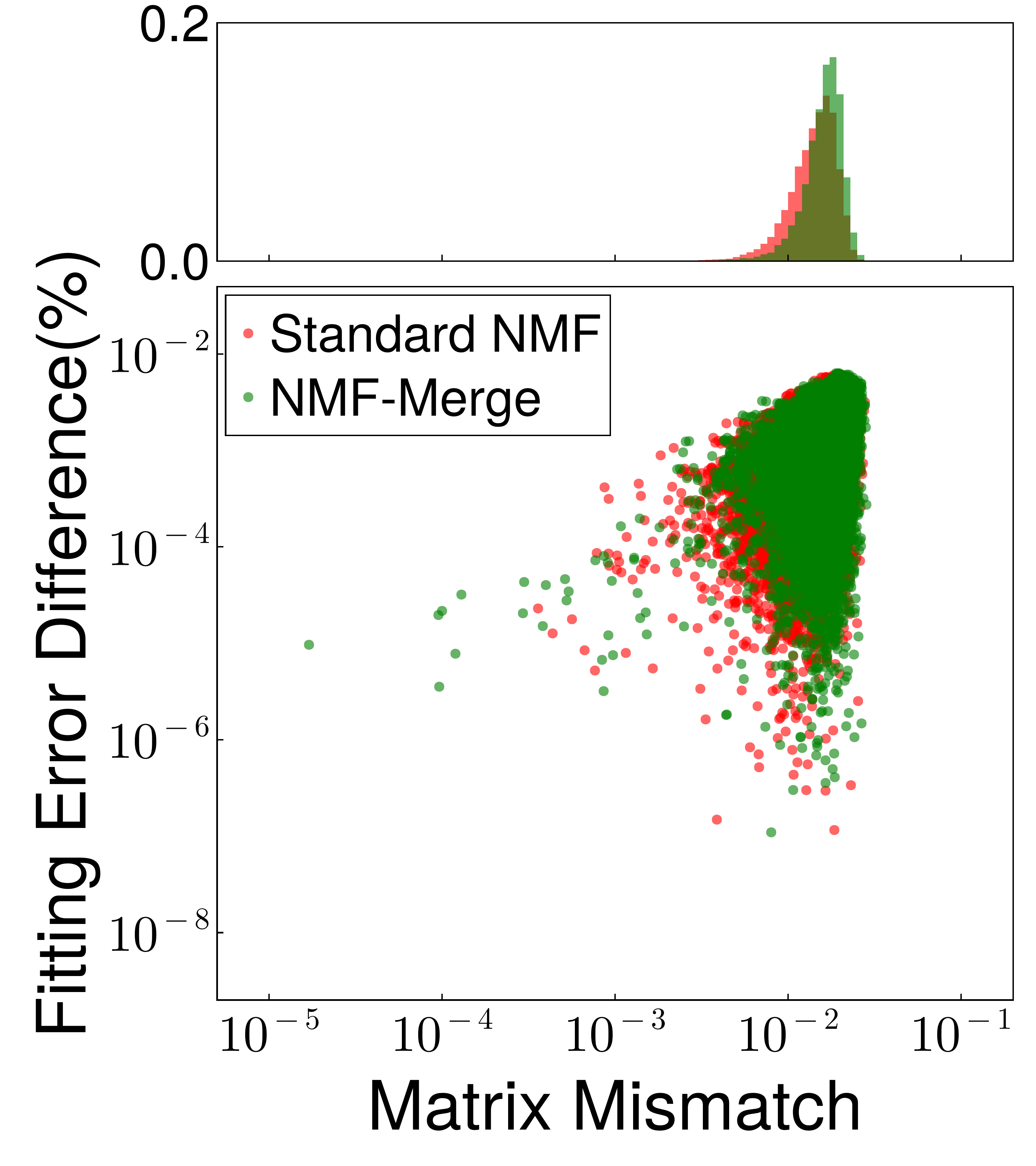}}%
            \subfloat[]{\includegraphics[width=1.18in]{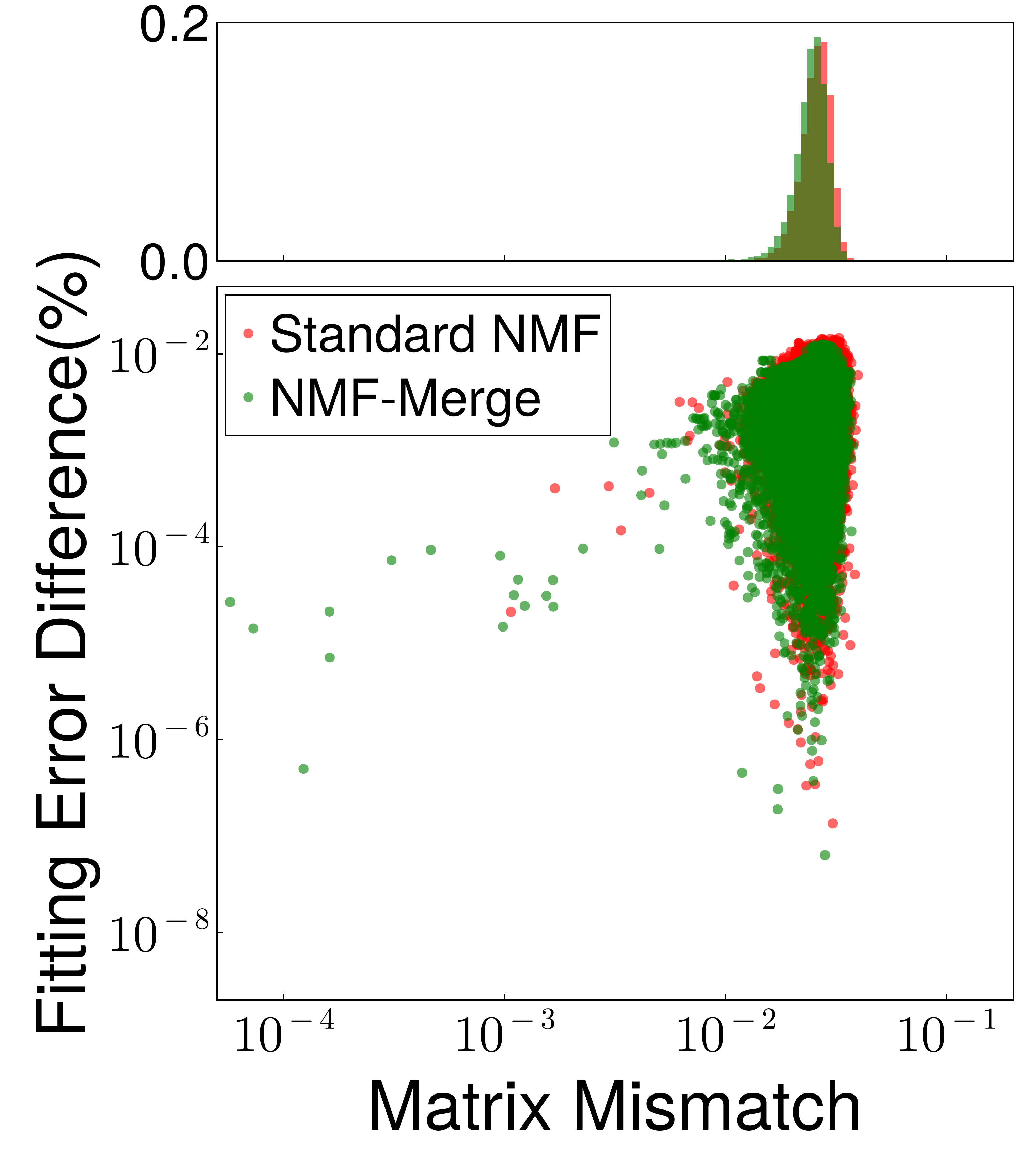}}%
            \caption{Consistency between $\mathbf{W}$ generated from standard and NMF-Merge: (a) LCMS1. (b) LCMS2. (c) Mary had a little lamb. (d) Prelude and Fugue No.1 in C major. (e) CBCL. (f) ORL.}
            \label{consistent_ana}
    \end{figure*}

\begin{figure*}[!b]
        \centering
            \subfloat[]{\includegraphics[width=1.18in]{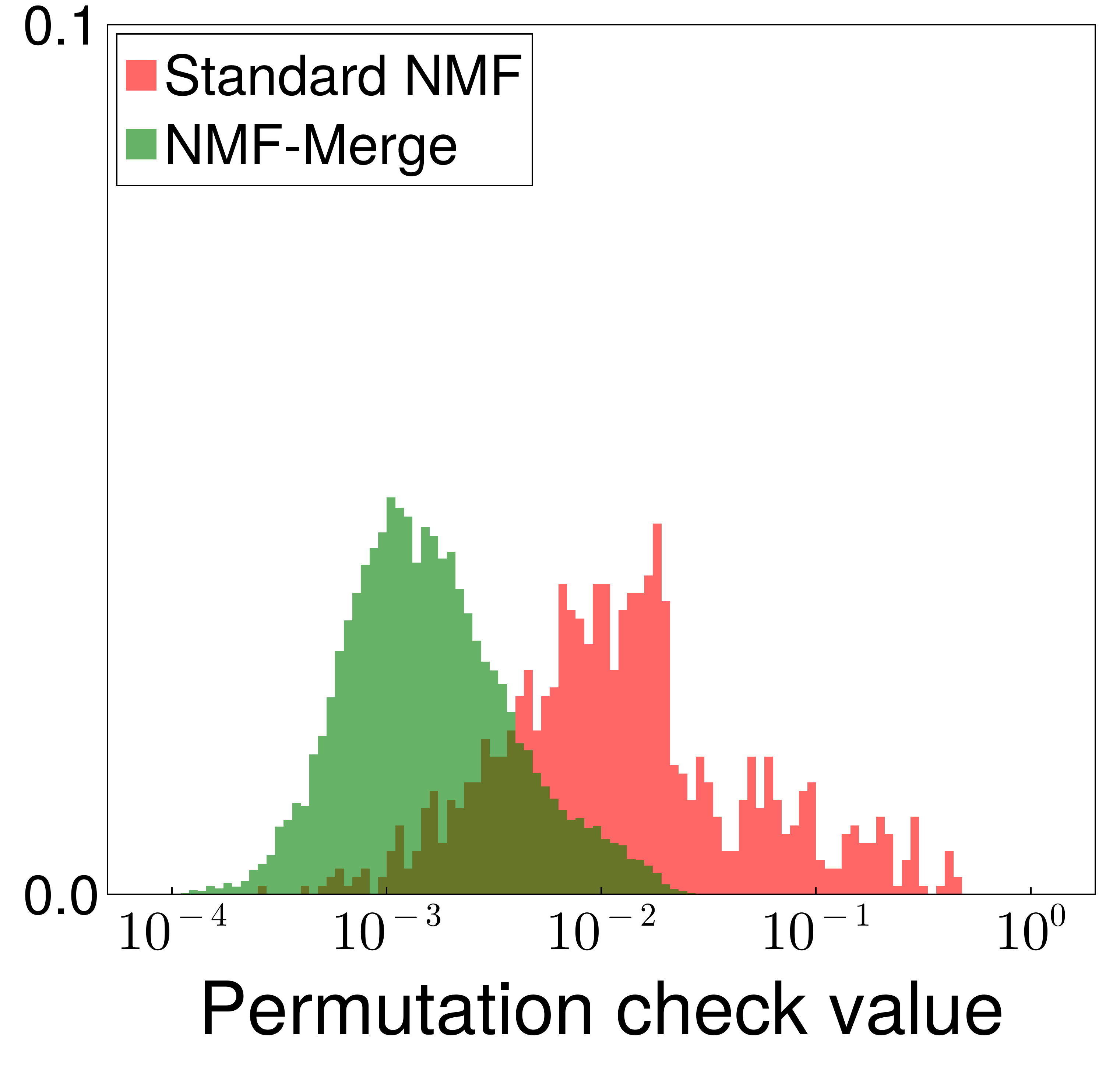}}%
            \subfloat[]{\includegraphics[width=1.18in]{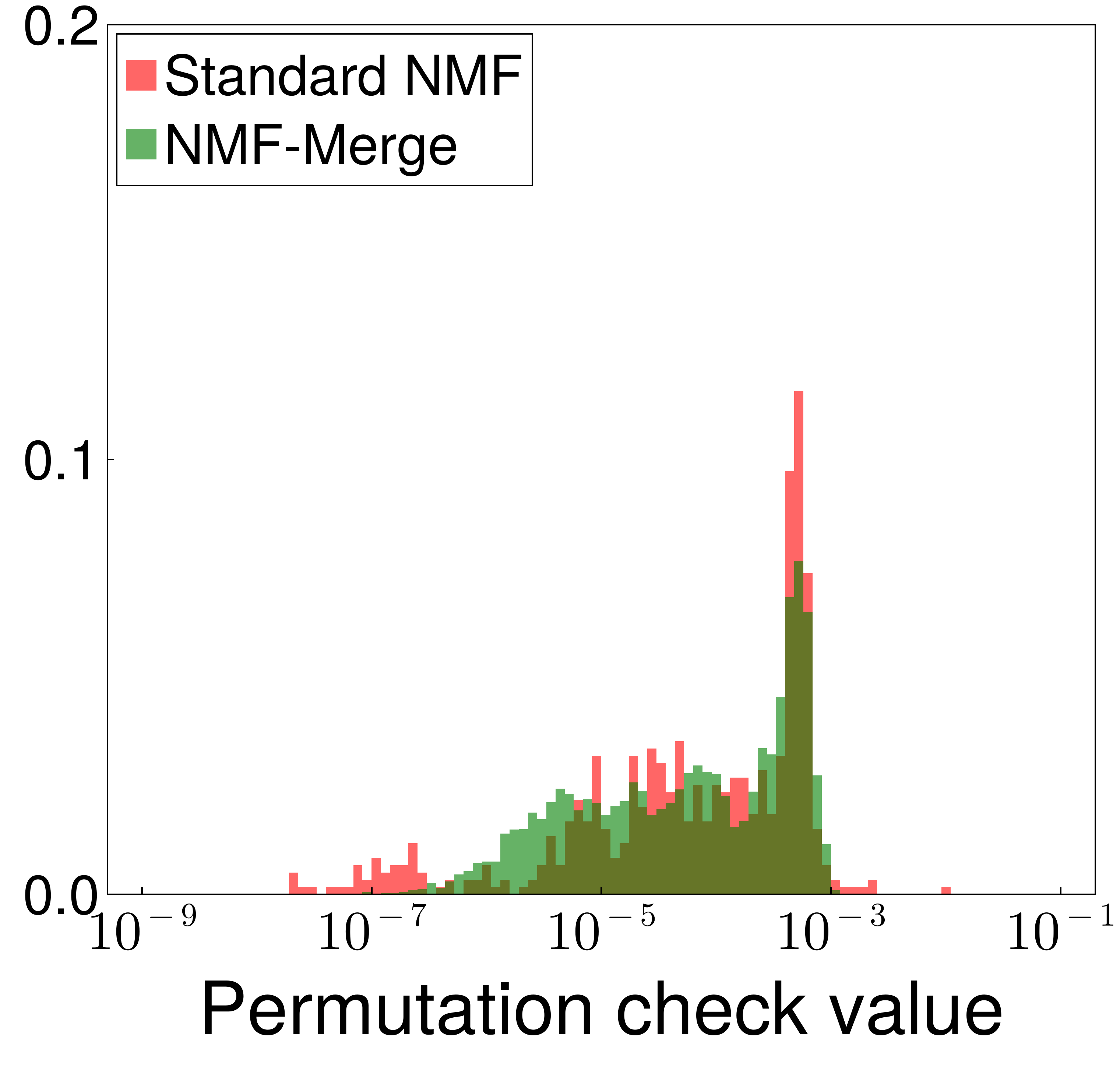}}%
            \subfloat[]{\includegraphics[width=1.18in]{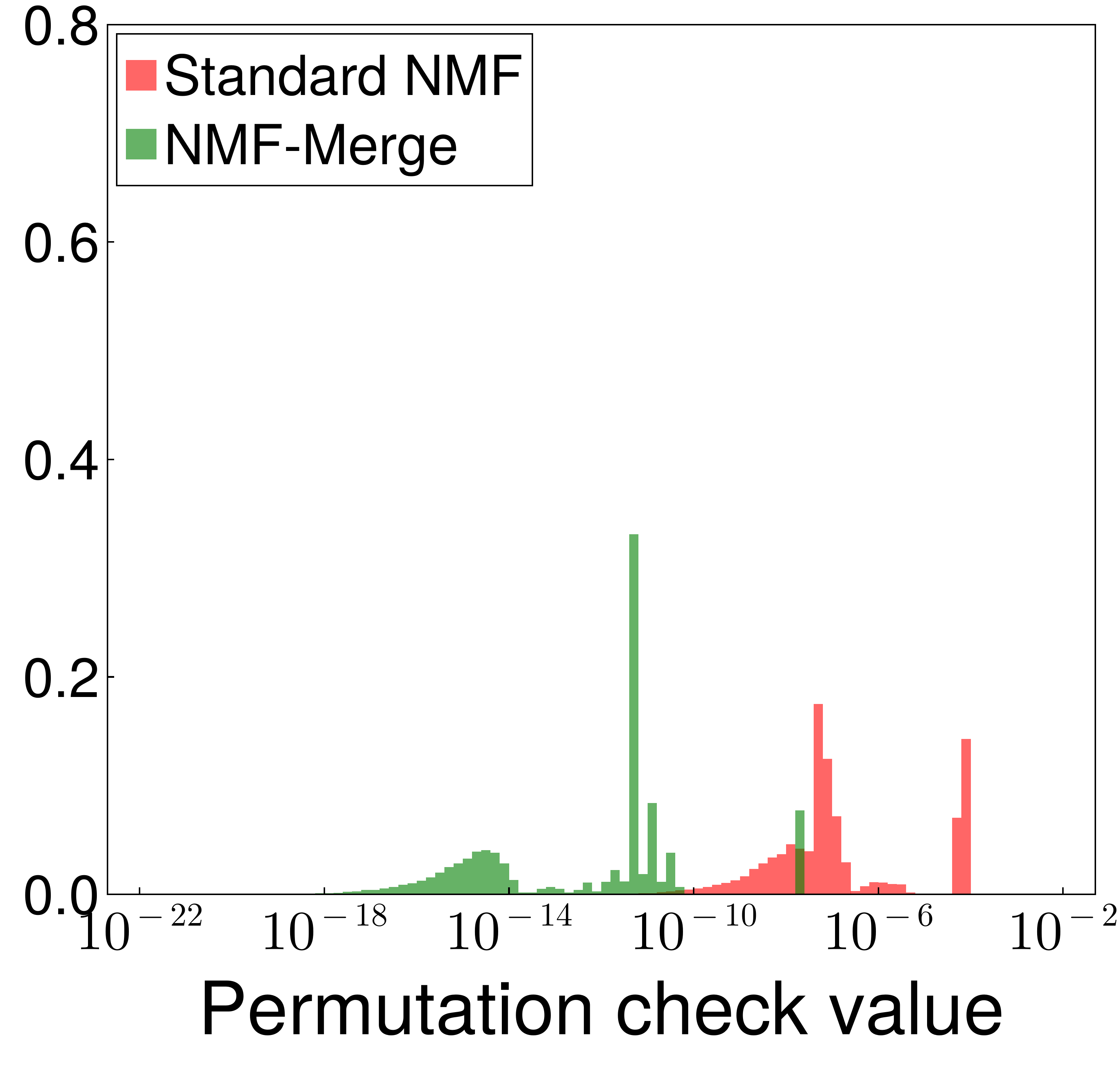}}%
            \subfloat[]{\includegraphics[width=1.18in]{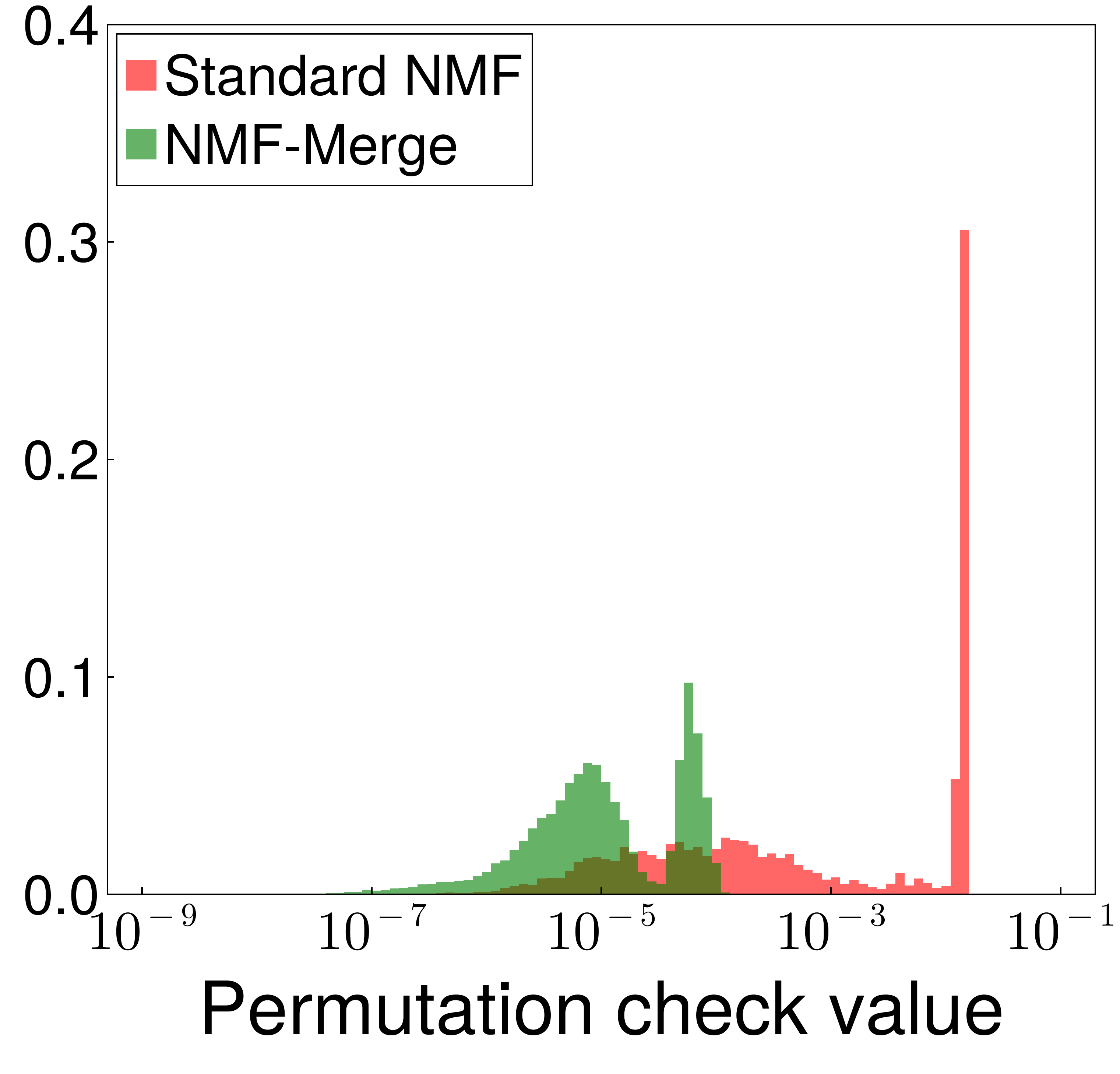}}%
            \subfloat[]{\includegraphics[width=1.18in]{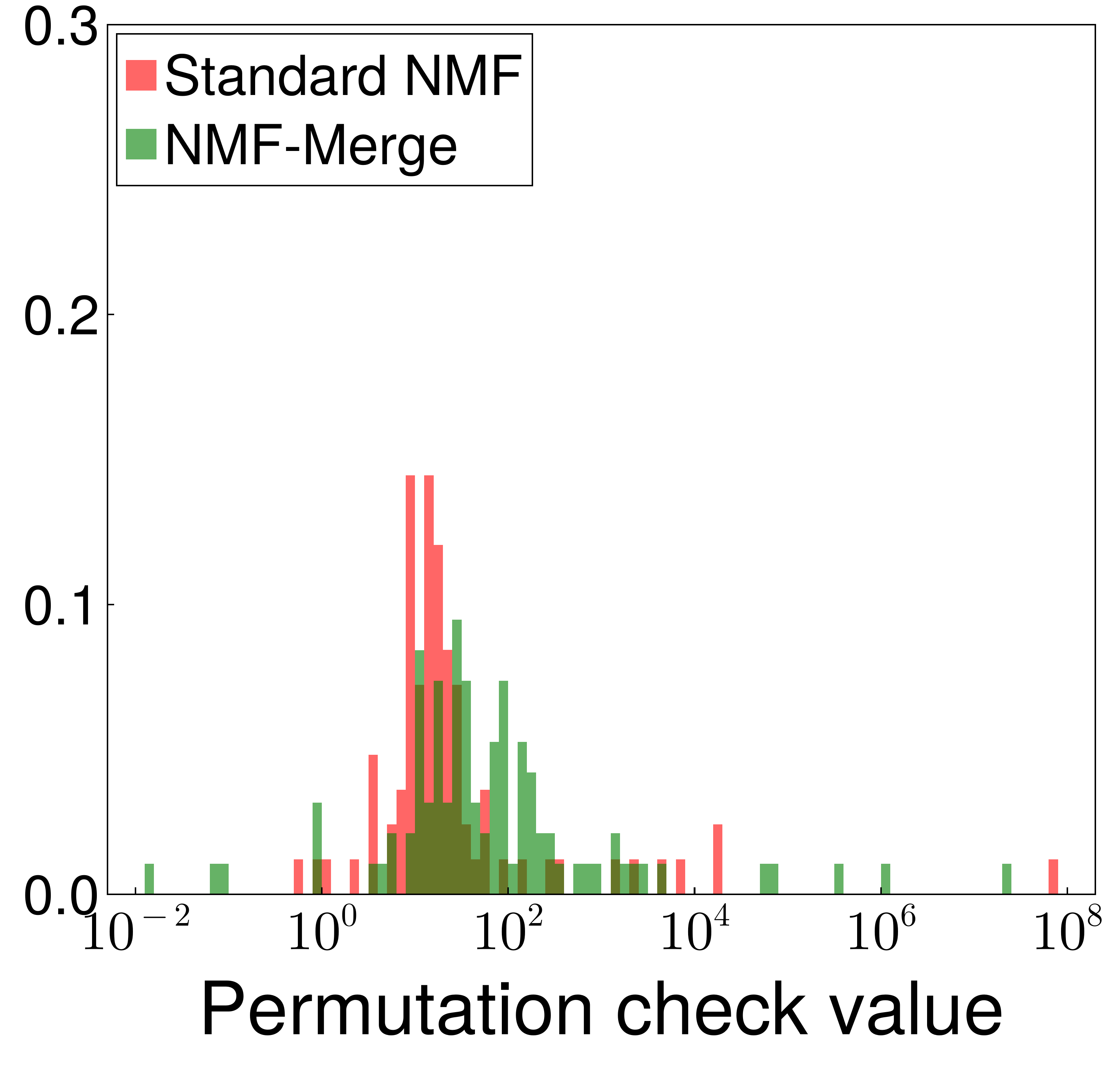}}%
            \subfloat[]{\includegraphics[width=1.18in]{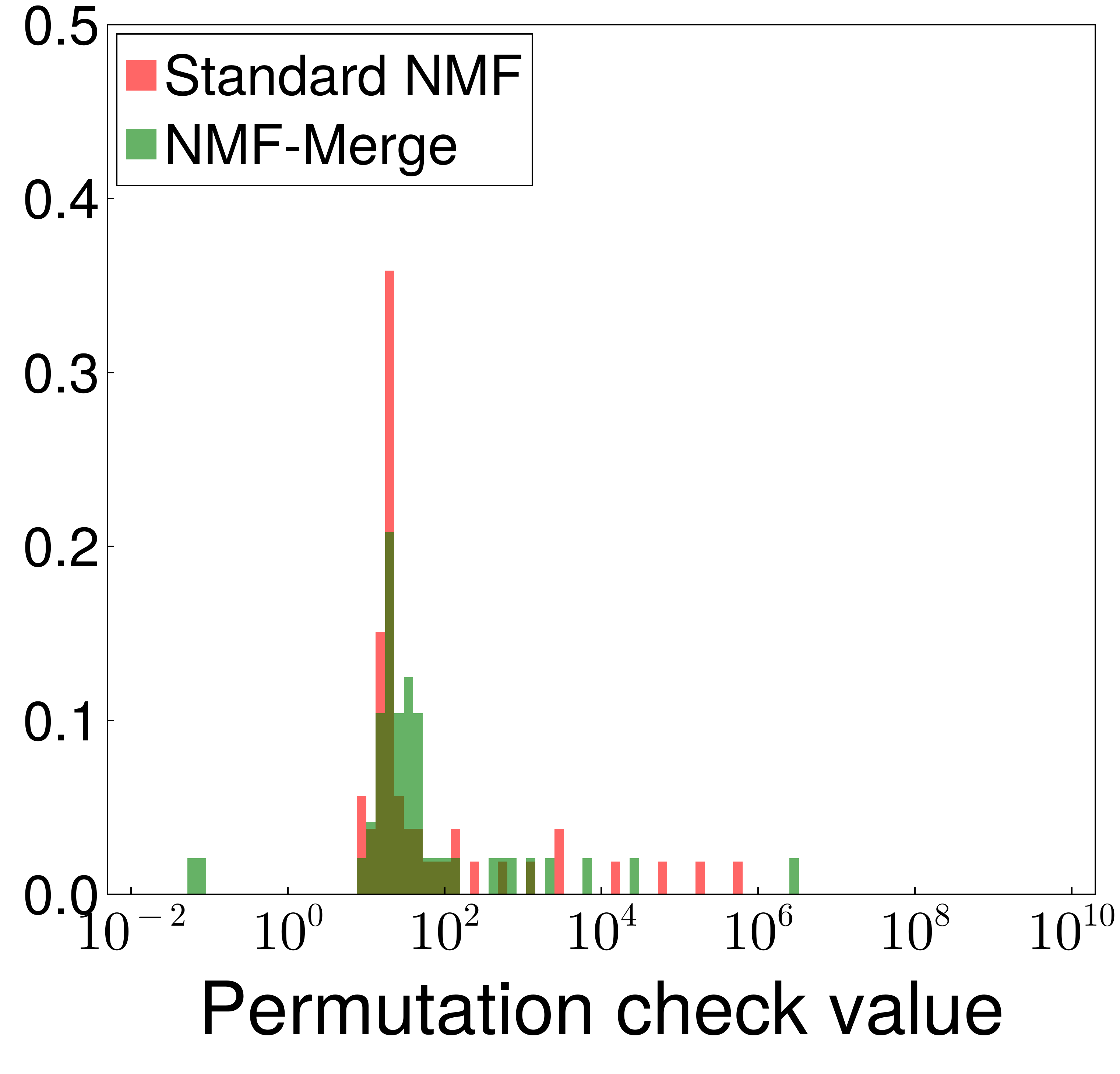}}%
            \caption{The histogram of the distribution of permutation consistency: (a) LCMS1. (b) LCMS2. (c) Mary had a little lamb. (d) Prelude and Fugue No.1 in C major. (e) CBCL. (f) ORL.}
            \label{permutation_check_fig}
    \end{figure*}
    
    The SVD has a global optimum that is typically unique (as long as all singular values are distinct) up to permutation. However, two NMF solutions can differ by more than a permutation and yet have identical objective value \cite{gillis2014nonnegative}. To investigate whether NMF-Merge improves the consistency of NMF solutions, we analyzed the feature parametrizations (components) from multiple runs using two different metrics. We first defined a subspace consistency metric as
    \begin{align}
        \begin{aligned}
        \label{consis_check}
            \mathcal{D}(\mathbf{W}_1, \mathbf{W}_2) =\lVert \mathbf{W}_1-\mathbf{W}_2\mathbf{R}_2 \rVert^2+\lVert \mathbf{W}_2-\mathbf{W}_1\mathbf{R}_1 \rVert^2
        \end{aligned}
    \end{align}
    where $\mathbf{R}_2 = \mathbf{W}_2^{+}\mathbf{W}_1$ and $\mathbf{R}_1 = \mathbf{W}_1^{+}\mathbf{W}_2$, and $\mathbf{W}_1^{+}, \mathbf{W}_2^{+}$ are the pseudoinverse of $\mathbf{W}_1, \mathbf{W}_2$ respectively.
    $\mathcal{D}(\mathbf{W}_1, \mathbf{W}_2)$ is zero only if $\mathbf{W}_1$ and $\mathbf{W}_2$ span the same subspace.
    We nominally restrict our analysis to the $\mathbf{W}$ matrices; because (\ref{SED_obj}) is convex for fixed $\mathbf{W}$, consistency of $\mathbf{W}$ also assures consistency of $\mathbf{H}$. 
    
    Outcomes from the six datasets are shown in Fig.  \ref{consistent_ana}. 
    The upper panel of each subplot is the histogram of the distribution of (\ref{consis_check}) for each pair of factorizations starting from the same random initialization.
    The analysis of the histograms in Fig. \ref{consistent_ana} reveals that NMF is capable of producing variant factorizations, which nonetheless may yield closely comparable fitting errors with respect to the original matrix. 
    NMF-Merge enhances the consistency of these factorizations for four datasets, sometimes substantially so, and yields similar results to standard NMF for the two face datasets.

    A potential caveat of the one-dimensional histograms in Fig.  \ref{consistent_ana} is that they include comparisons among solutions with differing objective values.
    To determine whether solutions with similar objective value (fitting error) can differ, the lower panel of each subplot shows the relationship between the fitting error difference and subspace mismatch (\ref{consis_check}).
    Scatter plots in Fig. \ref{consistent_ana}(a)-(d) demonstrate that on LCMS and audio data sets, NMF-Merge generates more consistent factorizations even when the objective values are nearly indistinguishable. 
    Notably, while the improvement from NMF-Merge in Fig. \ref{fig_scatter_spec_r_spec_tol}(c)-(d) appears subtle at the level of objective value, Fig. \ref{consistent_ana}(c)-(d) shows a marked improvement in the consistency of the components themselves. In all cases, the consistency of NMF-Merge exceeds or equals that of standard NMF.

    The analysis of (\ref{consis_check}) and Fig. \ref{consistent_ana} considers two solutions equivalent if they span the same subspace; however, prior work on the (non)uniqueness of NMF\cite{gillis2014nonnegative} employs a stricter notion, that two solutions are equivalent if they differ only by a permutation. (Here we assume that $\mathbf{W}$ has been normalized so that components are scaled consistently). 
    We therefore propose a second consistency measure (\ref{permutation_check}), permutation consistency, which measures the deviation of $\mathbf{R}_1$ and $\mathbf{R}_2$ from permutation matrices:
    \begin{align}
        \begin{aligned}
            \label{permutation_check}
            PC(\mathbf{R}) =& \sum_{i,j}r_{ij}^2(r_{ij}-1)^2\\
                            & -(\sum_i(\sum_jr_{ij}-1)^2+\sum_j(\sum_ir_{ij}-1)^2)
        \end{aligned}
    \end{align}
    We selected points in Fig. \ref{consistent_ana} that had a fitting error mismatch of less than $10^{-5}$. 
    The distribution of the resulting values from (\ref{permutation_check}) is presented as histograms in Fig. \ref{permutation_check_fig}.
    It demonstrates that for four datasets, the $\mathbf{R}_1$ and $\mathbf{R}_2$ are more similar to permutation matrices for NMF-Merge than standard NMF, whereas the two are approximately equivalent on the two face datasets. These two data sets have more local optima with substantially-different factorizations, and thus the permutation check values are large. 

    \begin{figure*}[!b]
        \centering
            \subfloat[]{\includegraphics[width=3in]{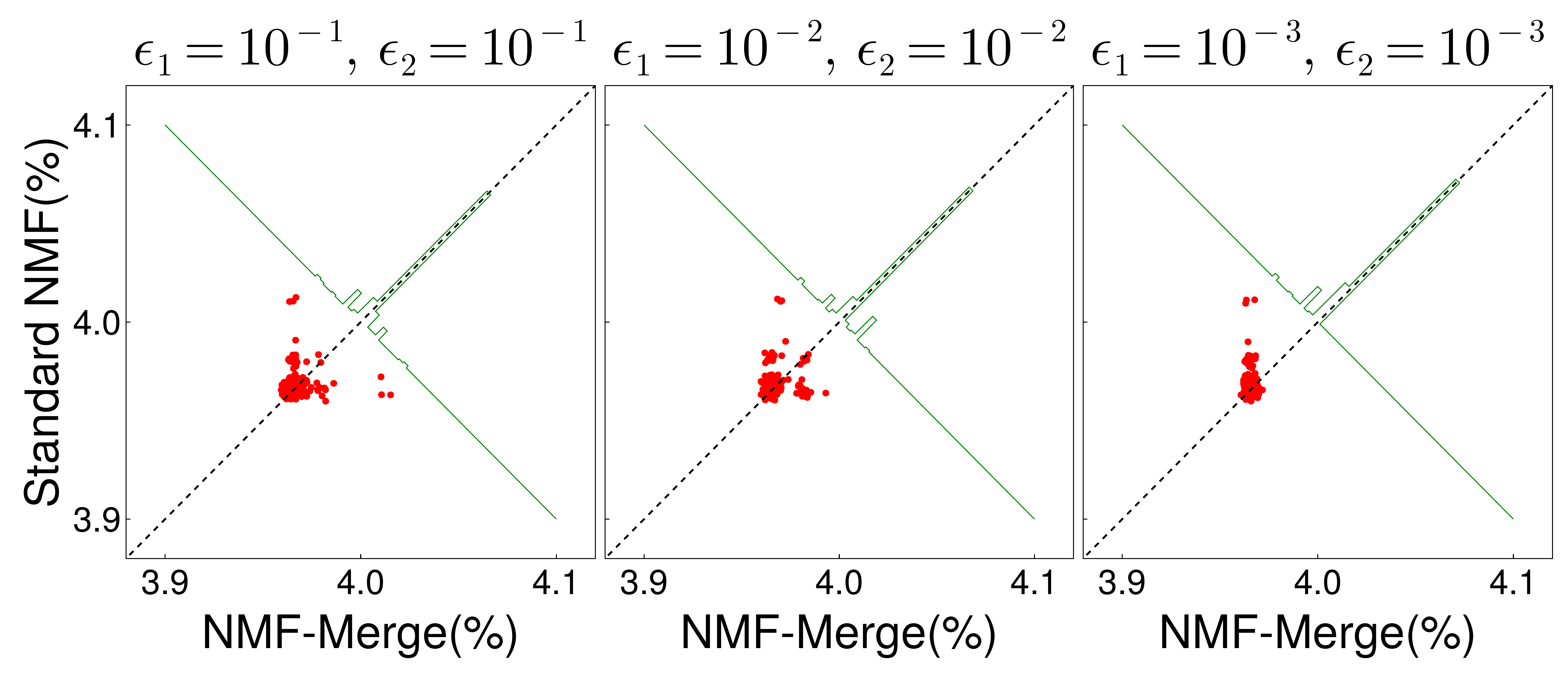}}%
            \hfil
            \subfloat[]{\includegraphics[width=3in]{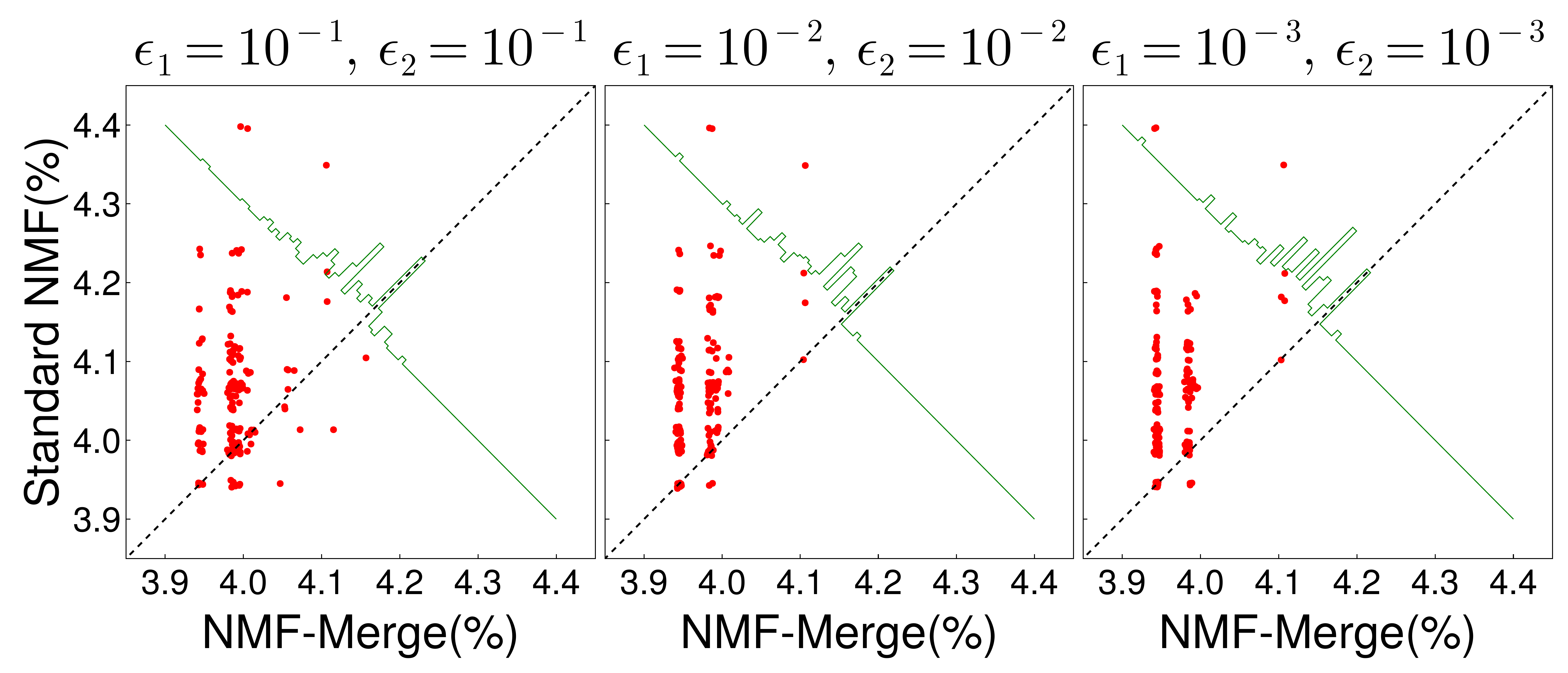}}%
            
            \subfloat[]{\includegraphics[width=3in]{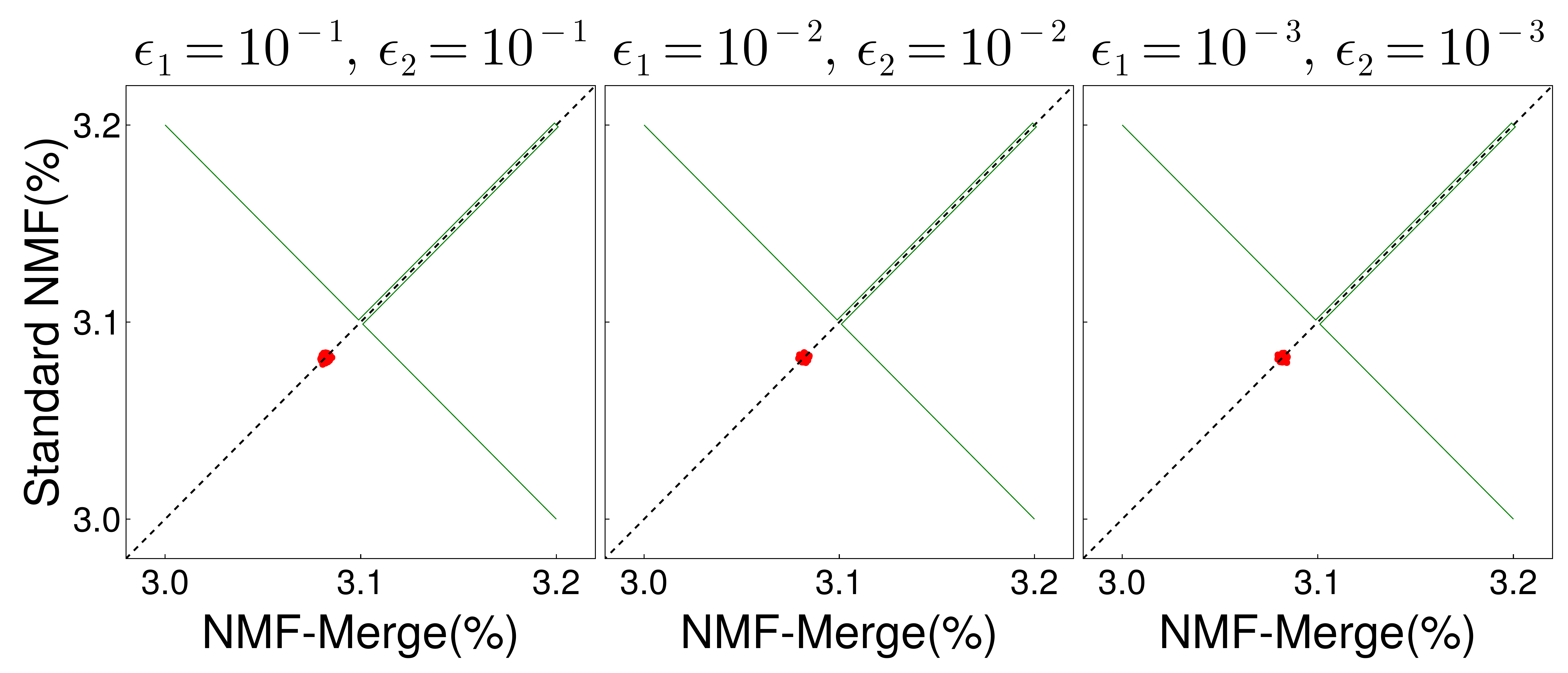}}%
            \hfil
            \subfloat[]{\includegraphics[width=3in]{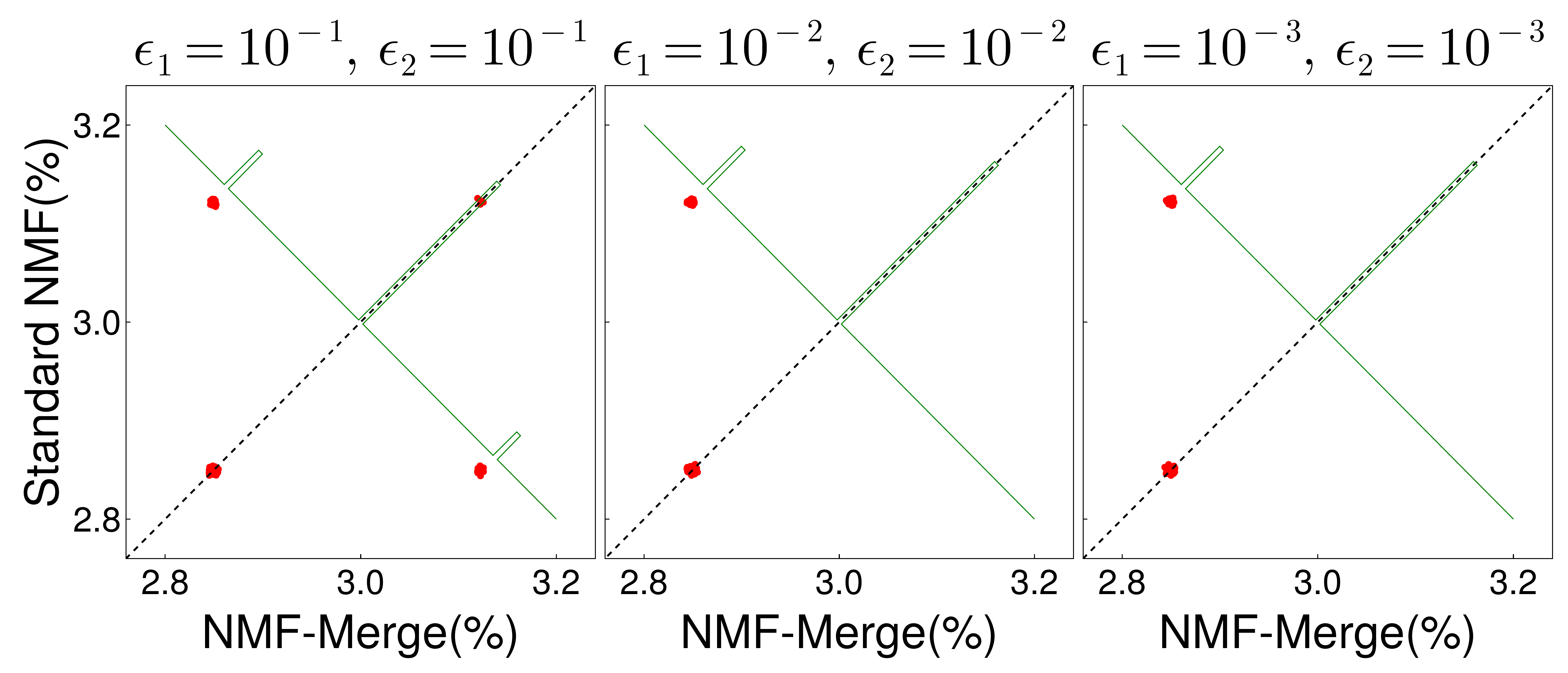}}%
    
            \subfloat[]{\includegraphics[width=3in]{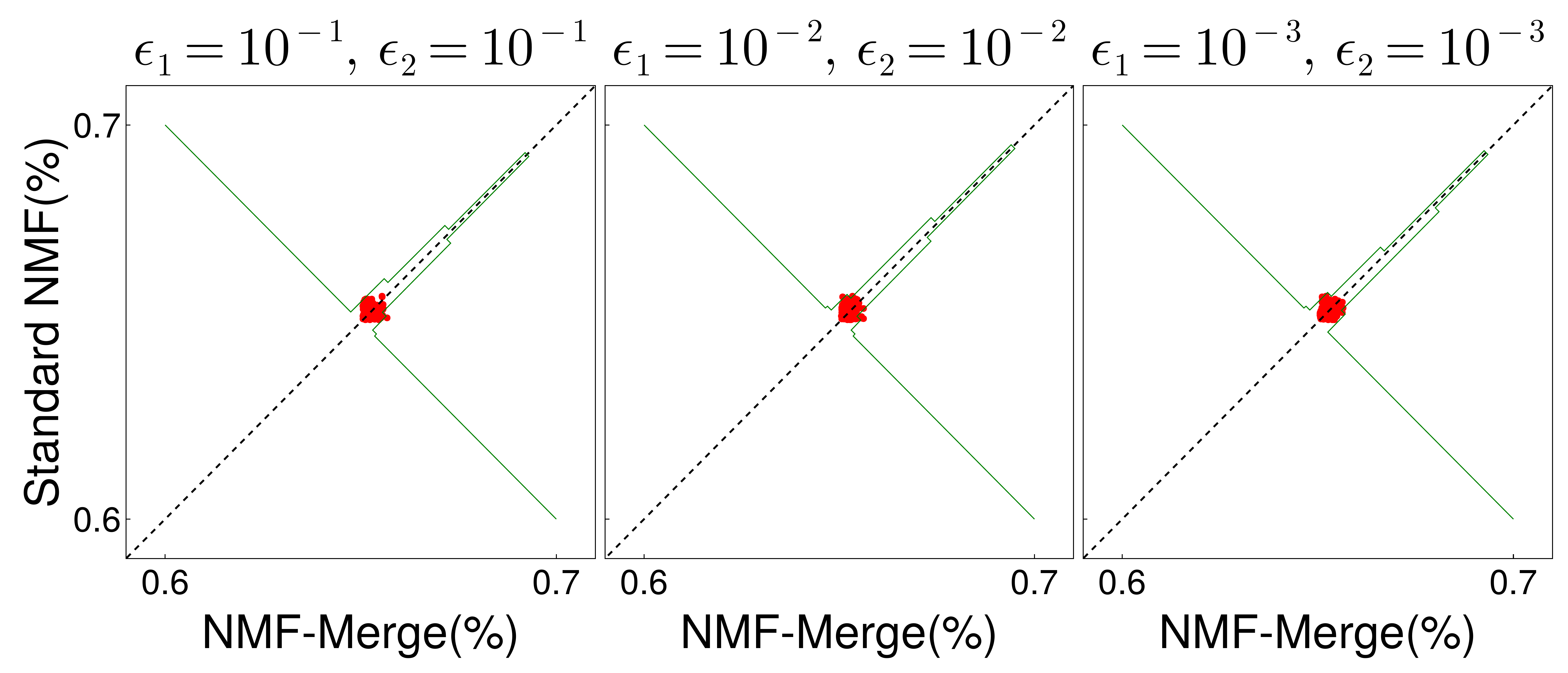}}%
            \hfil
            \subfloat[]{\includegraphics[width=3in]{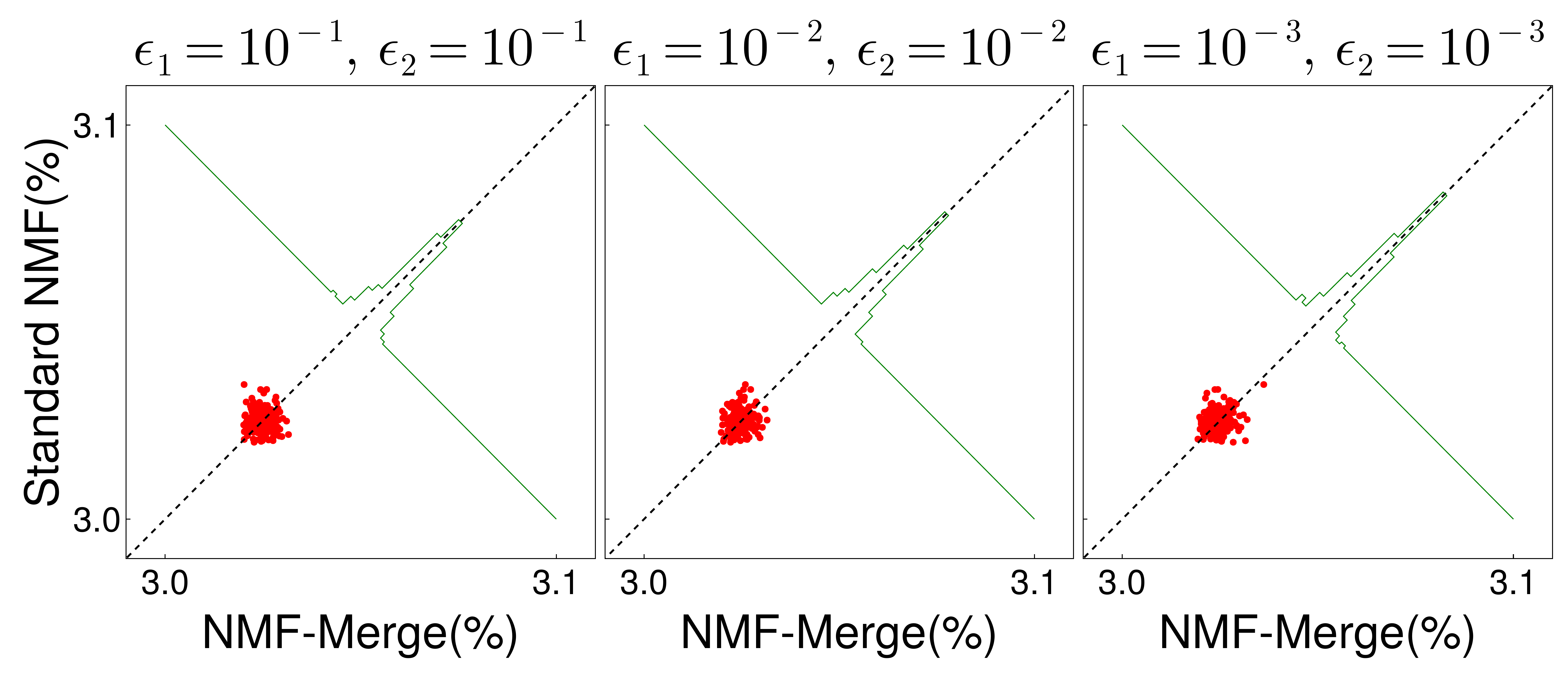}}%
    
            \caption{The effect of convergence tolerance $\epsilon_1$ and $\epsilon_2$ on final results of NMF-Merge. Panels should be compared to those of Fig.  \ref{fig_scatter_spec_r_spec_tol}, which used $\epsilon_1 = \epsilon_2 = 10^{-4}$. (a) LCMS1. (b) LCMS2. (c) Mary had a little lamb. (d) Prelude and Fugue No.1 in C major. (e) CBCL. (f) ORL.}
            \label{fig_scatters_spec_r_all_tol}
    \end{figure*}

    In aggregate, both measures show that NMF-Merge equals or improves the consistency of standard NMF. For applications of NMF that focus on the components more than the overall reconstruction of the original data matrix, the increased reliability of NMF-Merge may be a major asset.
    
    \subsection{The choice of $\epsilon_1$ and $\epsilon_2$}
    
    NMF-Merge runs standard NMF up to three times: the initial NMF, the over-complete NMF, and the final NMF (Fig. \ref{whole_pipe}). So far, all our results have employed the same $\epsilon = 10^{-4}$ in (\ref{hals_stop_condition}) for all three NMF runs. One might wonder whether all of these need to be run to convergence, or whether for computational efficiency one might be able to terminate early for all but the final NMF.
    
    The effects of reducing the stringency of the initial ($\epsilon_1$) and over-complete ($\epsilon_2$) NMF steps are shown in Fig.  \ref{fig_scatters_spec_r_all_tol}. 
    The tolerance has little effect on the two face image and ``Mary had a little lamb'' datasets. 
    On both LCMS and the ``Prelude and Fugue No.1 in C major'' datasets, if the initial and over-complete NMF use very high tolerance ($10^{-1}$), a minority of the data points are positioned below the diagonal; however, at $10^{-2}$ and especially $10^{-3}$ the results were roughly equivalent to our previous results using a tolerance of $10^{-4}$ throughout.

    These results demonstrate that in NMF-Merge, the first two stages of NMF can be run at higher tolerance (lower stringency).
    For an optimal trade-off between computational efficiency and the quality of the final outcome, in practical applications a tolerance threshold at or below  $10^{-2}$ is recommended for both the initial and over-complete NMF. 
  
    \subsection{NMF-Merge does not incur extra computational cost}
    \label{NMF-Merge avoids stalling in standard NMF}
    Given the extra steps depicted in Fig. \ref{whole_pipe}, it is natural to imagine that NMF-Merge might incur a cost in computational performance. 
    Here we show that NMF-Merge performs similarly to---and sometimes better than---standard NMF, largely by avoiding the plateau phenomenon shown in Appendix~\ref{Illustration of plateau phenomenon in NMF} and Fig. \ref{stalling_example}.
    
    To substantiate this claim, we track the temporal evolution of the objective function value. Specifically, starting from the same initialization we perform NMF twice: once to measure the average time per iteration (which varies with dataset and number of retained components), and a second time to measure the objective value after each iteration until convergence. Measuring these two quantities separately was necessary because computation of the objective value takes much longer than a single iteration of NMF. 
    For both standard NMF and the final NMF stage of NMF-Merge, the tolerance was set to $10^{-4}$. 
    The tolerances for the initial and over-complete NMF stages within the NMF-Merge framework were $\epsilon_1 = \epsilon_2 = 10^{-2}$.
    
\begin{figure}[!t]
        \centering
        \subfloat[]{\includegraphics[width=1.16in]{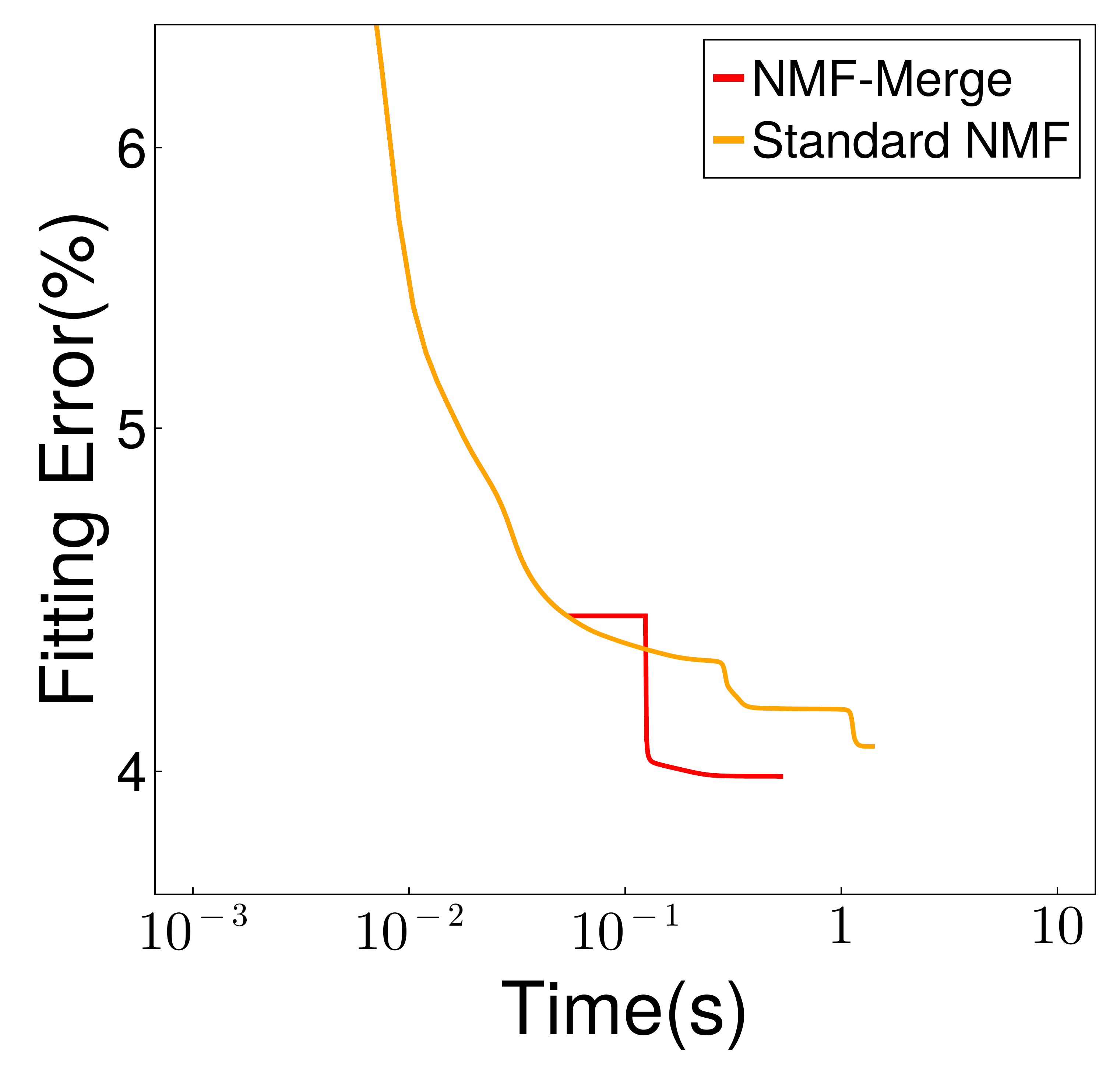}}%
        \subfloat[]{\includegraphics[width=1.16in]{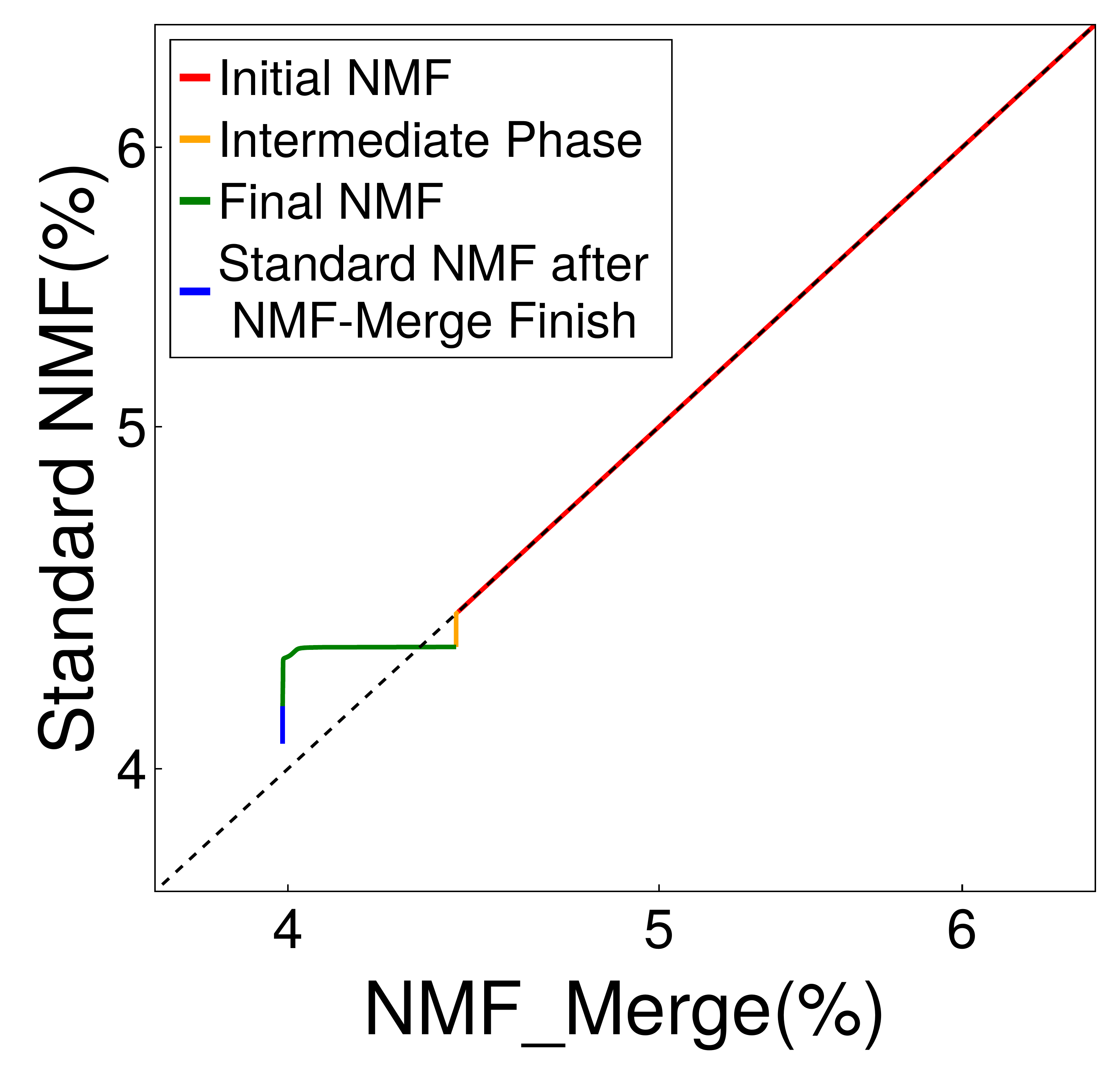}}%
        \subfloat[]{\includegraphics[width=1.16in]{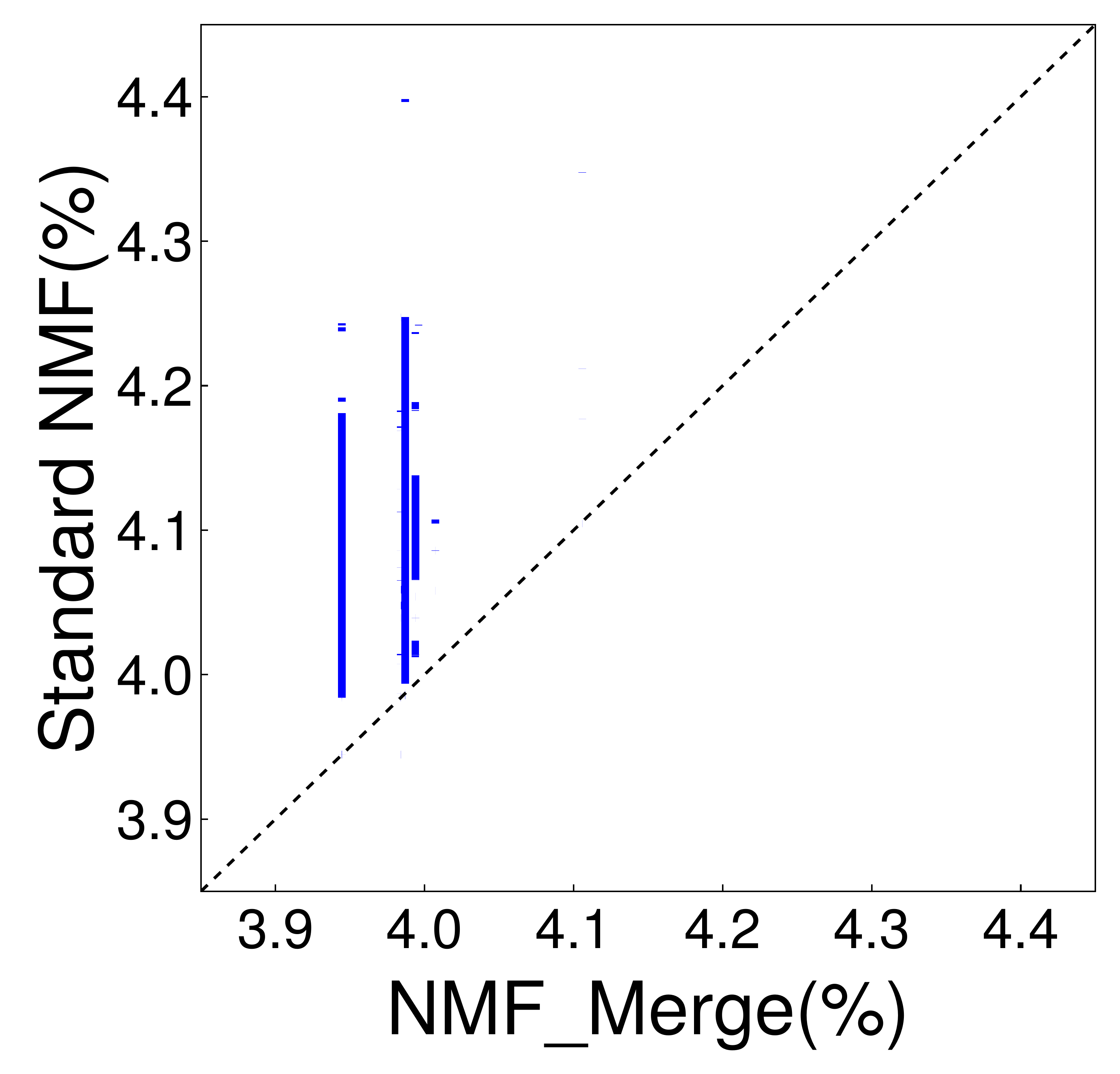}}%
        \caption{Computational performance of NMF-Merge on LCMS 2: (a) The fitting error versus execution time of NMF-Merge and Standard NMF for one specific initialization. (b) The same data as in (a) represented as a phase plot. (c) Phase plots as in (b) for 200 different random initializations; only the portion after one algorithm had already finished (the blue portion of the curve in (b)) is shown.}
        \label{fig_trace_t_example}
    \end{figure}

    Fig. \ref{fig_trace_t_example} introduces our evaluation strategy, using random initialization on LCMS 2. 
    Fig.  \ref{fig_trace_t_example}(a) shows a single favorable (for NMF-Merge) result chosen among the ensemble of 200 different random initializations. 
    The orange curve represents the relative objective function value vs.\ time for standard NMF, while the red curve documents the progression of NMF-Merge. 
    Since the initial phase of NMF-Merge is identical to standard NMF, they share the same fitting error trajectory for the first $\approx 0.05$s.
    NMF-Merge then commences with component augmentation, over-complete NMF, and merge, during which the two solutions are not comparable and hence the objective value is depicted as a constant horizontal line. During this time, standard NMF continues its iterations, and this is depicted in Fig.  \ref{fig_trace_t_example}(a) as a transient advantage for standard NMF. However, during this period standard NMF begins to stall, and hence progress is slow. Once NMF-Merge completes the merge phase, we can re-evaluate the objective value (the vertical jump in the red curve occurring at $\approx 0.2$s), which in this case results in an immediate improvement over standard NMF, and NMF-Merge maintains this advantage throughout convergence of both methods. With other initializations, the post-merge objective value is initially worse but rapidly improves (often surpassing standard NMF) during polishing in the final NMF.
        
    Fig.  \ref{fig_trace_t_example}(b) shows these same data as a phase plot, representing the trajectory of the paired objective values at each moment in time.
    The red, orange and green portions of the curve represent stages of NMF-Merge, the initial NMF, intermediate phase (component augmentation, over-complete NMF, and merge), and final NMF, respectively.
    The blue part denotes the change of objective value of standard NMF during the time after NMF-Merge had already converged.
    Fig.  \ref{fig_trace_t_example}(a)(b) illustrate (for one specific example) that NMF-Merge avoids the plateau phenomenon exhibited by standard NMF, resulting in faster convergence.

    \begin{figure*}[!t]
        \centering
        \subfloat[]{\includegraphics[width=3.2in]{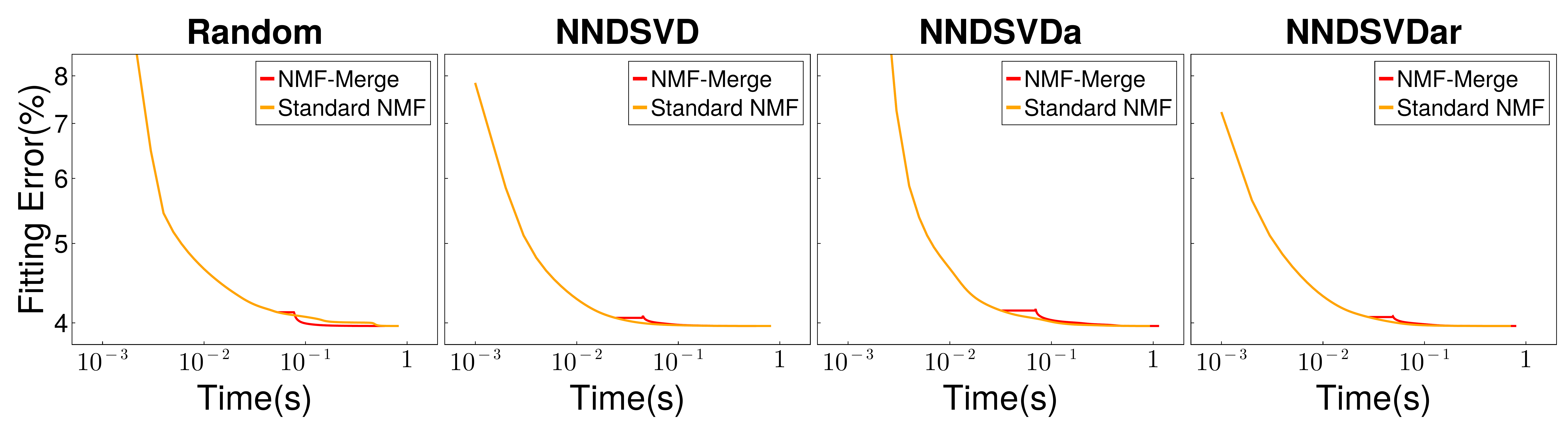}}%
        \hfil
        \subfloat[]{\includegraphics[width=3.2in]{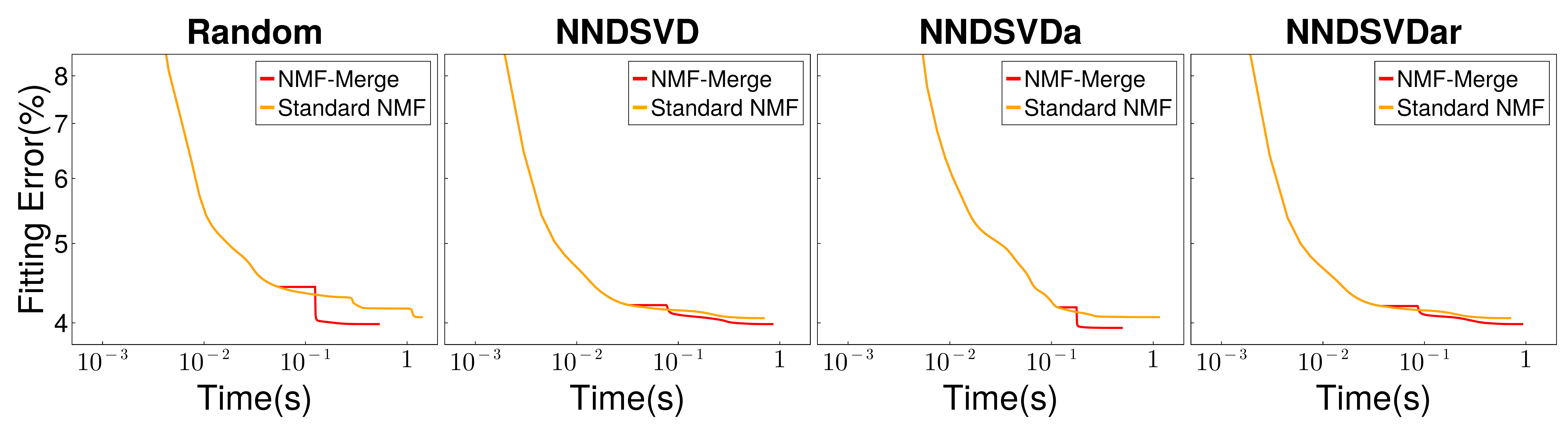}}%
        
        \subfloat[]{\includegraphics[width=3.2in]{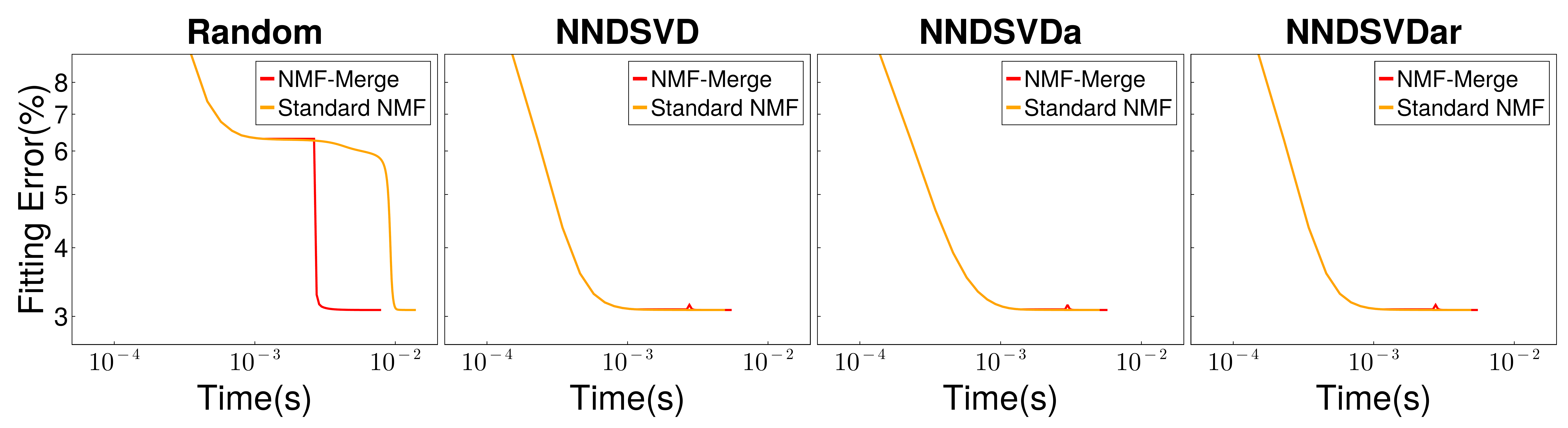}}%
        \hfil
        \subfloat[]{\includegraphics[width=3.2in]{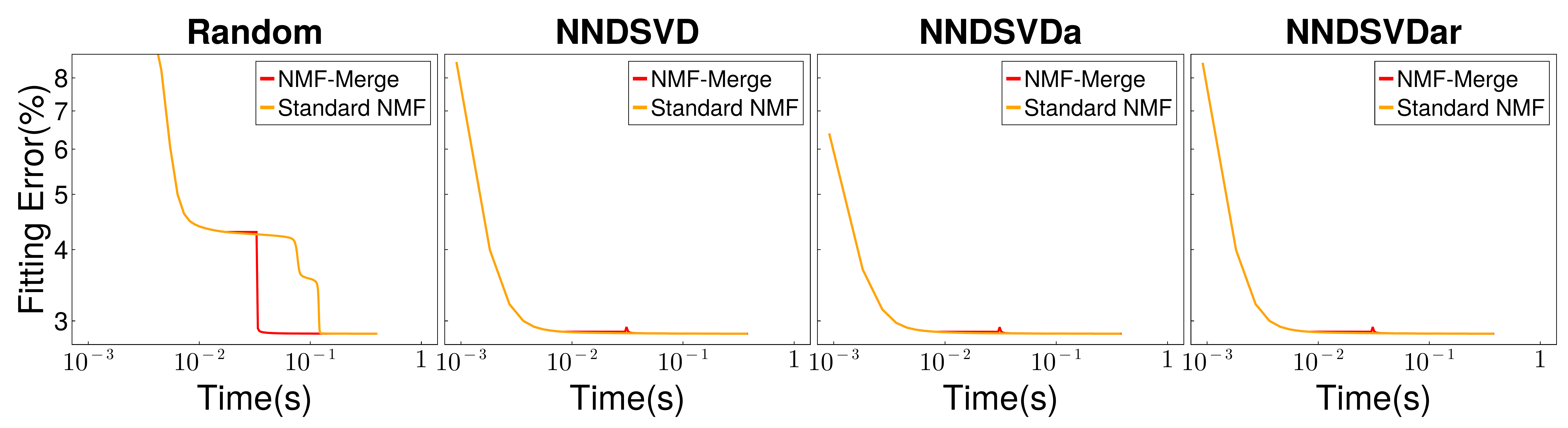}}%
        
        \subfloat[]{\includegraphics[width=3.2in]{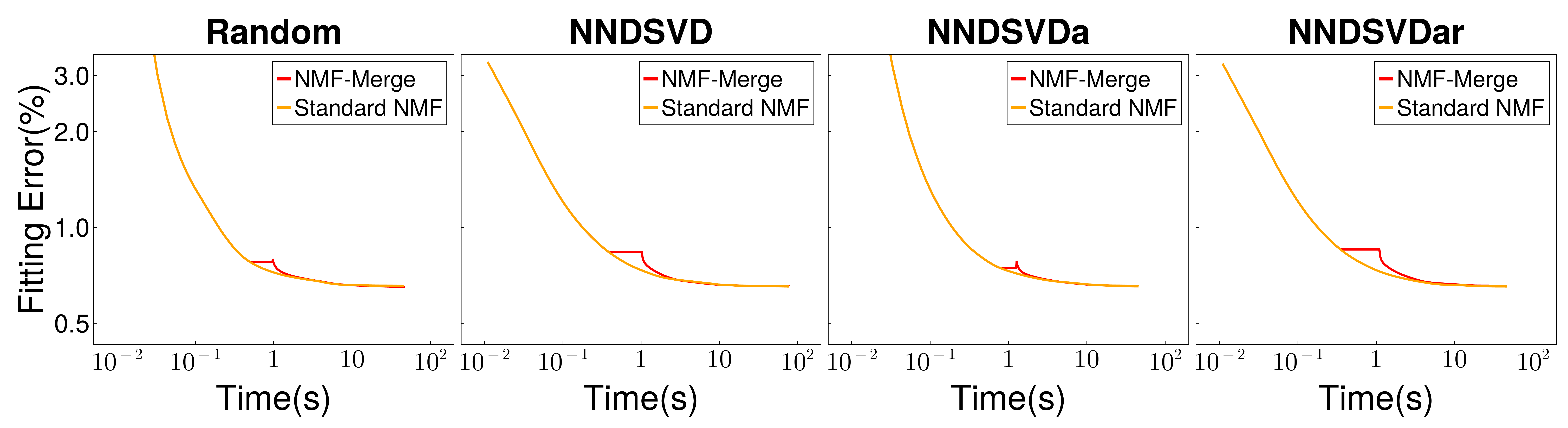}}%
        \hfil
        \subfloat[]{\includegraphics[width=3.2in]{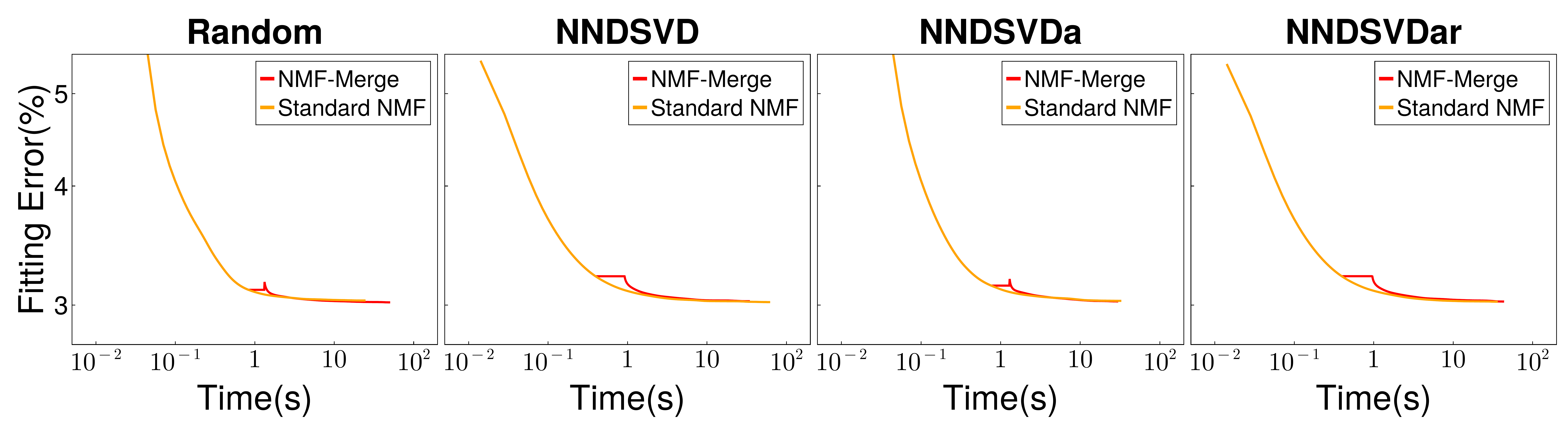}}%
        \caption{The change of the fitting error versus the time of NMF-Merge and Standard NMF: (a) LCMS1. (b) LCMS2. (c) Mary had a little lamb. (d) Prelude and Fugue No.1 in C major. (e) CBCL. (f) ORL. }
        \label{fig_trace_t}
    \end{figure*}
    
    \begin{figure*}[!t]
        \centering
        \subfloat[]{\includegraphics[width=3in]{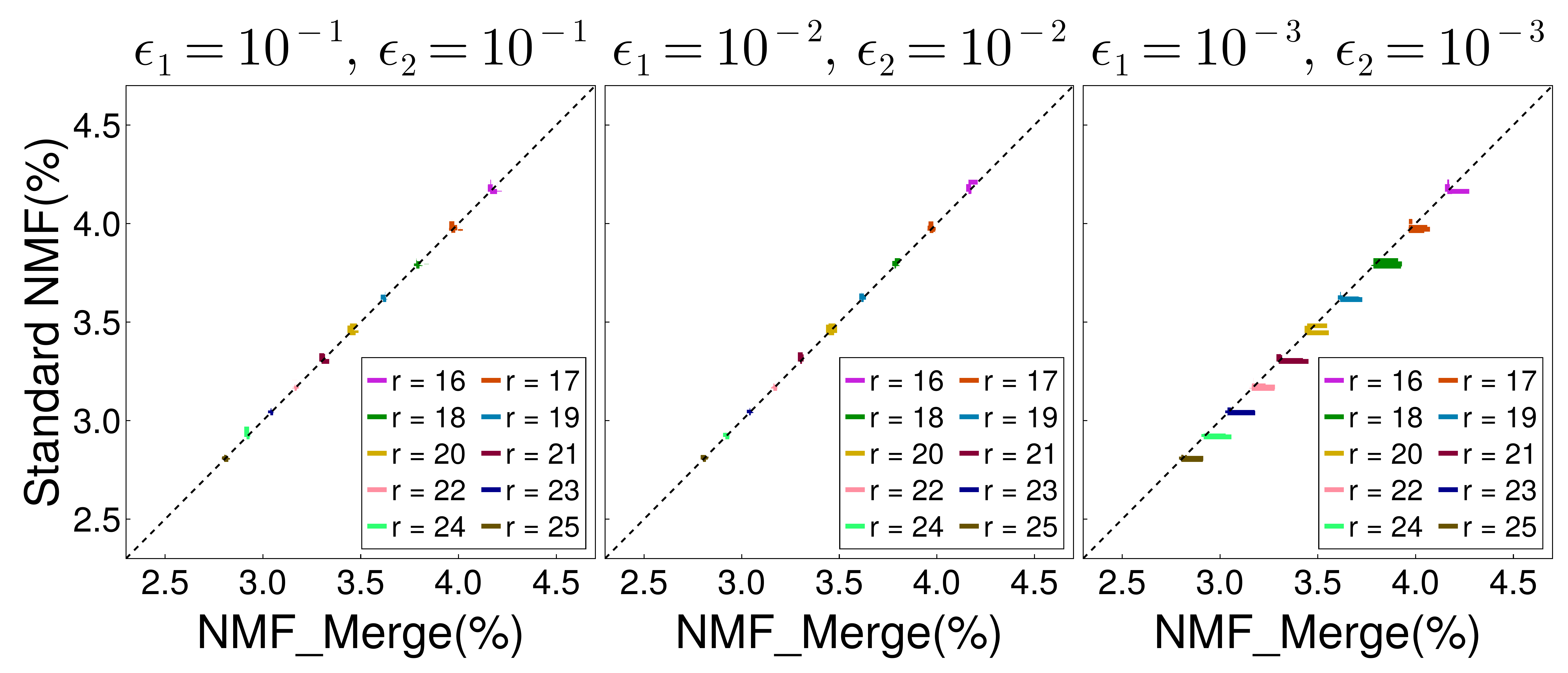}}%
        \hfil
        \subfloat[]{\includegraphics[width=3in]{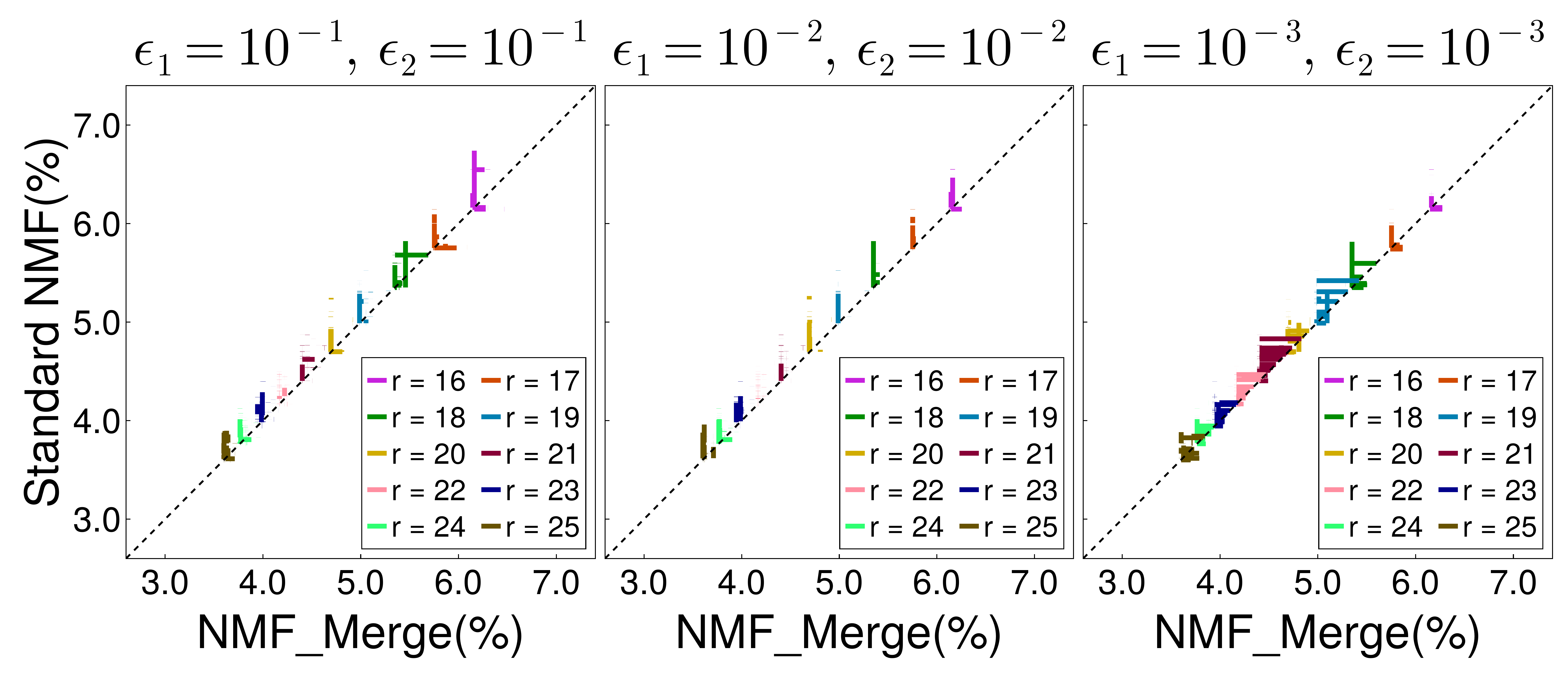}}%
        
        \subfloat[]{\includegraphics[width=3in]{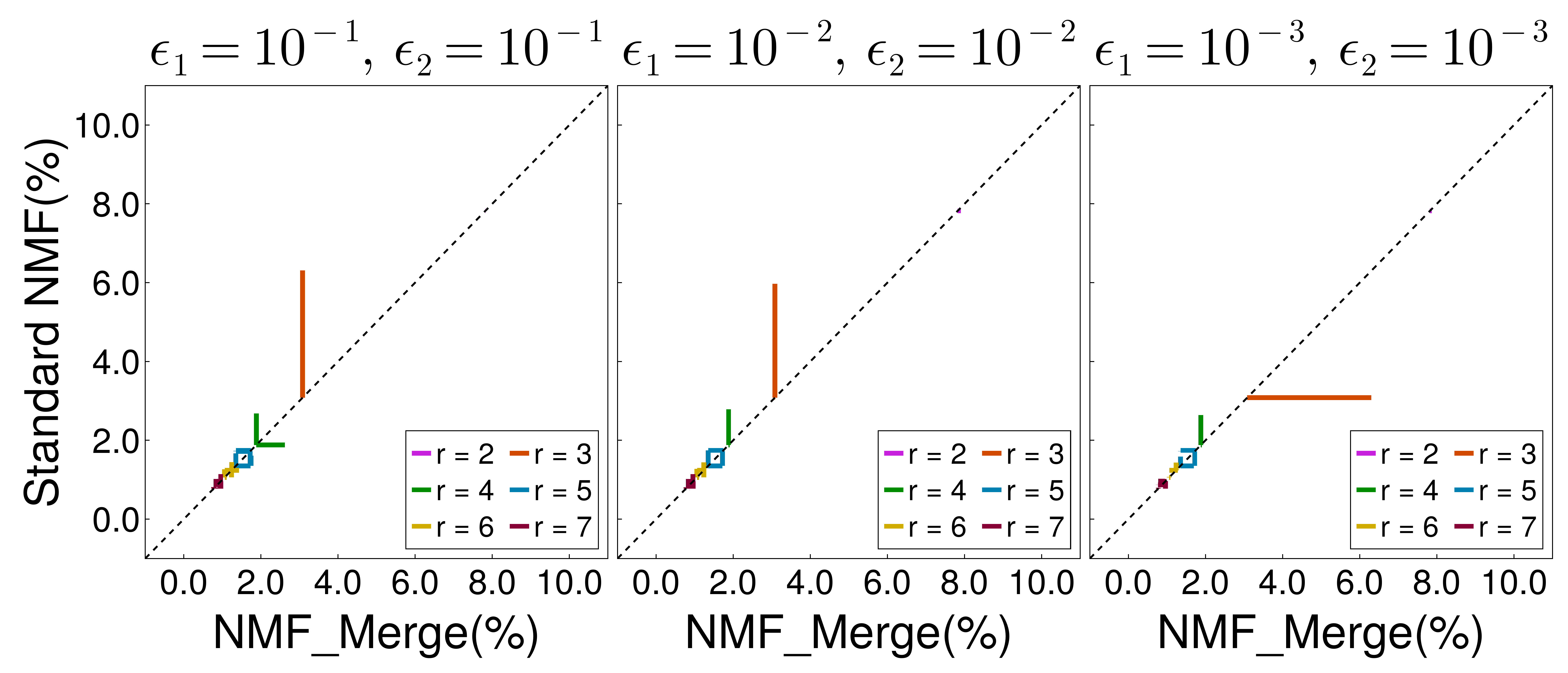}}%
        \hfil
        \subfloat[]{\includegraphics[width=3in]{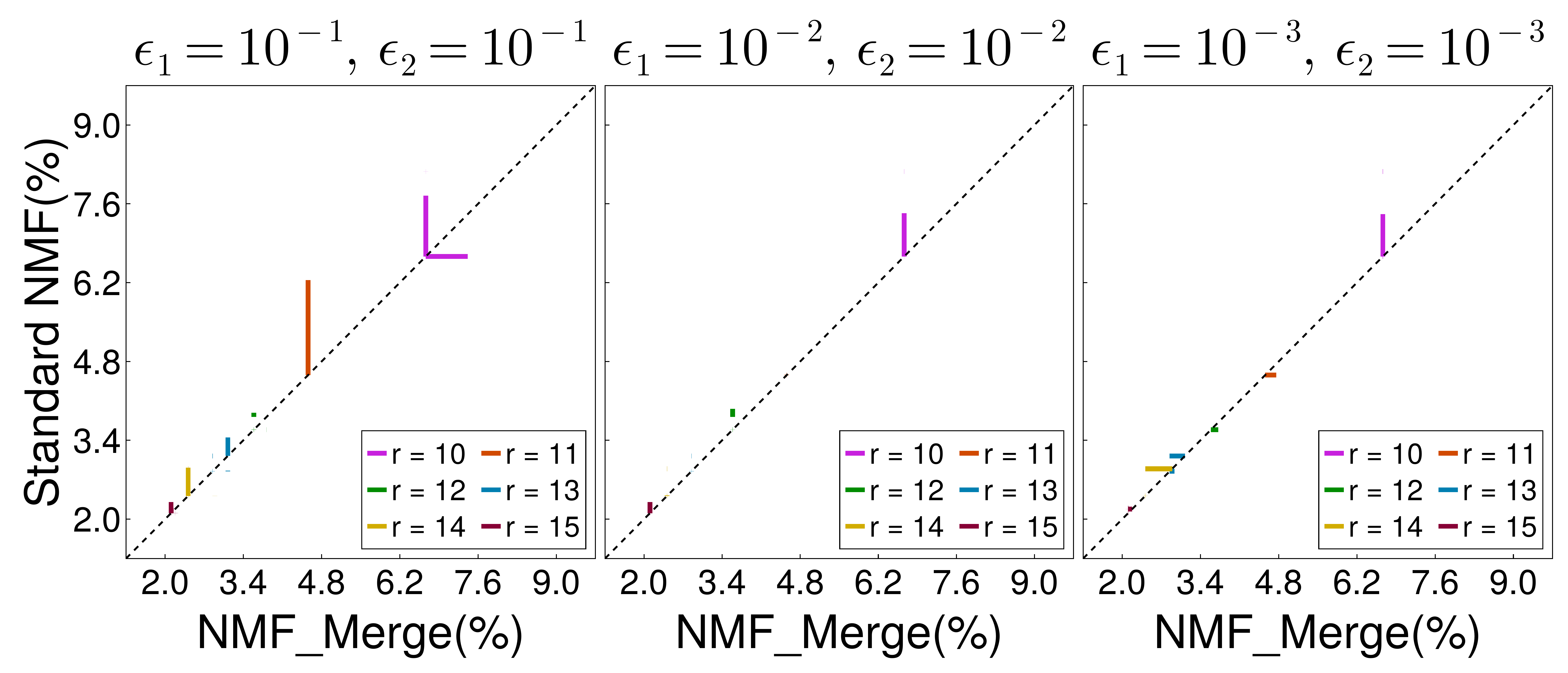}}%

        \subfloat[]{\includegraphics[width=3in]{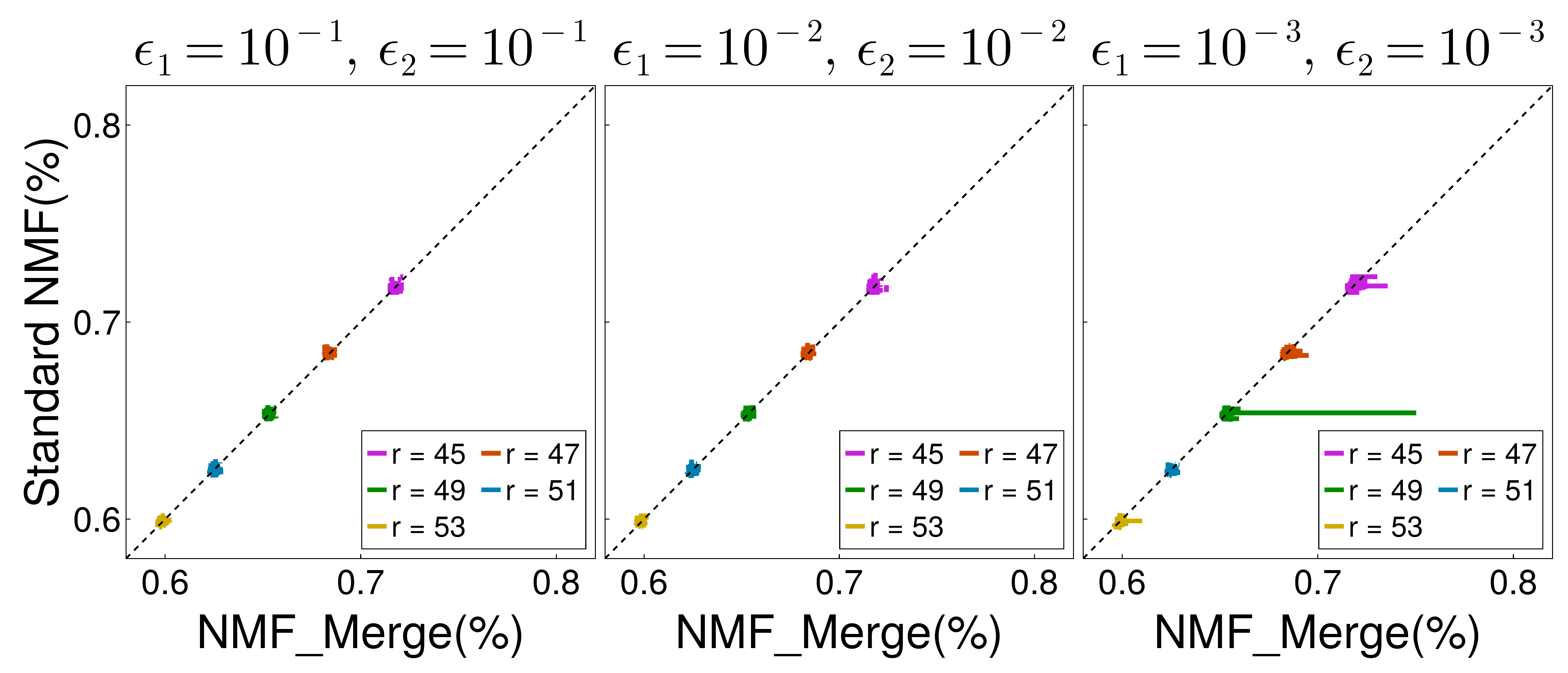}}%
        \hfil
        \subfloat[]{\includegraphics[width=3in]{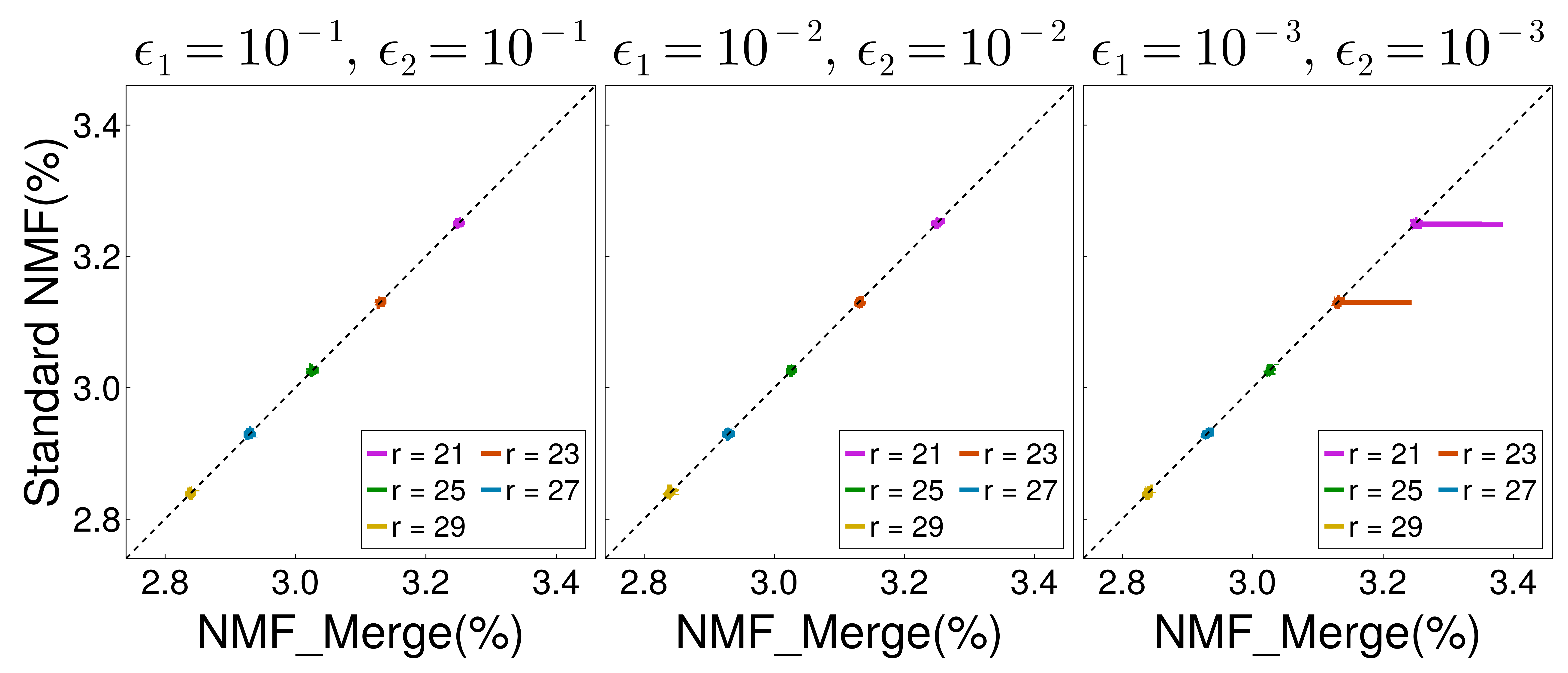}}%
        \caption{The change of the fitting error of NMF-Merge versus the fitting error of standard NMF at the same time after any one of the two methods finished: (a) LCMS1. (b) LCMS2. (c) Mary had a little lamb. (d) Prelude and Fugue No.1 in C major. (e) CBCL. (f) ORL. }
        \label{fig_trace_phase}
    \end{figure*} 

    To compare the two approaches more systematically, we adopt the phase-plot representation and overlay trajectories for many different random initializations (Fig.  \ref{fig_trace_t_example}(c)); to reduce complexity, only the portion of the curve after one of the methods has converged (the blue portion of Fig.  \ref{fig_trace_t_example}(b)) is shown.  
    Thus, horizontal lines mean standard NMF finished first, and vertical lines mean that NMF-Merge finished first. Because 200 independent random initializations are aggregated, both horizontal and vertical lines may be present. Fig. \ref{fig_trace_t_example} shows that in addition to yielding a better outcome, NMF-Merge also frequently converged first.
    
    Traces showing the objective value vs.\ time with different initialization methods for six datasets are in Fig.  \ref{fig_trace_t}. 
    The caption of each panel within the subplot specifies how the initial guess was chosen for the run of both NMF pipelines.
    Similar to Fig. \ref{fig_trace_t_example}, for random initialization, we illustrate a single favorable result; in contrast, the SVD-based initializations are deterministic and thus the traces are complete. 
    Fig.  \ref{fig_trace_t}(a)-(d) shows that NMF-Merge avoids plateau phenomenon exhibited by standard NMF on LCMS1, LCMS2, ``Mary had a little lamb'' and ``Prelude and Fugue No.1 in C major''. 
    On CBCL and ORL face images, standard NMF does not exhibit obvious plateau phenomenon with $\epsilon = 10^{-4}$ (Fig.  \ref{fig_trace_t}(e) and (f)), and NMF-Merge correspondingly exhibits no performance advantage.
    For these examples we used $\epsilon_1 =  \epsilon_2 = 10^{-2}$. Reducing $\epsilon_1, \epsilon_2$ may decrease efficiency, as indicated by the rightward movement of the red line in each figure, while enhancing the probability of identifying superior local optima. 
    Conversely, increasing the tolerance (e.g., $\epsilon_1 = \epsilon_2 = 10^{-1}$) can enhance efficiency, but it may also reduce the likelihood of finding better local optima. 
    However, in certain datasets like LCMS, setting $\epsilon_1 = \epsilon_2 = 10^{-1}$ resulted in superior local optima in most cases while also allowing NMF-Merge to consistently beat standard NMF in speed of convergence.

    \begin{table*}[]
    \renewcommand{\arraystretch}{1.5}
        \centering
        \caption{Time (mean$\pm$ std) of each stage of NMF-Merge}
        \label{tpart}
        \begin{tabular}{|c|c|c||c|c|c|c|c|}\hline
             \multirow{2}{*}{Data sets} &\multirow{2}{*}{$r$} & \multirow{2}{*}{$k$} & \multicolumn{5}{c|}{Time($s$), $(\epsilon_1, \epsilon_2) = (10^{-2}, 10^{-2})$} \\ \cline{4-8}
              & & & Initial NMF & Feature recovery  & Over-complete NMF & Merge & Final NMF \\ \hline \hline
             LCMS1& 17 & 3 & $0.0360\pm0.0103$ & $\mathbf{0.0022\pm0.0005}$ & $0.0267\pm0.0090$ & $\mathbf{0.0004\pm0.0000}$ & $0.9036\pm0.3380$ \\ \hline 
             LCMS2& 23 & 5 & $0.0683\pm0.0249$ & $\mathbf{0.0027\pm0.0004}$ & $0.0526\pm0.0249$ & $\mathbf{0.0005\pm0.0000}$ & $0.4490\pm0.1051$\\ \hline 
             MHLL & 3 & 1 & $0.0018\pm0.0007$ & $\mathbf{0.0002\pm0.0000}$ & $0.0014\pm0.0002$ & $\mathbf{0.0000\pm0.0000}$ & $0.0038\pm0.0018$\\ \hline 
             P\&F No.1 & 13 & 3 & $0.0250\pm0.0113$ & $\mathbf{0.0017\pm0.0003}$ & $0.0239\pm0.0079$ & $\mathbf{0.0003\pm0.0000}$ & $0.3105\pm0.0717$\\ \hline 
             CBCL & 49 & 10 & $0.5428\pm0.0902$ & $\mathbf{0.0256\pm0.0128}$ & $0.4312\pm0.0636$ & $\mathbf{0.0062\pm0.0004}$ & $44.7117\pm16.9786$ \\ \hline 
             ORL & 25 & 5 & $0.7844\pm0.0627$ & $\mathbf{0.0268\pm0.0084}$ & $0.4289\pm0.0592$ & $\mathbf{0.0077\pm0.0028}$ & $38.4424\pm15.3281$\\ \hline 
        \end{tabular}
    \end{table*}

    Fig.  \ref{fig_trace_t} chooses only a single random initialization for illustration; more comprehensive results are shown in Fig.  \ref{fig_trace_phase}, which illustrates the results in Fig.  \ref{fig_trace_t_example}(c) for different $\epsilon_1$, $\epsilon_2$, and datasets.
    Fig.  \ref{fig_trace_phase} shows that NMF-Merge typically finished before standard NMF on LCMS1, LCMS2 and ``Prelude and Fugue No.1 in C major'' when initial NMF and over-complete NMF tolerance were no more stringent than $10^{-2}$. 
    ``Mary had a little lamb'' is a special case since its small rank allows standard NMF to be highly efficient with little benefit from extra steps of NMF-Merge. 
    On CBCL face images and ORL face images (Fig.  \ref{fig_trace_phase}(e) and (f)), performance of the two methods was similar for $\epsilon_1 = \epsilon_2 \ge 10^{-2}$. At more stringent tolerances, there are still runs for which NMF-Merge is more efficient (e.g., LCMS2 in Fig.  \ref{fig_trace_phase}(b) and ``Prelude and Fugue No.1 in C major'' in Fig.  \ref{fig_trace_phase}(d)), but more commonly it becomes less efficient than standard NMF. As a consequence, we recommend $\epsilon_1 = \epsilon_2 = 10^{-2}$ (given a final tolerance of $10^{-4}$) as a good balance between quality and performance. 
    These experimental result show that NMF-Merge maintains competitive performance by avoiding plateau phenomena during iteration, but rigorous insights into the mechanism by which this occurs are a topic for future work.

    The contribution of separate stages of NMF merge are documented in  Table~\ref{tpart}, which shows the mean and standard variation of time for each portion of NMF-Merge. One sees that the ``feature recovery'' and ``merge'' steps are very efficient, and represent a nearly-negligible fraction of the time for most data sets. 
    Thus, the efficiency of NMF-Merge primarily depends on the time devoted to each of the three stages of NMF.
    In other words, $\epsilon_1$ and $\epsilon_2$ are the main control parameters for the runtime performance of NMF-Merge, and we find that they can be set to achieve a favorable balance of quality and performance.
    
    \subsection{The choice of extra components number}
    \label{The choice of extra components number}
    
     \begin{figure}[!t]
    \centering
        \subfloat[]{\includegraphics[width=1.1in]{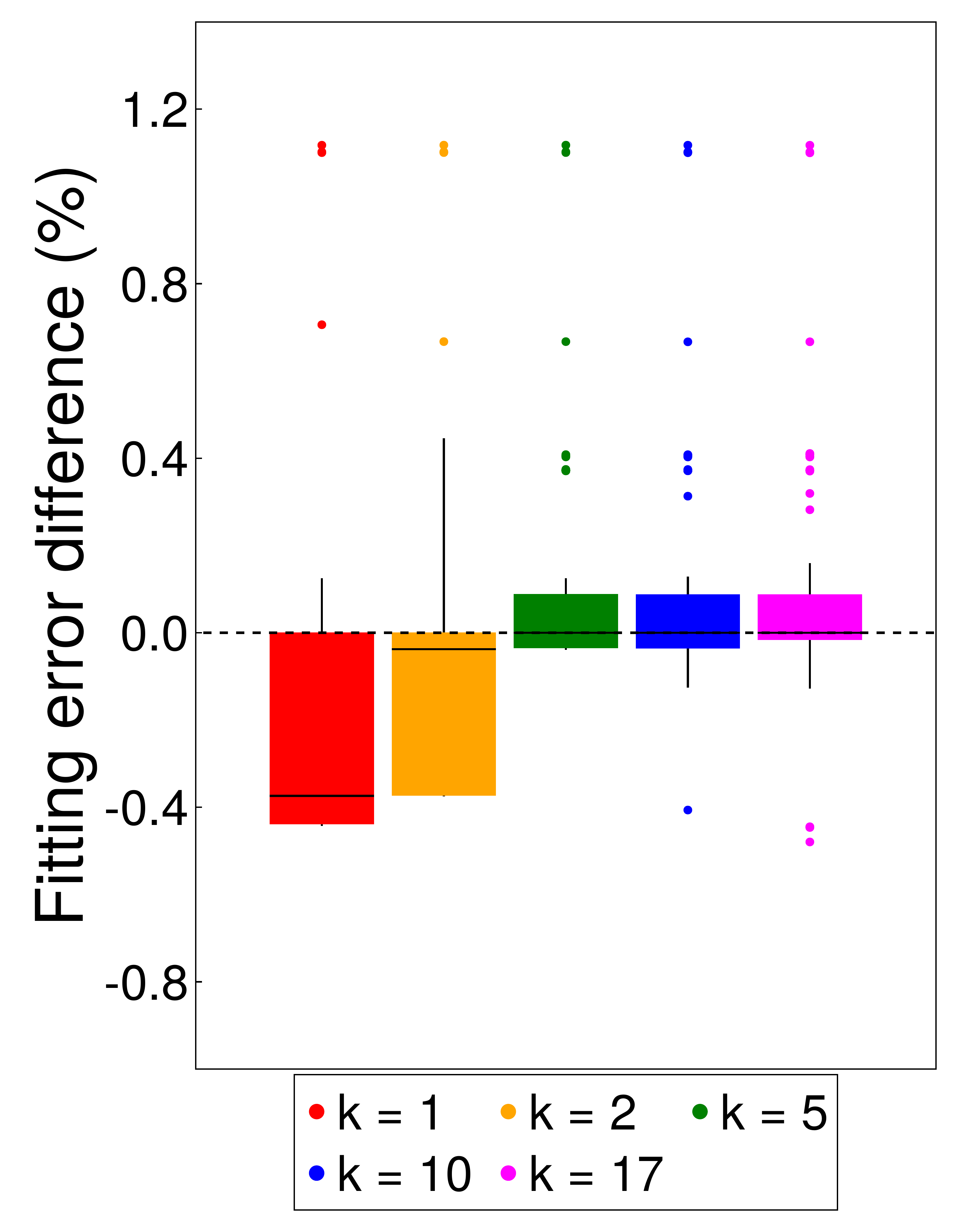}}%
        \subfloat[]{\includegraphics[width=1.1in]{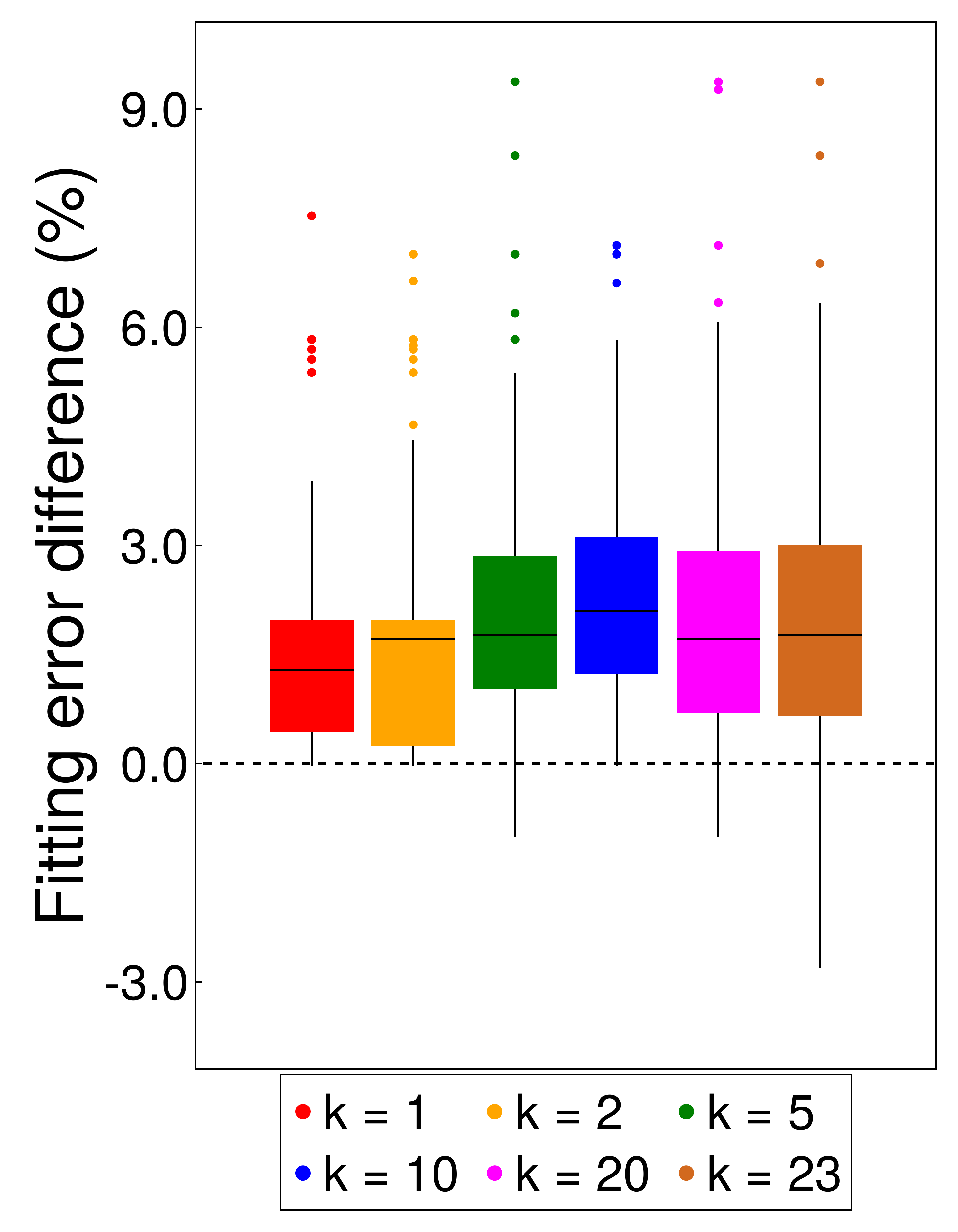}}%
        \subfloat[]{\includegraphics[width=1.1in]{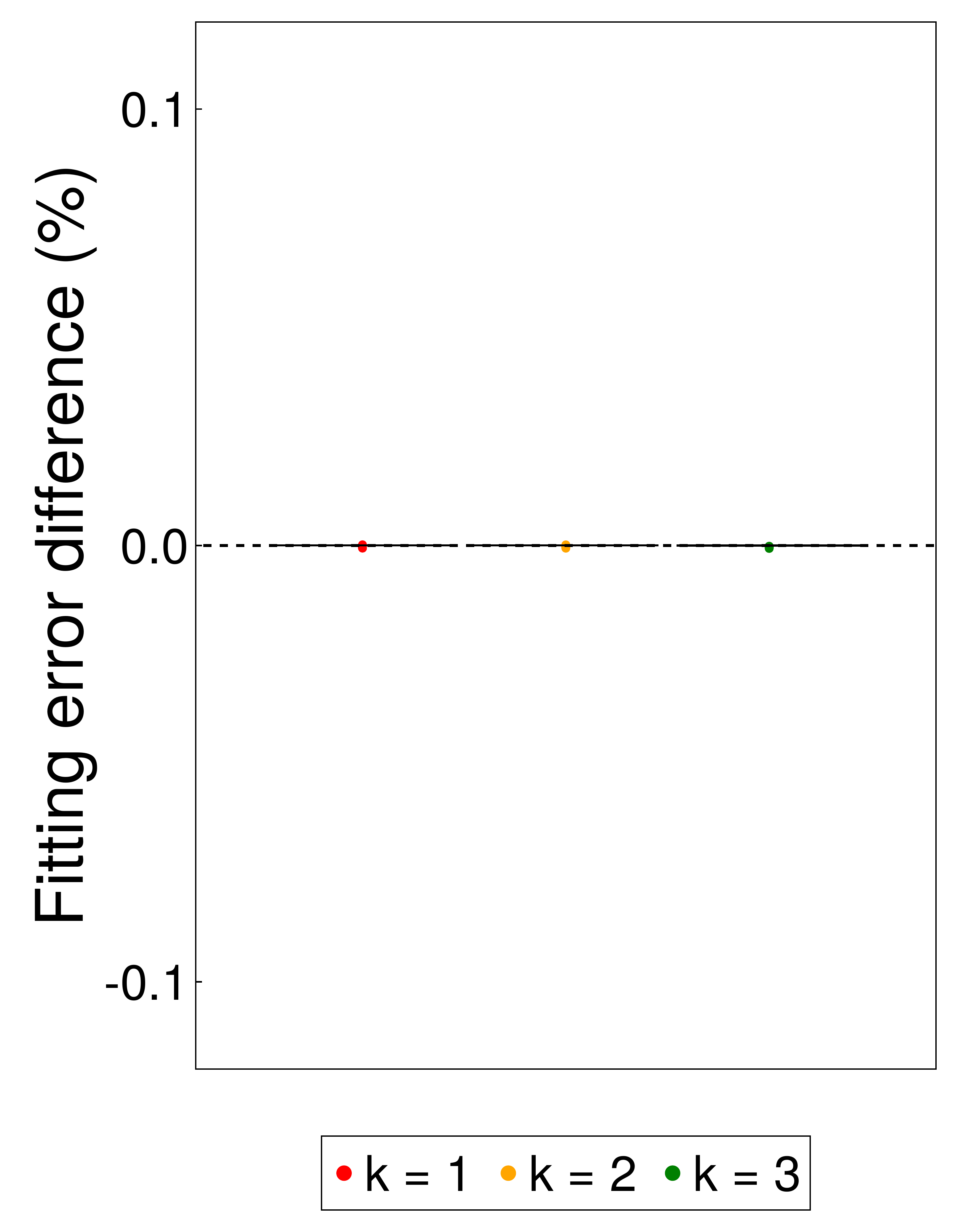}}%

        \subfloat[]{\includegraphics[width=1.1in]{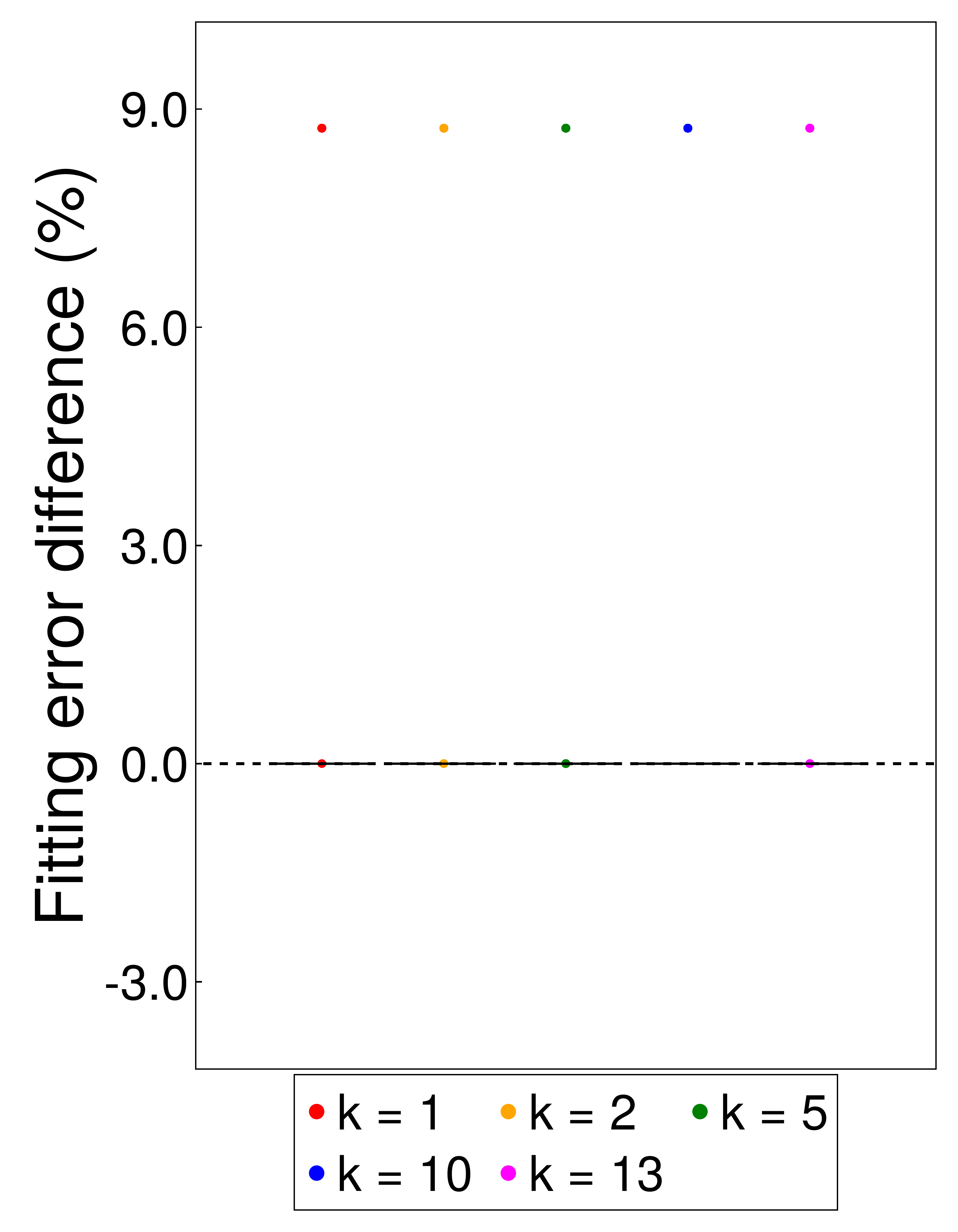}}%
        \subfloat[]{\includegraphics[width=1.1in]{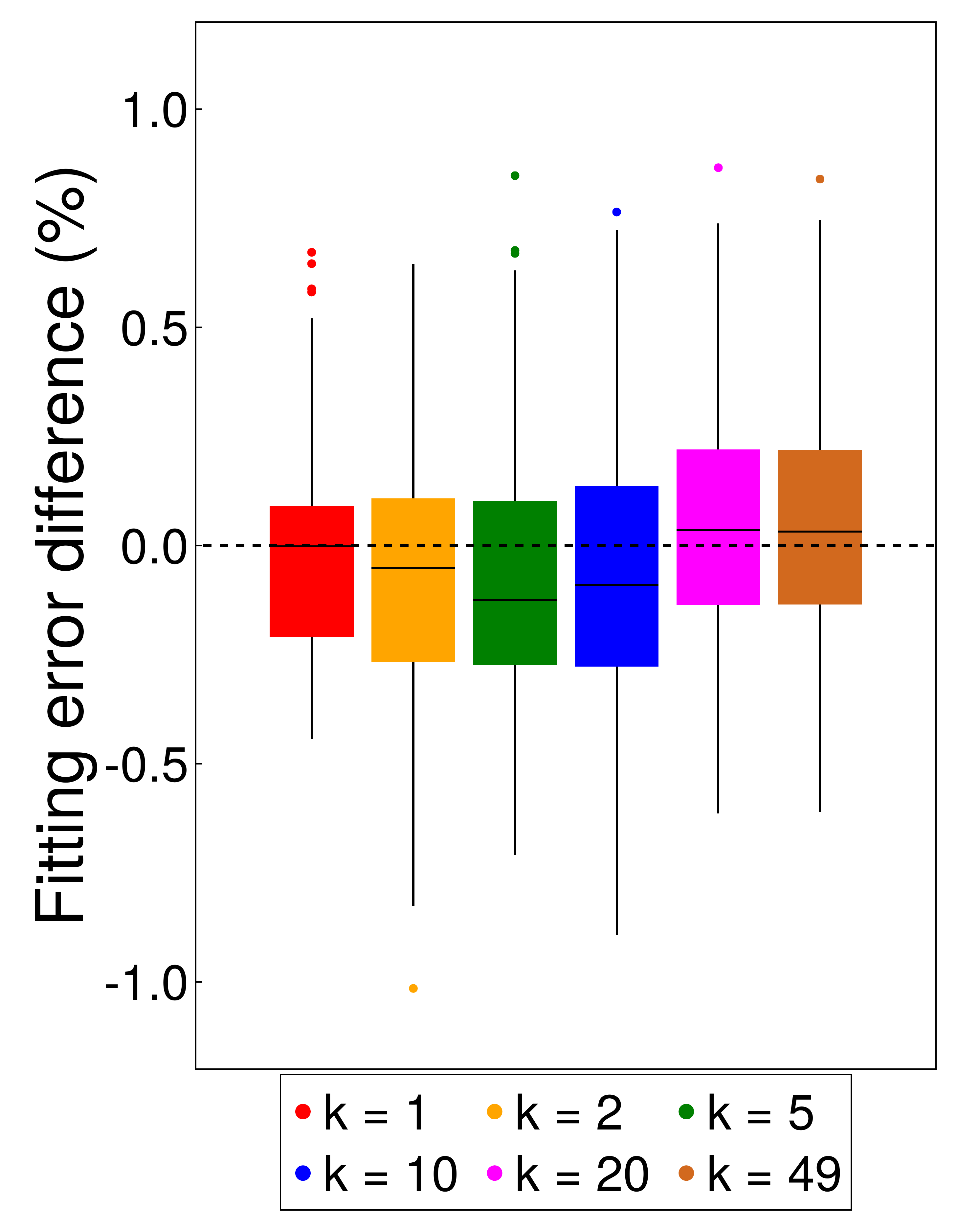}}%
        \subfloat[]{\includegraphics[width=1.1in]{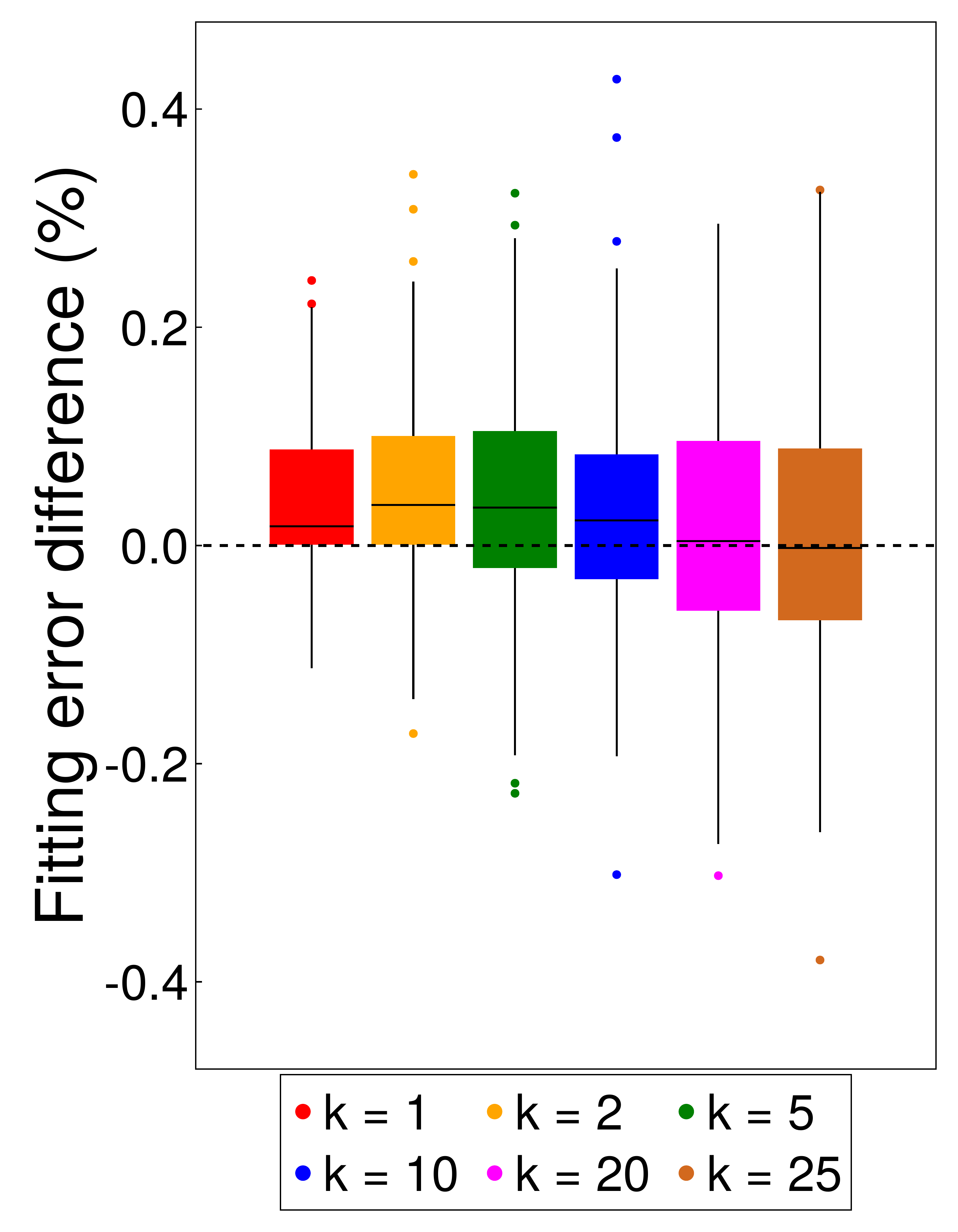}}%
        \caption{The difference of fitting error between standard NMF and NMF-Merge with different extra components number: (a) LCMS1. (b) LCMS2. (c) Mary had a little lamb. (d) Prelude and Fugue No.1 in C major. (e) CBCL. (f) ORL.}
        \label{boxplot_kadd}
    \end{figure}
    It should be noted that all previous results were obtained by generating 20\% additional components during the feature recovery stage, a systematic procedure that appears to be broadly successful.
    However, since adding more components might be expected to improve feature recovery, but also increase the amount of time required for over-complete NMF and merging, it is valuable to investigate the role of this parameter on final results.
    
    The experiment in Sec.\ref{NMF-Merge helps NMF converge to better local optima} (200 random initialized trials) was conducted with different numbers of extra components. 
    Fig.  \ref{boxplot_kadd} illustrates the difference between the fitting error of standard NMF and that of NMF-Merge; positive differences correspond to an advantage for NMF-Merge. 
    The outcomes are consistent with the results in Fig.  \ref{fig_scatter_spec_r_spec_tol}, with NMF-Merge broadly outperforming standard NMF for most numbers of extra components on LCMS1, LCMS2 and ``Prelude and Fugue No.1 in C major''.
    On ``Mary had a little lamb'', CBCL and ORL face images, the two methods perform similarly. 
    We note there are some cases where adding too few components can be detrimental to the solution quality (e.g., Fig.  \ref{boxplot_kadd}(a)). Aside from this case, however, the final results are relatively insensitive to small alterations in the number of extra components.

\section{Conclusion}
\label{Conclusion}
We proposed a new multistage NMF pipeline.
% It finds features that are not learned by standard NMF using generalized SVD-based feature recovery and optimizes them by over-complete NMF. 
It adds components to existing NMF solutions using generalized SVD-based feature recovery and optimizes them by over-complete NMF.
New components are recombined together by our analytically solvable pairwise merge algorithm and then optimized by final NMF. 
% We show that NMF-Merge can help efficiently estimate the number of components needed.
We showed that it outperformed standard NMF by helping existing non-ideal NMF solutions escape poor local optima, and improves not just the quality but also consistency of NMF solutions. 
% Moreover, because the NMF-Merge avoids \youdong{plateau phenomenon} during iteration, it is computationally efficient. 
Moreover, because NMF-Merge avoids a plateau phenomenon that slows convergence of other algorithms, it maintains competitive computational performance.
Overall, we believe that NMF-Merge is well-positioned to become a standard approach for performing nonnegative matrix factorization.
 
All code for this manuscript is open-source and available at \texttt{https://github.com/HolyLab/NMFMerge.jl}.

\section*{Acknowledgments}
We thank Jody O'Sullivan for comments on this manuscript.
This work was supported by NIH grants R01DC020034, R01DC010381 (T.E.H.), and training grant T32EB014855 (Y.G, PI: Joseph P. Culver).

{\appendices

\section{Illustration of plateau phenomenon in NMF}
\label{Illustration of plateau phenomenon in NMF}
The non-convexity of NMF can cause optimization algorithms to enter extended periods of very slow progress towards the solution that alternate with short periods of rapid decrease in the objective value, a phenomenon we call ``plateau phenomenon''. 
This phenomenon was previously investigated in neural network training and was found to be caused primarily by saddle points or poor local minima of the objective function\cite{fukumizu2000local,ainsworth2021plateau, yoshida2019data}.
To illustrate this plateau phenomenon for NMF, consider a case with ground truth
\begin{align} 
        \begin{aligned}
        \label{simu_matrix}
            \mathbf{W} = \begin{pmatrix}
                6 & 0 & 4 & 9 \\
                0 & 4 & 8 & 3 \\
                4 & 4 & 0 & 7 \\
                9 & 1 & 1 & 1 \\
                0 & 3 & 0 & 4 \\
                8 & 1 & 4 & 0 \\
                0 & 0 & 4 & 2 \\
                0 & 9 & 5 & 5 
            \end{pmatrix}, \quad
            \mathbf{H}^{\mathrm{T}} = \begin{pmatrix}
                6 & 0 & 3 & 4 \\
                10 & 10 & 5 & 9 \\
                8 & 2 & 0 & 10 \\
                2 & 9 & 2 & 7 \\
                0 & 10 & 4 & 7 \\
                1 & 6 & 0 & 0 \\
                2 & 0 & 0 & 0 \\
                10 & 0 & 8 & 0
            \end{pmatrix}
        \end{aligned}
    \end{align}
 These components are depicted in Fig. \ref{stalling_example}(a). Performing NMF on $\mathbf{X} = \mathbf{WH}$, the change of the objective value versus iteration number is illustrated in Fig. \ref{stalling_example}(b). 
 After initial rapid progress, optimization enters two separate plateau periods, the first roughly between 20 and 250 iterations, and the second between 350 to 5200 iterations.
 If the NMF is terminated at 100 iterations (during the first plateau), the recovered features differ substantially from the ground truth (Fig. \ref{stalling_example}(c)). Full recovery happens only with complete convergence (Fig. \ref{stalling_example}(d)), but executing $\approx 10000$ iterations is often viewed as impractical in applications. Thus, the poor convergence properties of NMF can have practical consequences for the quality of the obtained solutions.

\begin{figure}[!t]
    \centering
    \subfloat[]{\includegraphics[width=1.7in]{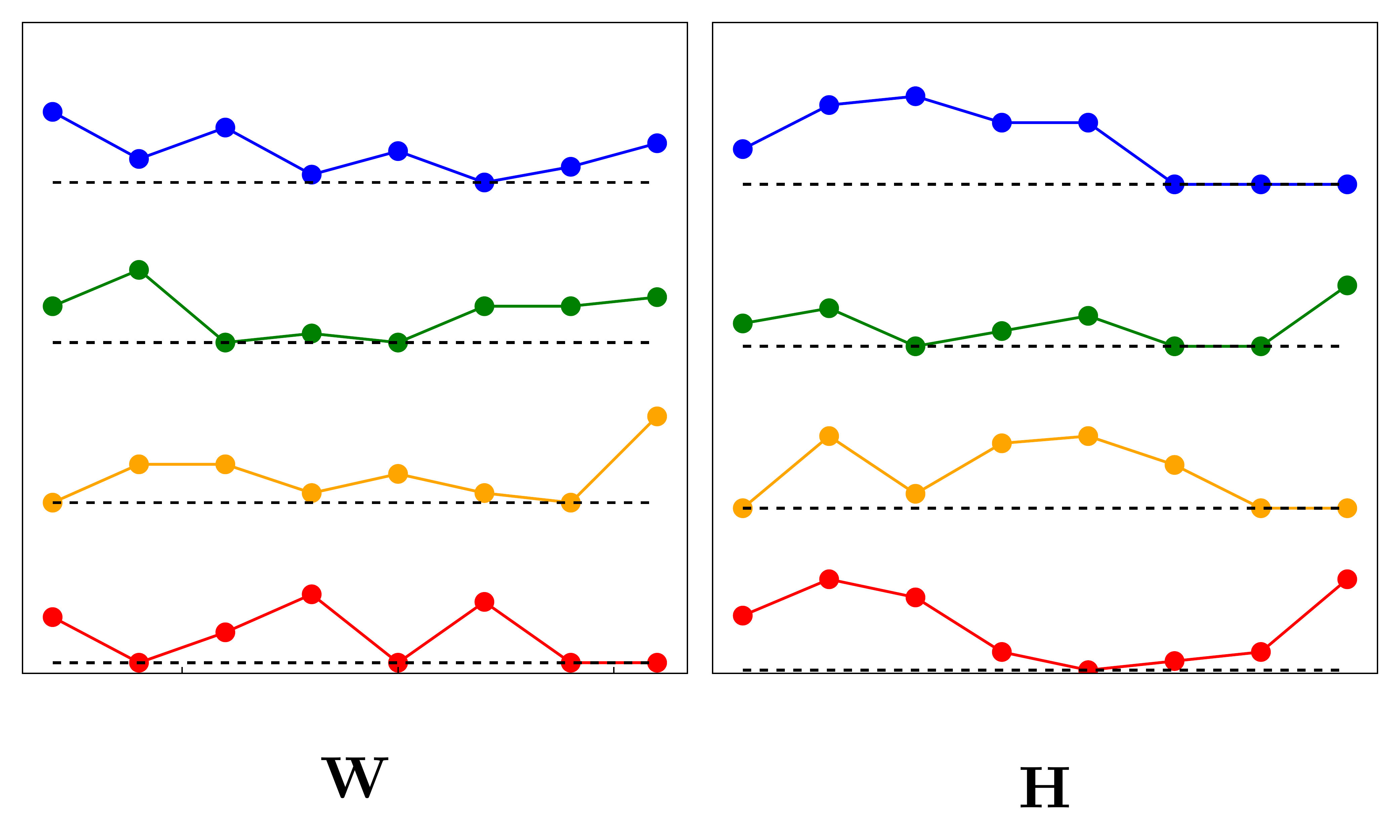}}%
    \subfloat[]{\includegraphics[width=1.7in]{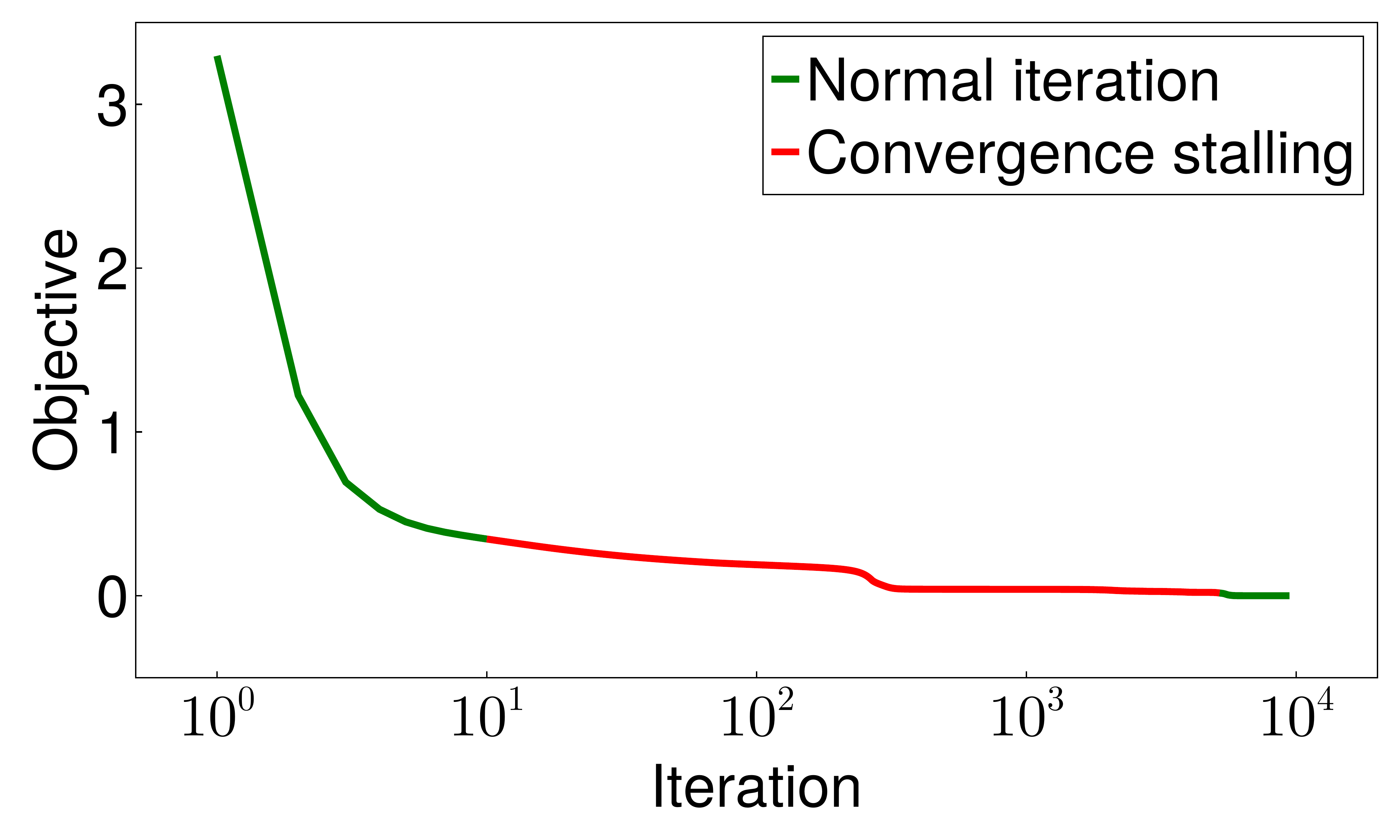}}%

    \subfloat[]{\includegraphics[width=1.7in]{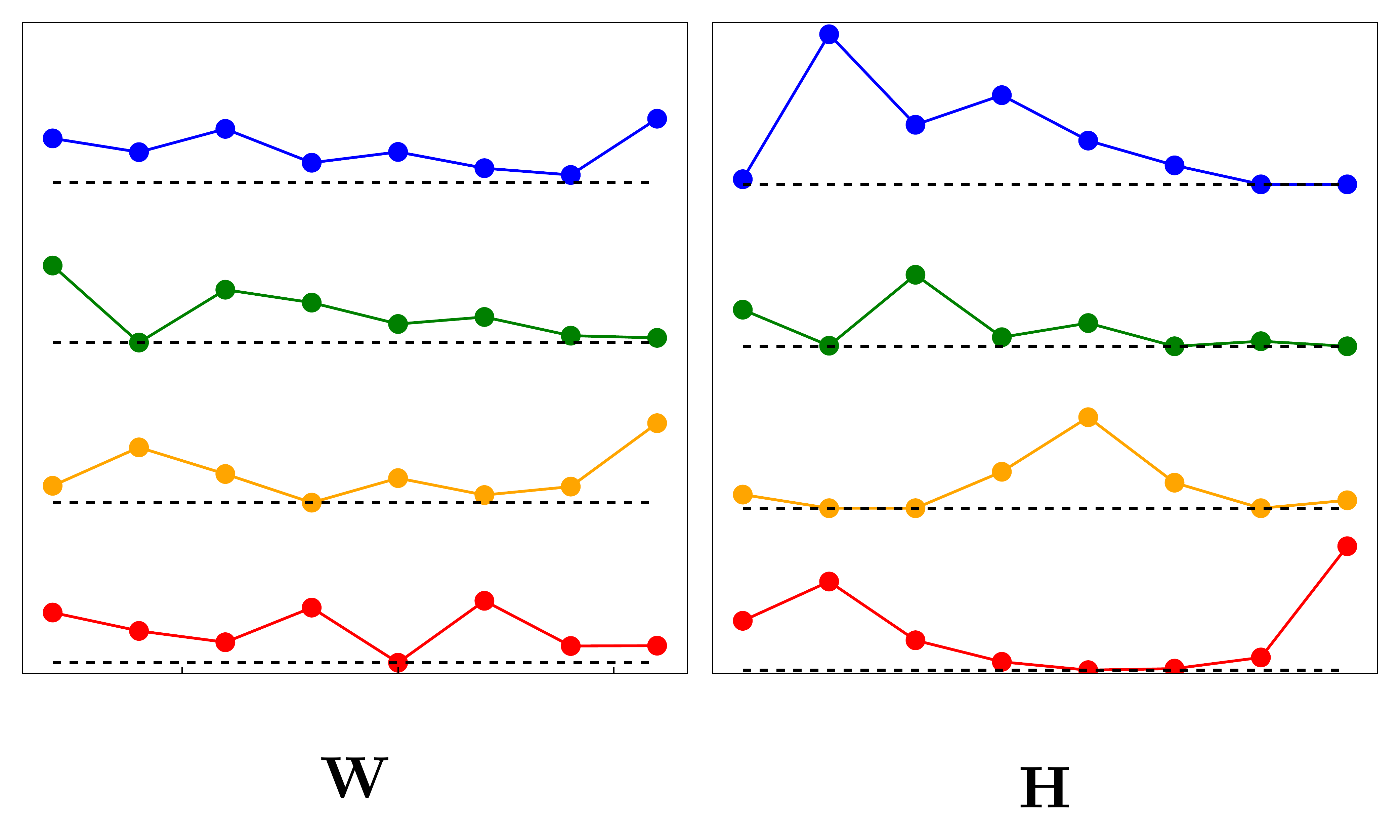}}%
    \subfloat[]{\includegraphics[width=1.7in]{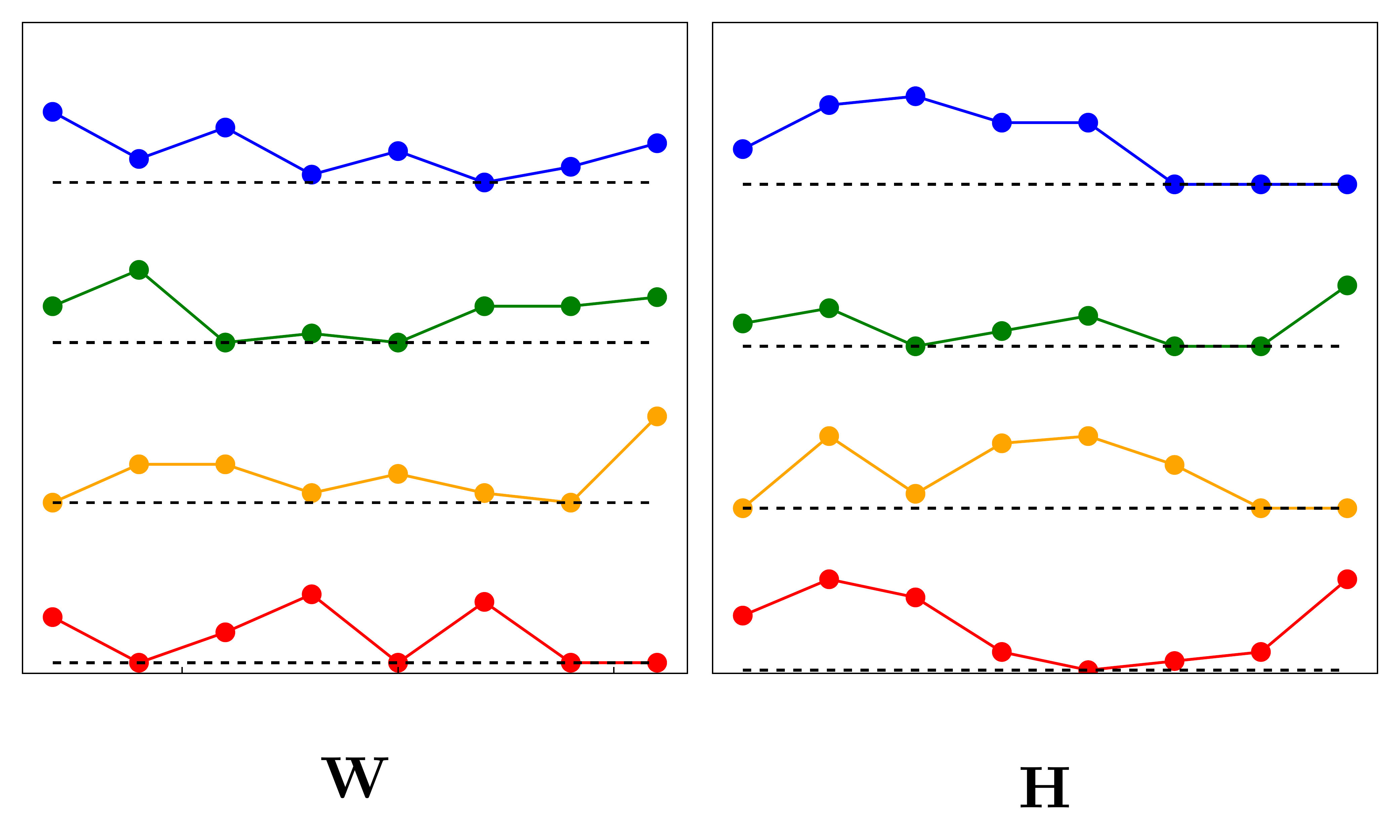}}%
    \caption{NMF on matrix $\mathbf{WH}$ in (\ref{simu_matrix}). (a) Ground truth of $\mathbf{W}$ and $\mathbf{H}$. (b) plateau phenomenon during iteration in NMF on $\mathbf{WH}$ in (3). (c) $\mathbf{W}$ and $\mathbf{H}$ after 100 iterations. (d) $\mathbf{W}$ and $\mathbf{H}$ after completely convergence.}
    \label{stalling_example}
\end{figure}

\section{Derivation of optimal merge model}
\label{Derivation of optimal merge model}
To optimize (\ref{DE_X_X_merge}), we add a Lagrange multiplier to enforce the normalization constraint, 
    \begin{align}
        \begin{aligned}
            \label{DE_X_X_merge_+lagrange}
            P = \lVert\mathbf{w}_p\mathbf{h}_p^\mathrm{T}+\mathbf{w}_q\mathbf{h}_q^\mathrm{T}-\mathbf{w}_\mathrm{m}\mathbf{h}_\mathrm{m}^\mathrm{T}\rVert^2+\lambda \left(\lVert\mathbf{w}_\mathrm{m}\rVert^2-1\right)
        \end{aligned}
    \end{align}
Computing the derivative of (\ref{DE_X_X_merge_+lagrange}) with respect to $\mathbf{w}_\mathrm{m}$ yields
    \begin{align}
        \begin{aligned}
            \label{DE_X_X_merge_d_w_m}
            \frac{1}{2}\frac{\partial P}{\partial\mathbf{w}_\mathrm{m}}=h_\mathrm{m}^2\mathbf{w}_\mathrm{m}-\mathbf{h}_p^\mathrm{T}\mathbf{h}_\mathrm{m}\mathbf{w}_p-\mathbf{h}_q^\mathrm{T}\mathbf{h}_\mathrm{m}\mathbf{w}_q+\lambda\mathbf{w}_\mathrm{m}
        \end{aligned}
    \end{align}
where $h_\mathrm{m} = \lVert\mathbf{h}_\mathrm{m}\rVert$. Letting $\frac{\partial P}{\partial\mathbf{w}_\mathrm{m}}=0$ yields
    \begin{align}
        \begin{aligned}
            \label{optial_w_m}
                \mathbf{w}_\mathrm{m} = \frac{(\mathbf{h}_p^\mathrm{T}\mathbf{h}_\mathrm{m})\mathbf{w}_p+(\mathbf{h}_q^\mathrm{T}\mathbf{h}_\mathrm{m})\mathbf{w}_q}{h_\mathrm{m}^2+\lambda}
        \end{aligned}
    \end{align}
All but the $\mathbf{w}$ terms are (unknown) scalars, and hence $\mathbf{w}_\mathrm{m}$ is a linear combination of $\mathbf{w}_\mathrm{p}$ and $\mathbf{w}_\mathrm{q}$.
Similarly, at optimum 
% $\mathbf{h}_\mathrm{m}$ is
    \begin{align}
        \begin{aligned}
            \label{optial_h_m}
                \mathbf{h}_\mathrm{m} = (\mathbf{w}_p^\mathrm{T}\mathbf{w}_\mathrm{m})\mathbf{h}_p+(\mathbf{w}_q^\mathrm{T}\mathbf{w}_\mathrm{m})\mathbf{h}_q
        \end{aligned}
    \end{align}
\section{Derivation of final solution}
\label{Derivation of final solution}
The $\mathbf{Q}_1$ and $\mathbf{Q}_2$ in (\ref{gen_eigen}) are given by
\begin{align}
        \begin{aligned}
            \label{Q1&Q2}
            &\mathbf{Q}_1  = 
            \begin{pmatrix}
                q_{11} & q_{12}\\
                q_{21} & q_{22}
            \end{pmatrix}\\
            &\mathbf{Q}_2  = 
            \begin{pmatrix}
                1 & c\\
                c & 1
            \end{pmatrix}\\
            &q_{11} = h_p^2+2cgh_ph_q+c^2h_q^2\\
            &q_{12} = ch_p^2+\left(1+c^2\right)gh_ph_q+ch_q^2\\
            &q_{21} = q_{12}\\
            &q_{22} = c^2h_p^2+2cgh_ph_q+h_q^2
        \end{aligned}
    \end{align}
Decomposing $\mathbf{Q}_2$ by Cholesky factorization
    \begin{align}
        \begin{aligned}
            \label{cholesky}
            \mathbf{Q}_2=\mathbf{L}_2\mathbf{L}_2^{\mathrm{T}}
        \end{aligned}
    \end{align}
and plugging (\ref{cholesky}) in to (\ref{gen_eigen}) yields
    \begin{align}
        \begin{aligned}
            \label{eigen}
            \mathbf{M}\mathbf{v} = \lambda\mathbf{v}
        \end{aligned}
    \end{align}
where $\mathbf{v}=\mathbf{L}_2^{\mathrm{T}}\mathbf{u}$ and
  \begin{align}
        \begin{aligned}
            \label{M_mat}
            &m_{11}  = q_{11}\\
            &m_{12} = \sqrt{1-c^2}\left(gh_ph_q+ch_q^2\right)\\
            &m_{21} = m_{12} \\
            &m_{22}  = \left(1-c^2\right)h_q^2
        \end{aligned}
    \end{align}
(\ref{eigen_values}) - (\ref{u_vec}) can be easily solved from (\ref{eigen}) and (\ref{M_mat}). 
One also recognizes that $\tau$ and $\delta$ are the trace and determinant of $\mathbf{M}$, respectively.
\section{Proof of non-negativity}
\label{sec:proof-nonneg}
Here, we demonstrate the non-negativity of (\ref{xi}). Let  
    \begin{align}
        \begin{aligned}
        \label{eta_define}
            \eta = \lambda_\mathrm{max}-h_q^2-cgh_ph_q
        \end{aligned}
    \end{align}
which is the numerator of $\xi$ in (\ref{xi}).
Plugging (\ref{eigen_values}) into (\ref{eta_define}) yields
    \begin{align}
        \begin{aligned}
        \label{eta_define_2}
            \eta =& \frac{h_p^2-h_q^2}{2}+\sqrt{\frac{\tau^2}{4}-\delta}
        \end{aligned}
    \end{align}
It could be found from (\ref{tau&delta}) that 
\begin{align}
        \begin{aligned}
            \frac{\tau^2}{4}-\delta 
            =& \frac{\left(h_p^2-h_q^2\right)^2}{4}\\
            &+(ch_p+gh_q)(gh_p+ch_q)h_ph_q\\
            >& \frac{\left(h_p^2-h_q^2\right)^2}{4}
        \end{aligned}
    \end{align}
}
Thus, $\eta > 0$ and $\mathbf{u} > 0$. Therefore, $\mathbf{w}_\mathrm{m}$ is always non-negative and $\mathbf{h}_\mathrm{m}$ is also non-negative from (\ref{optimal_h_m}) and (\ref{optial_h_m}).

\bibliographystyle{IEEEtran}
% \bibliography{bibliography}

\end{document}